# A Comprehensive Survey on Architectural Advances in Deep CNNs: Challenges, Applications, and Emerging Research Directions

Saddam Hussain Khan[1]*, Rashid Iqbal[1]

[1]Artificial Intelligence Lab, Department of Computer Systems Engineering, University of Engineering and Applied Sciences (UEAS), Swat 19060, Pakistan

**Email:** saddamhkhan@ueas.edu.pk

## Abstract

Deep Convolutional Neural Networks (CNNs) have significantly advanced deep learning, driving breakthroughs in computer vision, natural language processing, medical diagnosis, object detection, and speech recognition. Architectural innovations including 1D, 2D, and 3D convolutional models, dilated and grouped convolutions, depthwise separable convolutions, and attention mechanisms address domain-specific challenges and enhance feature representation and computational efficiency. Advancements in activation functions, optimization methods, and regularization techniques, supported by large-scale datasets and high-performance computing, have further accelerated CNN development. Structural refinements such as spatial-channel exploitation, multi-path design, and feature-map enhancement contribute to robust hierarchical feature extraction and improved generalization, particularly through transfer learning. Efficient preprocessing strategies, including Fourier transforms, structured transforms, low-precision computation, and weight compression, optimize inference speed and facilitate deployment in resource-constrained environments. This survey presents a unified taxonomy that classifies CNN architectures based on spatial exploitation, multi-path structures, depth, width, dimensionality expansion, channel boosting, and attention mechanisms. It systematically reviews CNN applications in face recognition, pose estimation, action recognition, text classification, statistical language modeling, disease diagnosis, radiological analysis, cryptocurrency sentiment prediction, 1D data processing, video analysis, and speech recognition. In addition to consolidating architectural advancements, the review highlights emerging learning paradigms such as few-shot, zero-shot, weakly supervised, and self-supervised learning, as well as lifelong and federated learning frameworks. Future research directions include hybrid CNN-transformer models, vision-language integration, generative learning, 6D vision, and neural architecture search for scalable, interpretable, and energy-efficient DL solutions. This review provides a comprehensive perspective on CNN's evolution from 2015 to 2025, outlining key innovations, challenges, and opportunities.

## 1. Introduction

Convolutional Neural Networks (CNNs) have become powerful algorithms for interpreting image content, excelling in tasks like segmentation, classification, detection, and retrieval [1]. CNNs are also important in academia as this CNN architecture is being explored by industry giants like Google, Microsoft, etc., which have formed a separate team that keeps looking for new CNN architecture [2]. At the moment, all of the top deep learning (DL) competition winners mostly use CNN techniques and seem to have high-level domain proficiencies. Further, CNNs appeal for their ability to exploit spatial and temporal data correlations. The structure of CNN is made up of several layers and is completed by different stages: Convolutional operation, Non-linear activation function units, and subsampling [3]. It is a feed-forward, hierarchical multilayer network where each layer employs different convolutional kernels [4]. Convolution captures important local-related structures, processed by using a non-linear activation. This will promote a couple of abstraction learning and inherit the nonlinear nature of feature maps. The unique patterns of activation for different responses allow us to easily grasp the semantic variation/subtle in images. Later, subsampling aggregates obtained features and increased geometric transformation invariance [5]. The inherent feature extraction capability of CNN obviates the necessity for a distinct extraction of features [6]. CNN is characterized by graded learning, dynamic feature extraction, multitasking capabilities, and weight-sharing [7].

CNNs originated from LeCun's 1989 work, focusing on processing grid-structured data such as images and time series [7]. The architectural framework of CNNs drew inspiration from Hubel and Wiesel's research, closely mirroring the foundational organization of the visual cortex in primates [8]. The different learning stages of CNNs exhibit a striking similarity to the dorsal pathway of the visual cortex in primates [9]. In primates, the visual cortex initially processes information from the region mapping to the retina, where the lateral geniculate nucleus conducts filtering across multiple scales and normalizes contrast. Subsequently, various regions of the visual cortex engage in detection. These cortex regions exhibit similarities to layers of convolution and subsampling, while the inferior temporal area resembles the upper layers of CNNs responsible for image inference [10].

CNNs train using the backpropagation algorithm, adjusting weights based on target outcomes, akin to human brain learning mechanisms. They employ hierarchical feature learning, where lower- and mid-level features contribute to higher-level representations. This deep, layered learning process mirrors the columnar organization of the human neocortex, enabling dynamic feature extraction from raw input [11]. CNNs have significantly advanced computer vision, natural language processing (NLP), medical diagnosis, object detection, and speech recognition. Architectures such as 1D–3D CNNs, dilated and grouped convolutions, depthwise separable convolutions, attention-based models, and Neural Architecture Search (NAS) address domain-specific challenges. Structural refinements, including spatial-channel exploitation, multi-path design, increased depth and width, and feature-map refinement, enhance representation and computational efficiency. Additionally, preprocessing techniques like Fast Fourier Transform, structured transforms, weight compression, and low-precision computation further optimize performance. CNNs extract hierarchical features through convolution, non-linear activations, and subsampling, mirroring the human visual cortex.

Weight-sharing and graded learning facilitate multi-level feature representation, improving transformation invariance. The evolution from LeCun's early models to modern deep architectures underscores CNNs' ability to learn invariant representations, even from extensive unlabeled data. Transfer Learning (TL) further strengthens generalization across diverse recognition tasks.

Deep models have been demonstrated to be highly useful in solving tough learning problems since they can generate complex representations across varying levels of abstraction. One of the experiments showcased in the paper was the recognition of 200 diverse classes of images, showing the proficiency of deep CNNs in modern vision-based tasks [12]. It has been widely accepted that deep architectures increase the representational capacity of CNNs and they have become popular for tasks, associated with image classification and segmentation [13]. Therefore, structured taxonomy categorizes CNNs based on spatial exploitation, multi-path design, depth- and width-based models, dimensionality enhancement, channel boosting, and attention mechanisms, facilitating comparative analysis. Recent successes of deep CNNs are primarily linked to the interaction of abundant data and the increased power of the hardware. According to empirical evidence, deep CNNs can learn invariant representations in case they are trained on a sufficient amount of data, achieving the performance of humans. Beyond conventional supervised learning, deep CNNs exhibit significant proficiency in extracting discriminative features from vast amounts of unlabeled data. Contemporary research underscores the transferability of both low-level and high-level features across diverse recognition tasks, leveraging the principles of TL [14]. Furthermore, CNNs support various applications, including image classification, object detection, semantic segmentation, and integration with ViTs. Advanced paradigms like few-shot, zero-shot, weakly, and self-supervised learning have expanded CNNs' utility in low-labeled data scenarios. Moreover, CNNs contribute significantly to speech processing and natural language understanding, supported by large datasets and improved hardware acceleration. Emerging research frontiers encompass 6D vision, generative modeling, and meta-learning, offering novel directions for exploration. Collectively, these consolidated advancements in CNN architecture and learning strategies illuminate persistent research gaps and delineate prospective opportunities for future DL innovation.

### 1.1. Limitation of Existing Surveys

Prior surveys on CNN architectures, exemplified by Aloysius and Geetha (2017) [15] and Gu et al. (2018) [16], provided chronological overviews of architectures prevalent during specific periods. These works catalogued design patterns, yet often omitted explicit research questions, systematic evaluations, and critical discussions on inherent architectural design challenges. Earlier reviews, such as those by Rawat and Wang (2017) [17] and Liu et al. (2018) [18], focused on foundational CNN components and applications in narrow domains, such as image recognition or segmentation. These studies neglected dataset analyses, specialized taxonomies, or discussions on emerging paradigms, such as large vision models and multimodal frameworks (Dosovitskiy et al., 2020; [19] Radford et al., 2021) [20]. Consequently, their perspectives remained fragmented, lacking a unified analysis of architectural evolution and interdisciplinary challenges.

The absence of a principled taxonomy based on design principles, rather than temporal release, further limited the utility of prior surveys. Dhillon and Verma (2019) [21] analyzed classical architectures, such as AlexNet and VGGNet, but did not investigate why newer models, such as MobileNetV3 (Howard et al., 2019) [22] or InceptionV4 (Szegedy et al., 2017) [23], achieved superior performance. Similarly, Ajmal et al. (2020) [24] concentrated on segmentation applications without addressing advancements in efficiency-driven designs, such as NAS (Zoph & Le, 2017)[25] or dynamic convolutions (Chen et al., 2021) [26]. Such omissions hindered a holistic understanding of CNN innovation trajectories.

Existing surveys frequently prioritized domain-specific applications, such as object detection or medical imaging, over structural or algorithmic advancements. Zhang et al. (2019) [27] proposed a taxonomy centered on acceleration techniques but overlooked critical trends, such as federated learning (Konečný et al., 2016) [28] and explainability methods (Selvaraju et al., 2017) [29]. Earlier works inadequately addressed the ethical and computational challenges of deploying CNNs in resource-constrained environments, such as edge devices (Han et al., 2016; Li et al., 2021) [30], [31]. This narrow focus restricted their relevance to contemporary research demands.

The rapid proliferation of CNN variants since 2012 exacerbated these gaps. ShuffleNet (Zhang et al., 2018) [32] and Vision Transformers (ViTs) [19] represent paradigm shifts that prior surveys rarely anticipated or contextualized. While Gu et al. (2018)[16] reviewed architectures up to 2015, subsequent innovations in attention mechanisms (Wang et al., 2018) [33] and lightweight designs remained underexplored in comparative frameworks. Additionally, the generalized form of the limitations is mentioned as:

- Prior surveys provided basic overviews of CNN components, typically limited to standard convolutional layers and conventional pooling techniques, without addressing the emergence of advanced convolutional variants. Moreover, focused mainly on traditional pooling layers, neglecting an in-depth analysis of recent advanced pooling mechanisms that enhance spatial invariance, multi-scale representation, and information preservation. Furthermore, lacked a unified taxonomy and critical evaluation of the impact of these advanced convolutional and pooling variants on computational efficiency, scalability, and applicability across diverse tasks, including classification, detection, and segmentation.
- Existing reviews predominantly focus on domain-specific implementations, such as vegetation remote sensing and network intrusion detection, while lacking a systematic analysis of critical architectural advancements in CNNs. Consequently, they fail to provide a unified perspective on the evolution of CNN design strategies across diverse fields, including computer vision, NLP, and medical imaging.
- Existing studies predominantly emphasize supervised learning, with insufficient coverage of emerging paradigms such as weakly supervised, self-supervised, few-shot, and zero-shot learning. Discussions on adaptive pooling methods and advanced feature extraction strategies—vital for improving model generalization in data-scarce settings—remain limited.
- Optimization techniques critical for efficient CNN deployment, including structured transforms, low-precision computation, weight quantization, and energy-efficient architecture design, receive minimal attention. Similarly, the integration of CNNs with ViTs, graph neural networks, meta-learning frameworks, and NAS is rarely addressed in depth, hindering a comprehensive understanding of hybrid and scalable model development.

- In addition, several reviews offer only cursory attention to core challenges such as data heterogeneity, adversarial robustness, interpretability, ethical fairness, and catastrophic forgetting. Emerging domains—including 6D vision, generative learning, lifelong and continual learning, vision-language models, and NAS-driven optimization—lack systematic analysis. Benchmarking of CNN-powered models for classification, segmentation, and detection tasks remains largely absent, limiting insights into their practical deployment and performance in real-world scenarios.

### 1.2. Origins and Milestones

In the late 1990s to the early 2000s, substantial progress has been achieved in the methodologies and architecture of CNNs, making them adaptable to extensive, diverse, complex, and multi-class problems. These advancements involved alterations to processing units, strategies for optimizing parameters and hyperparameters, design patterns for layers, and adjustments to connectivity. The visibility of CNN applications significantly increased AlexNet's remarkable performance on the ImageNet dataset in 2012 [13]. Afterward, considerable innovations in CNNs have emerged, largely driven by the reconfiguration of processing units and the introduction of novel components. Instantly, Zeiler and Fergus [34] introduced layer-wise visualization concepts in CNNs, emphasizing feature extraction at low spatial resolutions, as demonstrated in VGG [35]. Modern architectures often follow the principle of a straightforward and uniform network topology inspired by VGG. Google's DL group introduced the concept of split, transform, and merge, exemplified by the inception block. This block incorporates branching within a layer to abstract features across multiple spatial scales [36]. In 2015, ResNet [37] introduced skip connections for deep CNN training, a technique that gained widespread acceptance and was later adopted by various subsequent networks such as Inception-ResNet, Wide ResNet, and ResNeXt [23], [38], [39].

Several architectural concepts, including Wide ResNet, Pyramidal Net, and PolyNet, have investigated the influence of multilevel transformation learning capabilities of CNNs. These technical variations have introduced concepts of cardinality adjustment and increased architectural width [40], [41]. Consequently, research has shifted to fine-tuning parameters and adjusting channel connections leading to the development of innovative architectural channel-boosting designing, utilizing spatial and feature-map information, and attention-based data processing [42]. Recent years have witnessed numerous insightful surveys on deep CNNs, elucidating their core components and alternative methodologies. Notably, analyses of well-known architectures from 2012 to 2015 have categorized their fundamental elements [16]. Recent surveys have explored diverse algorithms and applications of CNNs [6], [43], [44].

### 1.3. Contributions of This Survey

This paper surveys CNNs, outlining foundational concepts, architectures, and recent advancements. The analysis includes the advantages and limitations of CNNs, recommendations for practitioners, a review of available platforms and libraries, and an evaluation of computational costs. This survey systematically integrates the historical evolution, architectural innovations, and cross-domain applications of CNNs into an analytical framework. The primary focus examines CNN's limitations and highlights emerging strategies aimed at addressing these challenges. A concise overview of CNN architectures clarifies core operations, key networks, datasets, and the significance of CNNs. This survey is intended for a diverse audience seeking a comprehensive understanding of CNN mechanisms and constraints. The contributions are organized as follows.

#### 1.3.1. Comprehensive Taxonomy of Advanced Convolutional Layers

A systematic classification of classical and advanced convolutional layers is provided, encompassing transposed, grouped, deformable, steerable convolutions, and graph convolutional networks. Each variant is critically analyzed concerning its role in enhancing feature representation, adaptability, and computational efficiency. This taxonomy addresses the lack of systematic categorization in prior surveys.

#### 1.3.2. Advanced Pooling Techniques

The survey presents a detailed analysis of traditional and advanced pooling strategies, including mixed, stochastic, spatial pyramid, edge-aware pyramid, and genetic-based pooling. Their design principles, operational mechanisms, and contributions to spatial invariance and multi-scale feature aggregation are thoroughly evaluated.

#### 1.3.3. Architectural Evolution and Paradigm Shifts (2015–2025)

The evolution of CNN architectures is mapped from foundational models (LeNet, AlexNet) to recent innovations emphasizing simple-deeper architectures (VGG, Inception), depth-based networks (ResNet, Highway Networks), and attention-driven designs (CBAM, RAN). Lightweight architectures (MobileNet v3, GhostNet) and hybrid CNN-transformer models are analyzed for their efficiency in resource-constrained environments.

#### 1.3.4. Granular Analysis of Core CNN Components and Optimization Strategies

Core architectural components such as normalization (batch, group), activation functions, and regularization techniques (dropout) are discussed. In addition, the survey reviews optimization techniques, including Fast Fourier Transform, structured transforms, low-precision computation, weight compression, and energy-efficient designs tailored for edge AI applications.

#### 1.3.5. Critical Evaluation and Emerging Strategies

Milestone models (AlexNet, DenseNet, Inception) are examined alongside emerging strategies such as meta-learning, weakly supervised, and self-supervised learning frameworks. Feature-map exploitation mechanisms (Squeeze-and-Excitation Networks) and attention-based modules are evaluated for their role in real-time and low-resolution applications.

#### 1.3.6. Design Taxonomy for CNNs

Insights from over 30 architectural models, 15 convolution variants, and 25 pooling strategies are synthesized into a unified taxonomy. This framework offers a roadmap for developing efficient, interpretable, and ethically aligned CNNs, addressing challenges in scalability, sustainability, and cross-domain adaptability.

#### 1.3.7. CNN Applications Across Diverse Domains

CNN applications are systematically explored in computer vision (face recognition, pose estimation, action recognition), NLP (text classification, statistical language modeling), medical imaging (disease detection, radiology), cryptocurrency analysis (sentiment prediction, portfolio optimization), speech recognition, video analysis, and 1D sensor data processing.

#### 1.3.8. Challenges Identification and Proposed Solutions

The survey addresses persistent challenges such as data availability, interpretability, adversarial robustness, catastrophic forgetting, hyperparameter sensitivity, hardware constraints, and ethical biases. Proposed solutions include hybrid learning approaches, federated frameworks, adaptive pooling strategies, and robust feature extraction techniques.

#### 1.3.9. Emerging Research Directions

Underexplored avenues such as neuromorphic computing, pipeline parallelism, vision-language models, 6D vision, and high-energy physics applications are highlighted. The survey also examines advanced learning paradigms, including few-shot, zero-shot, weakly supervised, self-supervised, lifelong, and continual learning.

#### 1.3.10 CNN-Powered Models for Diverse Challenges

Specialized CNN architectures for classification (CVR-Net, Bayesian CNNs, and Ensemble CNN), segmentation (U-Net, FCNN, and Multi-Path CNN), and object detection (YOLO variants, Faster R-CNN, and Dynamic CNN) are benchmarked based on their efficiency, scalability, and deployment feasibility in constrained environments.

#### 1.3.11. Integration with ViTs and Capsule Networks

The synergistic integration of CNNs with ViTs, capsule networks, and graph neural networks is analyzed, with an emphasis on hybrid architectures. Deployment considerations, such as hyperparameter tuning, adversarial defenses, and energy-efficient implementations on edge devices, are discussed. Standardized metrics for bias detection and ethical AI guidelines are proposed to bridge theoretical advancements with practical applications.

This survey is structured to provide a comprehensive analysis of CNNs and their advancements. **Section 2** outlines the fundamental components of CNNs. **Section 3** traces the architectural evolution of deep CNNs, from early models to advanced frameworks. **Section 4** examines recent innovations, including attention mechanisms, hybrid CNN-transformer architectures, and lightweight models for resource-constrained environments. **Section 5** reviews CNN applications. **Section 6** highlights key challenges such as data limitations, interpretability, adversarial robustness, and ethical considerations while identifying future research directions. **Section 7** presents fast processing techniques, including structured transforms, low-precision computation, and weight compression to enhance efficiency. **Section 8** explores emerging research fields, such as ViTs, few-shot learning, lifelong learning, multi-task learning, and NAS. **Section 9** benchmarks CNN models for classification, segmentation, and detection tasks, emphasizing hybrid integrations. Finally, **Section 11** concludes the survey by summarizing key insights and proposing directions for future advancements in CNN research and applications. The paper layout is illustrated in Figure 1.

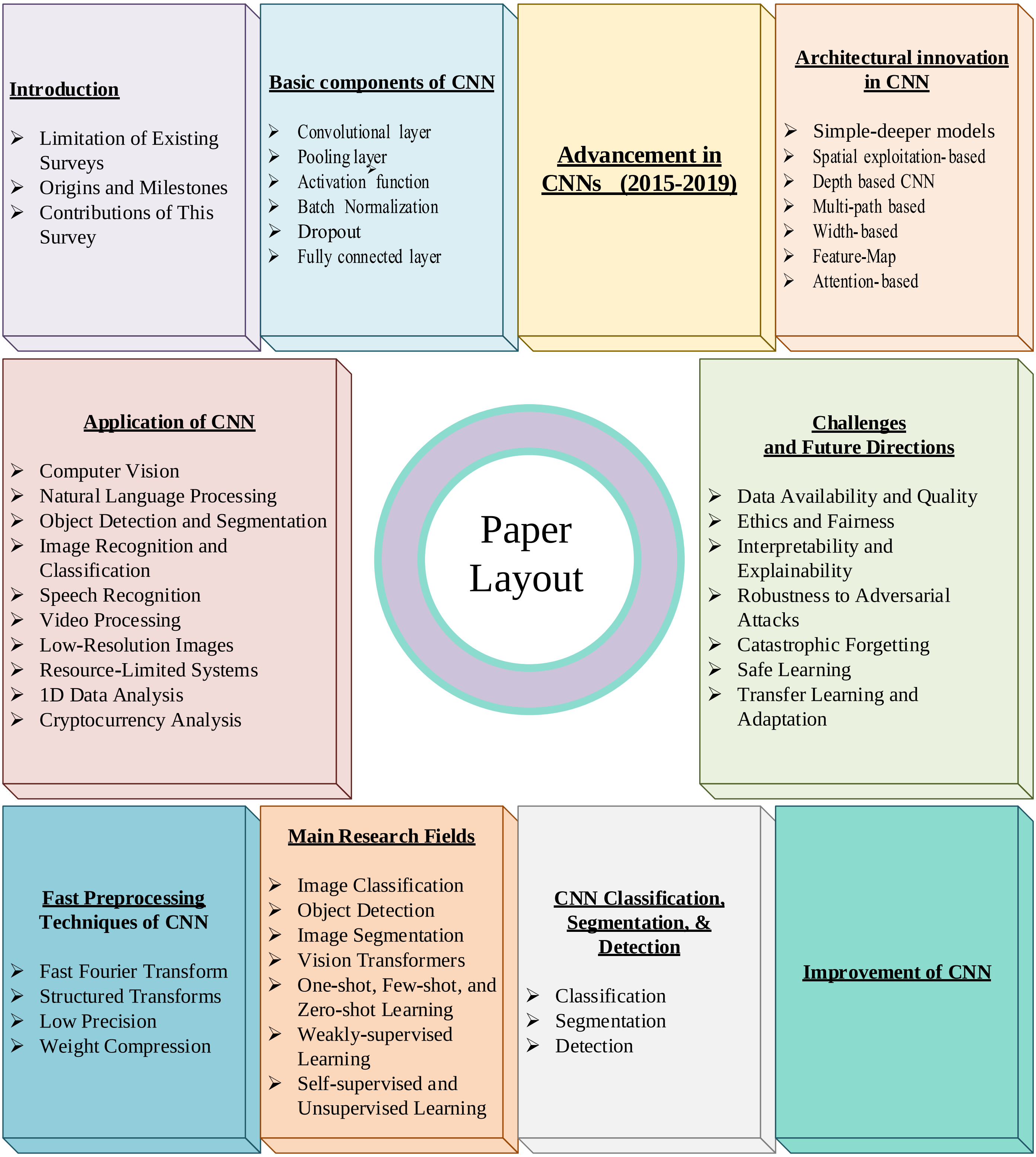

Figure 1: Paper Layout.

## 2. Basic components of CNN

Currently, CNNs have solidified their position as one of the most extensively utilized techniques within the realm of DL, particularly in applications associated with visual data. CNNs exhibit a remarkable ability to extract meaningful representations from grid-like data, and their recent advancements in performance across various DL tasks have been highly commendable. Furthermore, a standard CNN architecture typically comprises a sequence of layers that alternate with convolutional layers and pooling layers, culminating in one or multiple fully connected layers towards the network's end. In certain scenarios, a global average pooling layer may be employed as a substitute for a fully connected layer. CNNs achieve remarkable performance through a combination of essential components, including convolutional layers and activation functions, which extract features and introduce non-linearity. Moreover, regularization methods like batch normalization and dropout mitigate overfitting and enhance training.

**2.1.Convolutional layer**

An array of convolutional kernels is employed within the convolutional layer, where each neuron effectively operates as an individual kernel. It's important to emphasize that in cases where a kernel exhibits symmetry, the conventional convolution operation shifts into a correlation operation [45]. The primary role of these convolutional kernels lies in the segmentation of the input image into smaller units commonly referred to as receptive fields. This segmentation of the image into compact blocks serves as a critical step for the extraction of distinct patterns representing various features. The convolutional kernel, or filter, functions like a scanner moving across the input image. At each position, it multiplies the image's corresponding elements by their weights and sums the resulting products as shown in Figure 3. This operation captures local patterns within the image and is mathematically expressed as below:

$$f_l^k(p,q) = \sum_c . \sum_{x,y} i_c\,(x,y) e_l^k(v,u) \qquad (1)$$

In this equation, $i_c(x,y)$ represents an element extracted from the input image tensor, It undergoes element-wise multiplication with $e_l^k(v,u)$ which signifies the index of the $k^{th}$ convolutional kernel within the $l^{th}$. Meanwhile, the resulting feature map generated by the $k^{th}$ convolutional operation is formally stated as $\mathbf{F}_l^k = [f_l^k(1,1), \dots, f_l^k(p,q), \dots, f_l^k(P,Q]$. Definitions of the mathematical symbols are provided in equation (1).

An exceptional characteristic of convolutional operations is their intrinsic ability to share weights. Using a kernel that is situated across the image and has a consistent set of weights allows for a diverse set of features to be extracted from the image. Because a kernel that has a uniform set of weights will run over the different parts of the image, the key property enhances the parameter efficiency of CNNs in comparison to fully connected networks. However, it should be noted that Convolution is not a universal concept. The models are designed to solve specific problems by using different types of convolutions, including filter type and size, padding method, and the orientation of convolution [4].

**2.1.1. Transposed Convolutions**

Upsampling of feature maps is carried out utilizing transposed convolutions also as deconvolutions or fractionally stride convolutions [46]. While standard convolutions reduce the spatial dimensions, transposed convolutions enlarge them. Transposed convolutions are relevant for tasks such as image segmentation, image generation, and image-to-image translation, in which high-resolution images should be produced from low-resolution ones [47]–[49]. They make use of learnable parameters to control upsampling, including the modification of stride and padding with an aim towards achieving a desired output size. However, if not appropriately managed, there is the possibility of overlapping receptive fields that give rise to artifacts and checkerboard patterns. Therefore, modifying stride, padding, and dilation parameters as necessary can manage the final resolution while minimizing these artifacts. In image generation, low-resolution images are resized to high-resolution images using transposed convolutions. This helps to avoid both checkerboard-patterned artifacts and improve the quality of the generated images through a carefully tuned stride, padding, and dilation.

### 2.1.2. Grouped Convolutions

Grouped convolutions were first introduced in AlexNet, which allowed training deep CNNs on limited GPU memory. It shall also allow the model to make wider networks with fewer parameters compared to standard convolutions. This would reduce training time and increase the representation learning of the model for better accuracy. Without a grouped convolution, AlexNet would not be that efficient and accurate [41]. Furthermore, grouped convolution realizes these benefits by rearranging the filter bank in the channel dimension into a sparse, block-diagonal structure. This architecture reduces the number of parameters to be learned, similar to regularization methods, as overfitting is restrained and deeper, more precise, and efficient networks can be built. This is utilized in ResNeXt, a generalization to the ResNet, using small, shifted blocks rather than full-sized blocks as described above. This maintains accuracy but dramatically reduces model complexity: in these architectures, the same much smaller block is repeated uniformly. Further studies revealed that growing such stacked blocks can lead to better improvements rather than increasing the width and depth of the network. Huang et al. [50] presented a two-stage process to train grouped convolutions. The first stage involves the pruning of less important filters based on a sparsity-inducing regularization technique. This is followed by an optimization stage where the weights on the groups are fixed, and further training ensures that the desired sparsity pattern is maintained in each group. In their work, Zhang et al. [51] have tackled the limitations brought in by manual selection concerning the number of groups and introduced Groupable ConvNet. This work combines grouped convolutions with NAS to get the model to automatically determine the number of groups in an end-to-end fashion and avoid manual selection.

### 2.1.3. Deformable Convolutions

Typically, CNNs have fixed geometric structures; hence, it limits their ability to handle objects well under varying shapes or poses. Geometric transformation invariance is one of the major robustness issues in computer vision. Traditional ways to avoid this may include increasing the variety in the training data by augmentation or features or algorithms resistant to geometric variations need specific pre-defined knowledge which may not generalize well to new situations. To improve these limitations, Dai et al. [52] proposed deformable convolutions and deformable ROI pooling. Deformable convolutions focus on the relevant object areas to obtain more informative feature maps with learned offsets added to the original kernel positions in convolution. These enable the kernel to adapt to some geometric transformations such as translation and rotation by increasing its receptive field area.

Similar is the deformable ROI pooling, which, prior to pooling, learns the feature offsets. Zhu et al. [53] pointed out that, while the geometric transformations can indeed be adapted by the deformable convolutions, these can be distracted at training time by nearby features that may be irrelevant. The authors proposed the model of deformable ConvNets v2 with an added modulation mechanism that was then used throughout the network. This in turn gives the model more control over the spatial distribution and influence of its samples.

Furthermore, this is highly desirable for hardware implementation in terms of improving computational efficiency. Dong et al. [54] combined depthwise separable convolutions with deformable convolutions. In the offset prediction, it replaced all the convolution layers with depthwise convolutions, which drastically reduced the computational costs.

#### 2.1.4. Steerable Convolutions

Equivariance is an essential attribute of a neural network involved in dealing with visual tasks. While invariance refers to a type of network output that remains the same irrespective of changes in the input, equivariance presupposes that a network should return outputs that would transform predictably with the input. Thus, if there is a rotation in an image, a network should return a rotated output. A regular CNN faces problems in this regard because it has been designed for positional invariance. After this limitation, Cohen and Welling [55] suggested the steerable CNNs. The authors presented a framework in which the filters can adapt other things to more complex transformations than just the positional shifts, such as rotations and reflections. By incorporating various linear transformations within a group structure, steerable CNNs are more flexible and computationally efficient compared to traditional methods. This is a very promising approach in the handling of continuous transformations, although detailed research will be needed regarding its effectiveness in high-dimensional spaces.

Building on this, the work of Weiler et al. [56] extended the idea of steerable CNNs to 3D space. The data representation used was scalar, vector, and tensor fields, and they showed the equivariant convolutions for these representations that allow the network to commute with rigid body motions. The Euclidean group E(2) [57], along with its subgroups, provides the mathematical basis to adapt kernel space to rotations and reflections in 2D images.

#### 2.1.5. Graph Convolutional Networks

Recently, Graph Convolutional Networks (GCNs) have become a powerful tool for analyzing graph-structured data. Early models, such as those by [58], convert graphs to matrices with the result of losing topological information. In contrast to traditional CNNs that excel with Euclidean data, GNNs are designed to manage non-Euclidean structures. [59] investigated GNN architectures inspired by CNNs, emphasizing graph spectral and hierarchical clustering methods. Kipf and Welling [60] introduced a simplified and efficient GCN model based on spectral graph convolutions, which aggregates information from neighboring nodes to learn node representations. Graph convolution techniques can be classified into spatial and spectral domains. Spatial domain methods operate directly on graph structures, whereas spectral domain methods exploit graph spectral properties. To improve GCNs, researchers have developed various enhancements. [61] proposed Adaptive GCN, which learns adaptive graph structures. [62] created Dual-GCN to efficiently capture both local and global information. [63] pioneered spatial domain graph convolutions with the NN4G model, which directly aggregates neighbor information and incorporates residual connections.

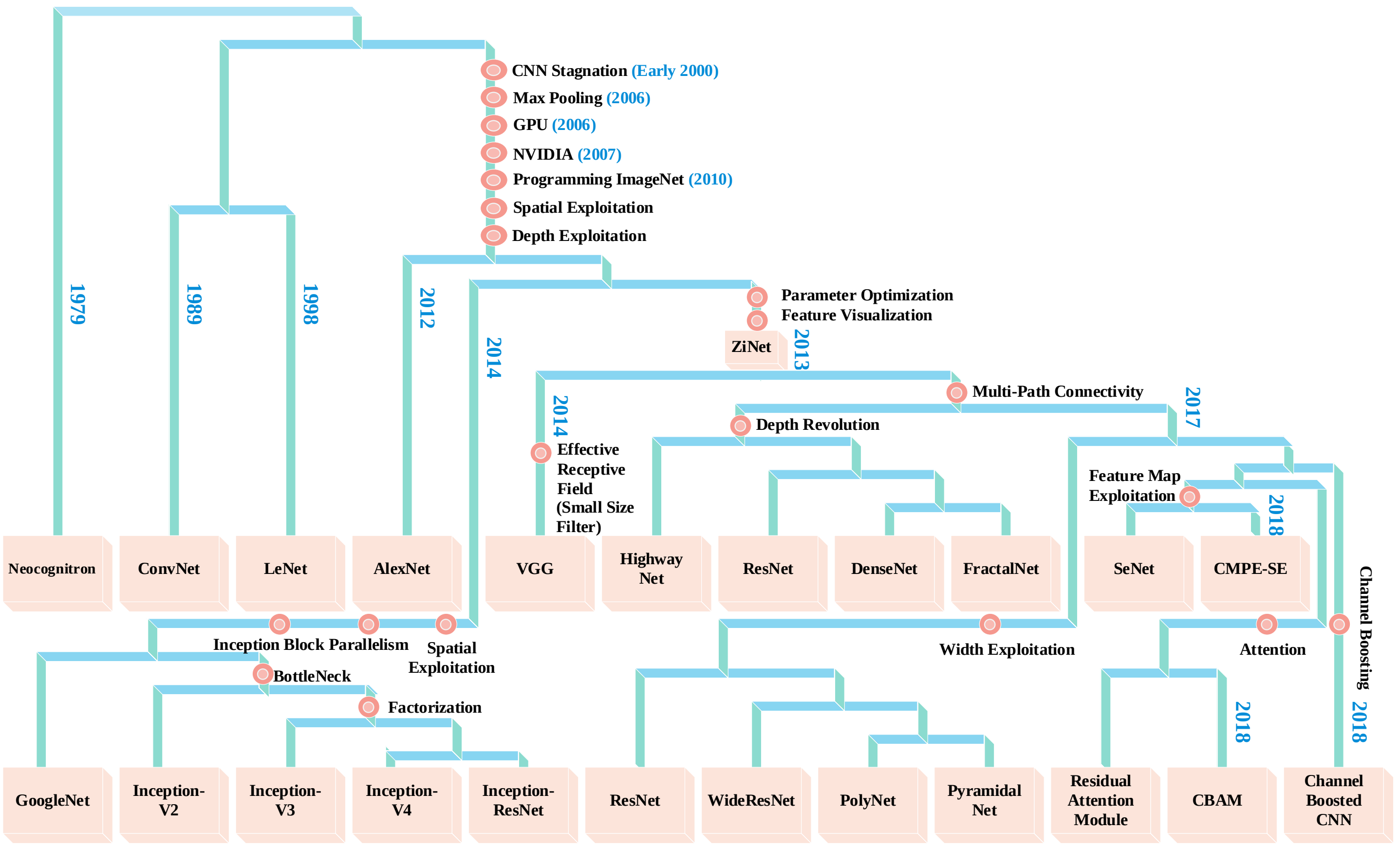

Figure 2: Architectural Innovations in Deep CNNs: From ConvNet to Modern Architectures.

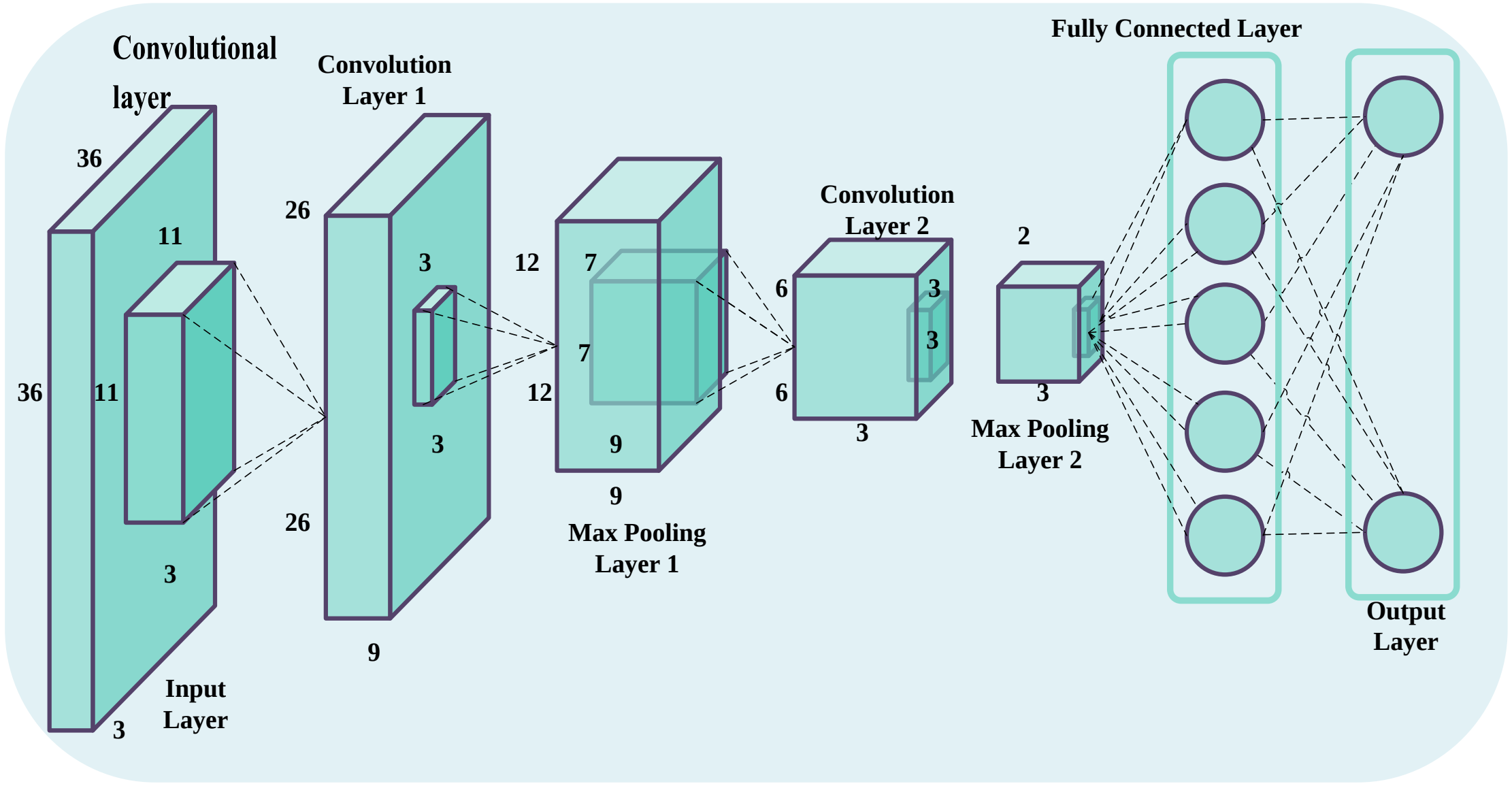

Figure 3: Convolutional Layer: Learnable filters slide across input data, extracting meaningful features.

## 2.2. Pooling layer

Convolution identifies significant patterns within images, where the exact feature locations may not be crucial as long as their relative positions are preserved. Pooling, or downsampling, plays a role here by summarizing information within local image areas to capture dominant features. This reduces the network's sensitivity to small shifts or distortions [64] [5]. Pooling reduces feature map sizes, simplifying the network and mitigating overfitting, thereby improving generalization as indicated in Figure 4. Various pooling techniques like max pooling (selecting the highest value) and average pooling offer specific task advantages.

$$\mathbf{z}_l^k = g_p(\mathbf{F}_l^k) \qquad (2)$$

In equation (2), the pooling operation is depicted, where $\mathbf{z}_l^k$ denotes the pooled feature map of the $l^{th}$ layer corresponding to the $k^{th}$ input feature map $\mathbf{F}_l^k$. The function $g_p(.)$ specifies the type of pooling operation employed. Pooling operations aid in extracting feature combinations that exhibit invariance to translational shifts and minor distortions [5], [65] This reduction of feature map size to an invariant feature set helps regulate network complexity and improves generalization by mitigating overfitting. CNNs employ various pooling techniques, such as max pooling, average pooling, L2 pooling, overlapping pooling, and spatial pyramid pooling [37], [66].

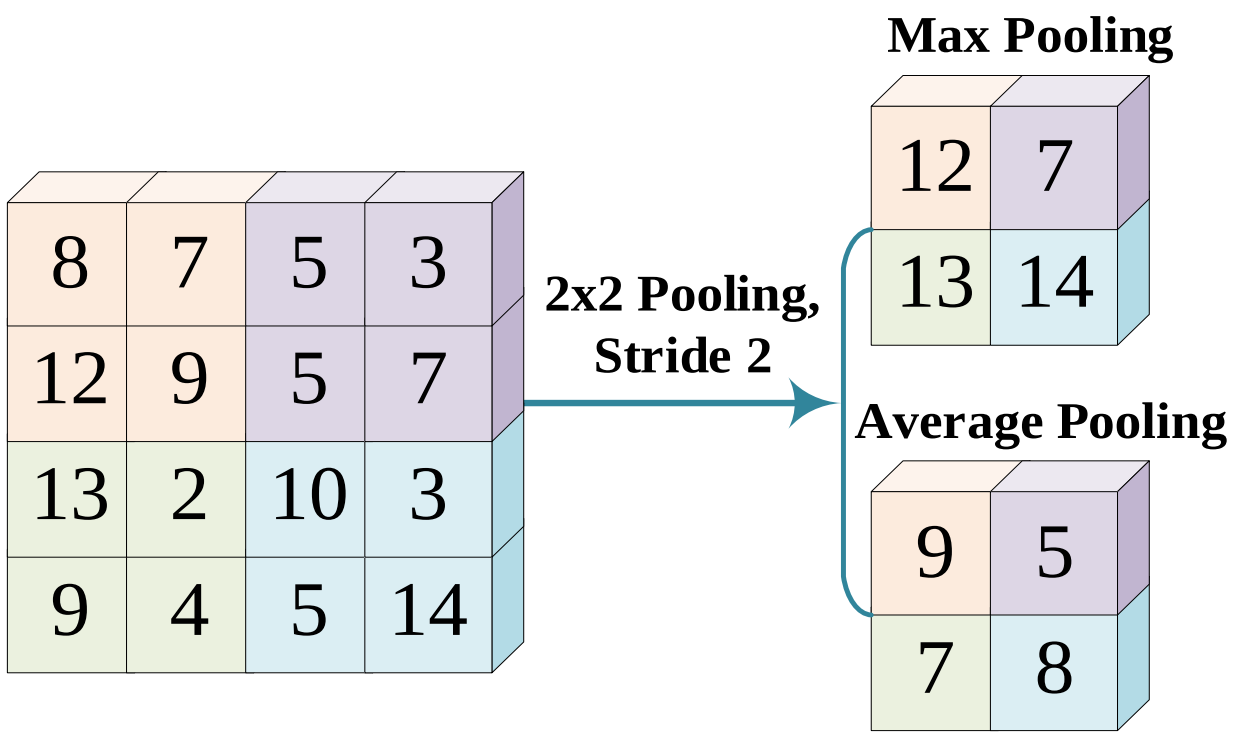

Figure 4: Pooling Layer: Downsamples feature maps by reducing spatial dimensions using max-pooling or average-pooling.

### 2.2.1. Average Pooling

The concept of averages applied to feature extraction harks back to the earlier days of image processing. The average pooling in a neural network refers to a process where the original input is divided into rectangular regions; then, the average value for each is computed. This operation downsamples the input by retaining only the important information. Even with the improvement of spatial dimensions, this may not capture minute variations in a feature or salient details concentrated in only a portion of the image using average pooling.

### 2.2.2. Max Pooling

Max pooling in CNNs reduces the dimensions of feature maps spatially but with useful information. In this approach, the maximum value inside a window of interest in an input is selected. So, salient features can be captured well, in particular, edges and textures. It is useful in the case of image processing. The process is different from average pooling, which takes the mean value because it focuses on the strongest values. Yet, it may also lead to a loss of some details and a decline in the accuracy of certain features.

### 2.2.3. Mixed Pooling

Mixed pooling combines the merits of max and average pooling in CNN. Unlike other pooling methods where either max or average pooling is used consistently, the contribution of mixed pooling is known with the introduction of randomness, usually performing a random turn between max and average during training. Mathematically, mixed pooling is defined in equation (3)

$$s_j = \lambda \cdot max(a_i) + (1-\lambda) \cdot mean(a_i) \qquad (3)$$

In this formula, $\mathbf{S}_j$ is the output of the pooling for region j, $a_i$ are the activations within region j, and λ is a random variable that can take either the value 0 (average pooling) or 1 (max pooling). This mixed pooling chooses between max and average pooling at random to produce regularization in the network and improved performance.

#### 2.2.4. L$_P$ Pooling

Sermanet et al. [67] proposed L$_P$ pooling, arguing that it provides a better generalization than max pooling. This has been implemented by computing the weighted average of features in a pooling region using the equation (4) below:

$$\boldsymbol{S}_j = \left(\frac{1}{|\boldsymbol{R}_j|}\sum_{i \epsilon \boldsymbol{R}_j} \boldsymbol{a}_i^p\right)^{1/p} \qquad (4)$$

In this equation, $\boldsymbol{S}_j$ denotes the result of the pooling layer at position j, $a_i$ value of a feature at position i in the pooling region $\boldsymbol{R}_j$, and p is a hyperparameter that varies from 1 to infinity. When p = 1, L$_P$ pooling becomes average pooling, and when p = ∞, it becomes max pooling. When p > 1, L$_P$ pooling produces a middle ground between the above two.

#### 2.2.5. Stochastic Pooling

Stochastic pooling, proposed by [34], is a kind of pooling in which, within a certain pooling region, there is a random selection of an activation. While max pooling always selects the maximum activation and average pooling treats all the activations equally, stochastic pooling assigns a probability to each activation depending on the value of that particular activation. These probabilities are determined through the process of normalization of activations in a given region, amounting to a multinomial distribution. A random selection is then made from this distribution, and the corresponding activation becomes the pooled output. The introduced randomness adds some noise to the network that prevents overfitting. Since stochastic pooling considers a variety of activations and not just the maximum, it can be said to combine some elements of both max and average pooling. While having a focus on stronger activations, stochastic pooling still allows weaker ones to contribute. Hence, for some applications, this may prove beneficial over traditional pooling techniques.

#### 2.2.6. Spatial Pyramid Pooling

Spatial Pyramid Pooling (SPP) divides an image into variable-size regions and aggregates features over these regions. This technique, borrowed from the Bag-of-Words model, has become a sort of imperative in computer vision. Inserting an SPP layer after some convolutional layers allows for fixed-length output representations, irrespective of the size of the input image, thus avoiding resizing or cropping of images, hence increasing flexibility. In deeper network stages, the inclusion of SPP has taken object detection algorithms such as YOLO to the next level. SPP is useful and efficient in feature extraction from images of variable sizes, hence finding its perfect application in the domain of computer vision.

#### 2.2.7. Region of Interest Pooling

Region of Interest (ROI) pooling is fundamental in CNN-based object detection and segmentation, focusing on specific image regions using bounding boxes. After convolutional layers generate feature maps, ROI pooling extracts relevant features from designated areas. RoI pooling, therefore, takes these feature maps and ROI coordinates as inputs. The challenge is that such RoIs can have different sizes. It achieves this by taking each ROI and dividing it into a grid of equal-sized cells and performing max pooling for every cell. So, these variable-sized ROI features have now become fixed-size outputs, thus preparing them for the subsequent fully connected layers. RoI Pooling The performance of an object detection model is enhanced by effectively extracting features from specific regions.

### 2.2.8. Multi-Scale Orderless Pooling

Multi-Scale Orderless Pooling (MOP) enhances the invariance of CNNs with no compromise in their discriminative power. It pools features both over the whole image and over smaller patches at multiple scales. It combines global features that capture the overall structure of the image with local features, conveying fine information. Specifically, VLAD encoding is insensitive to the order of features while aggregating local features. In this way, the proposed model shows further advanced performance in handling complicated variations in image content and perspective. The concatenation of global features with aggregated local features forms the final image representation and leads to a descriptor that effectively combines both pieces of information.

### 2.2.9. Superpixel Pooling

Superpixels in an image are regions in which pixels share similar properties. The final result is the over-segmentation of the whole image to have a representation with less complexity. This decreases the number of elements an image has and enhances computational efficiency by capturing what is usually referred to as perceptually significant structures. Superpixel pooling utilizes these over-segmented regions by defining the areas of pooling based on superpixel boundaries. This approach found a wide field of applications for object detection and semantic segmentation tasks. Performing superpixel pooling, it integrates the low-level image information into the learning process, hence possibly improving such models.

### 2.2.10. Principal Component Analysis Networks

Principal Component Analysis (PCA) networks leverage PCA for pooling by training large PCA filter stages with binary hashing and block histograms to represent them effectively. This results in a so-called PCA network, or PCANet for short, which is lauded due to its simplicity and efficiency in computation.

Further developing this idea, the Two-Stage Oriented PCA takes the help of PCA in a pooling layer for extracting noise-robust features. Unlike PCANet, OPCA focuses on noise covariance analysis rather than hashing or local histogram formation. Although PCANet is pretty efficient, its merger with OPCA for noise robustness will add more strength to PCANet.

### 2.2.11. Compact Bilinear Pooling

Bilinear models have demonstrated strong performance across various visual tasks but produce high-dimensional feature representations, which complicates their practical use. To address this challenge, compact bilinear pooling was introduced as a method for generating low-dimensional yet highly discriminative image features. While bilinear pooling effectively captures rich interactions between image features, its high dimensionality presents difficulties. Several approaches have been proposed to address this issue. Low-rank bilinear pooling reduces dimensionality prior to the bilinear transformation, whereas compact bilinear pooling employs a sampling-based approximation to achieve significant dimension reduction without performance loss. Additionally, second-order pooling convolutional networks incorporate bilinear interactions directly within convolutional layers. Compact bilinear pooling projects high-dimensional features into a lower-dimensional space using techniques such as Tensor Sketch. These projected features are then aggregated to create a compact, global image representation.

#### 2.2.12. Lead Asymmetric Pooling

Traditional pooling in CNNs normally down-samples the feature maps by a fixed pooling factor. Though effective in many scenarios, this may be far from being sufficient in extracting multiscale features, particularly multi-lead ECGs of varying scales. Therefore, Lead Asymmetric Pooling (LAP) overcomes this limit by borrowing experience from research into image recognition, which indicates that multilevel pooling is good for the extraction of more information from multiple scales. LAP further extended this concept to use different pooling factors at different levels within the network. This kind of strategy will provide a more adaptive and flexible way to capture the multiscale features of the ECG data, which may lead to better performance by CNNs.

#### 2.2.13. Edge-Aware Pyramid Pooling

The Pyramid Pooling Aware of Edges is a series of pooling to preserve the edge information. This approach combines the feature of edge detection in the pooling process to keep more information in the image representation. In turn, this will allow edges as a key feature and will improve the performance for detection and segmentation where the accuracy of boundary detection is essential.

#### 2.2.14. Spectral Pooling

Spectral pooling does this pooling in the frequency space. It does so by first taking the DFT of input features, then performing dimensionality reduction by cropping the frequency domain representation and converting it back to the spatial space using the inverse DFT. Unlike max-pooling, spectral pooling uses linear low-pass filtering; therefore, it avoids losing more information for a certain size of the output. Since image information is concentrated in lower frequencies, removing higher frequencies by cropping the spectrum preserves little effect on image content. This method is also computationally efficient by exploiting matrix truncation and fast Fourier transforms for convolutional operations.

#### 2.2.15. Row-Wise Max Pooling

Row-wise max pooling selects the highest value from each row of a feature map, forming an output vector. This should be a good method to achieve rotational invariance, which is important in the processing of 3D shape data. It would be possible to obtain features robust against object rotations by the use of the panoramic representation of 3D shapes combined with row-wise max pooling. It is effective in enhancing performance for tasks related to 3D shape classification and retrieval.

#### 2.2.16. Intermap Pooling

Intermap Pooling (IMP) is a pooling method that pools the convolutional filters and then conducts an intra-group max pooling. This captures both common and spectrally varied features by selecting the maximum activation value from each position of every group. As a result, this strengthens the robustness of the model to spectral variations; thus, it is particularly effective for tasks such as speech recognition. The technique has demonstrated strong performance in LVCSR tasks and achieved results competitive with leading methods.

#### 2.2.17. Per-Pixel Pyramid Pooling

Per-pixel pyramid pooling achieves multi-scale feature extraction by using different-sized window pooling operations at every pixel. Instead of utilizing one large-size pooling window, multiple windows are utilized to retain the fine details for capturing coarse and detailed information. This approach concatenates the results from performing the pooling operations at different scales for every pixel, hence producing feature maps that provide a comprehensive representation of the input, encompassing several levels of detail.

#### 2.2.18. Rank-Based Average Pooling

Rank-Based Average Pooling (RAP) strikes a good balance between average and max pooling. In the case of average pooling, it may be too radical in averaging out all values, since this may cause the critical few activations to vanish; on the other hand, max pooling discards all but the maximum. These issues are handled by RAP with the use of a ranking mechanism. A rank threshold is applied to RAP such that it averages out only those activations above a certain rank. Large activations and low-value ones result in better representation. The resulting flexibility means that RAP can be fine-tuned for more or less feature information, making the pooling solution at hand more versatile.

#### 2.2.19. Weighted Pooling

The weighted pooling handles the different influences of each neuron within a pooling region by assigning different weights to each neuron. Unlike other pooling techniques, where all neurons are treated equally, weighted pooling takes the relative importance of contributions for every neuron. The pooled value is computed by a sum of a product between activation at each neuron and its corresponding weight. Weighted pooling yields a resulting representation that is more representative of the information contained in the original features.

#### 2.2.20. Genetic-Based Pooling

Genetic-based pooling can be considered an optimization technique that leverages the power of concepts in evolutionary algorithms toward improving pooling in a neural network. In this approach, the weights over pooling are modeled as a population of candidates that evolve iteratively utilizing optimization. In practice, this method first generates a large number of diversified sets of weight configurations before each set trains the model. The former evaluates to select only the best-performing sets of weights to further evolve through genetic operations such as crossover and mutation. This process leads to optimal weights generated iteratively. Since more effective weight combinations pertaining to specific tasks could be discovered, Genetic-Based Pooling has a higher probability of outperforming traditional methods.

### 2.3. Activation function

Activation functions serve as a gateway within the network, determining which information proceeds forward as depicted in Figure 5. An appropriate activation function significantly enhances the learning process. Equation (5) illustrates its application to a feature map after convolution. In this equation (5), $\mathbf{F}_l^k$ represents the output of the convolution, which transforms the activation function $g_a(.)$. This activation function introduces non-linearity, resulting in a modified output $\mathbf{T}_l^k$ for the $l^{th}$ layer. Various activation functions, including sigmoid, tanh, $max_{out}$, SWISH, ReLU, and its derivatives like leaky ReLU, ELU, and PReLU, have been explored in academic literature. These investigations are well-documented in the works of researchers. [68]. However, ReLU and its variations have gained prominence due to their effectiveness in addressing the vanishing gradient problem, as discussed in research by [68]. A more recent addition to the field of activation functions is MISH, which, as demonstrated by [69], has showcased superior performance compared to ReLU in several cutting-edge deep neural networks applied to standard benchmark datasets.

$$\boldsymbol{T}_l^k = g_a(\boldsymbol{F}_l^k) \quad (5)$$

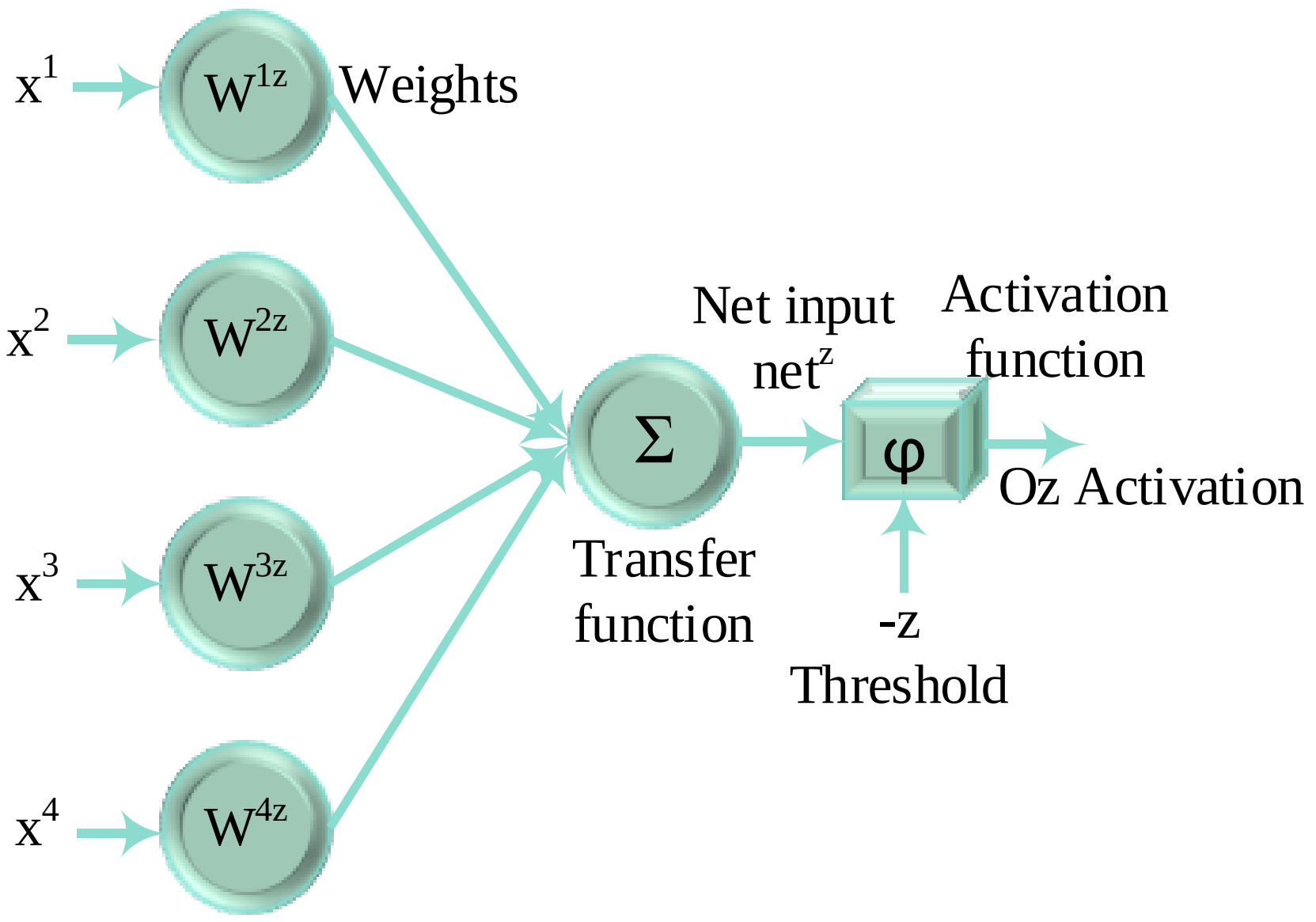

Figure 5: Activation functions guide information flow, enabling non-linear learning.

## 2.4. Batch Normalization

The input data at each layer fluctuates during training deep CNN and causes translational invariance. To mitigate this issue, batch normalization standardizes the data within each mini-batch before forwarding it to the subsequent layer, as represented in Figure 6. This approach improves both the speed of training and the stability of the network. Throughout the training process, the internal properties of the data flowing through the network can change, affecting learning speed (convergence). This necessitates the careful selection of initial parameter values and often requires using a very small learning rate. Equation (6) demonstrates how batch normalization addresses these challenges by transforming the feature maps.

$$\boldsymbol{N}_l^k = \frac{\boldsymbol{F}_l^k - \mu_B}{\sqrt{\sigma_B^2 + \varepsilon}} \quad (6)$$

In Equation (6), $\boldsymbol{N}_l^k$ signifies the normalized feature map, $\boldsymbol{F}_l^k$ represents the input feature map, while $\mu_B$ indicate the mean and variance of a feature map within a mini-batch. A small value denoted as epsilon (ε) is added to the variance to enhance numerical stability during calculations. This serves two primary purposes: preventing division by zero errors and ensuring smoother computations throughout the training process [70]. Additionally, it stabilizes gradient flow and contributes to regularization, enhancing the network's generalization.

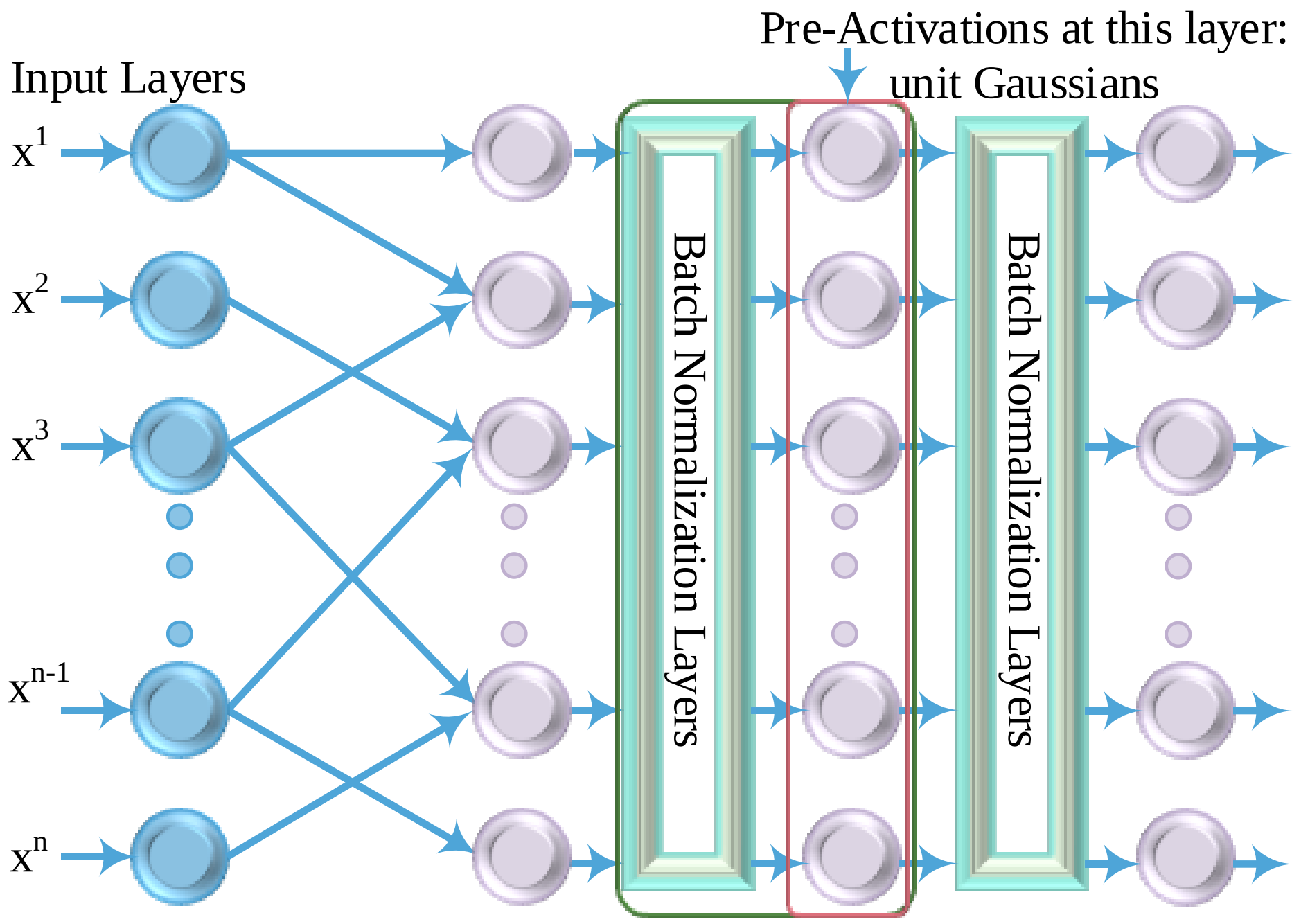

Figure 6: Normalizes intermediate feature maps within a mini-batch, improving training stability and accelerating convergence.

### 2.4.1. Group Normalization

Traditional image features, such as SIFT, HOG, and GIST, have inherent group structures. For instance, HOG features are computed by dividing pixels into cells and normalizing orientation histograms in each cell. Similarly, higher-order features, including VLAD and Fisher Vectors, take operations on grouped representations. Combining these with the previous techniques, Group Normalization divides the feature channels into groups and normalizes the features in each group, independent of the batch size. This approach recognizes the fact that channels within a group may share similar statistical properties. For instance, early convolutional layers often learn filters of similar responses, which might suggest that their corresponding channels could be grouped and normalized together.

Higher-level features, although more abstract, also benefit from grouping based on factors like frequency, shape, or texture. Neuroscience research supports normalizing neural responses across different receptive field centers and spatial frequencies. GN captures this intuition by applying group-wise normalization to deep neural network features, thus leveraging the shared statistical properties within each group.

## 2.5. Dropout

The dropout technique adds a regularization dimension to neural networks, thus enhancing generalization capability by randomly excluding specific units or connections based on a predetermined probability. In the neural network context, it is recognized that intricate, interrelated nonlinear dependencies can emerge among multiple connections and potential cause of overfitting [71], [72] Through the random omission of specific connections or units. Dropout results in the creation of various sparser network configurations. Eventually, a single network is chosen, characterized by its reduced weight parameters as illustrated in Figure 7.

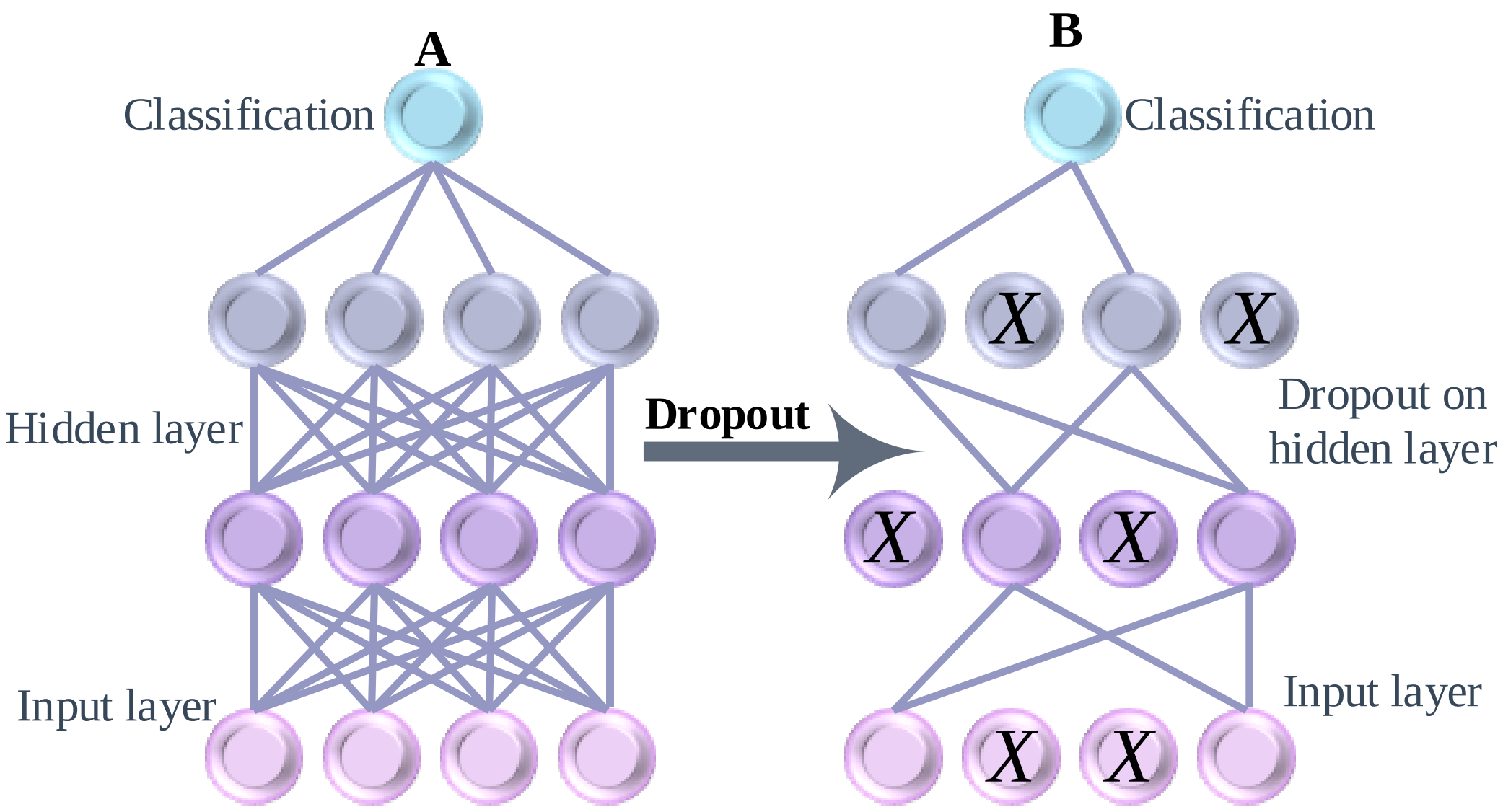

Figure 7: Randomly deactivates neurons during training to prevent overfitting and promote model generalization.

### 2.6. Fully connected layer

The fully connected layer is typically positioned at the neural network's target level, primarily designed for performing classification tasks. In contrast to pooling and convolution layers, this component engages in a comprehensive operation that gathers information from earlier feature extraction stages. This approach involves conducting a thorough analysis of the outputs from all preceding layers [73], as shown in Figure 8. Subsequently, it produces a non-linear sequence of distinctive features, which are subsequently utilized for data classification [7].

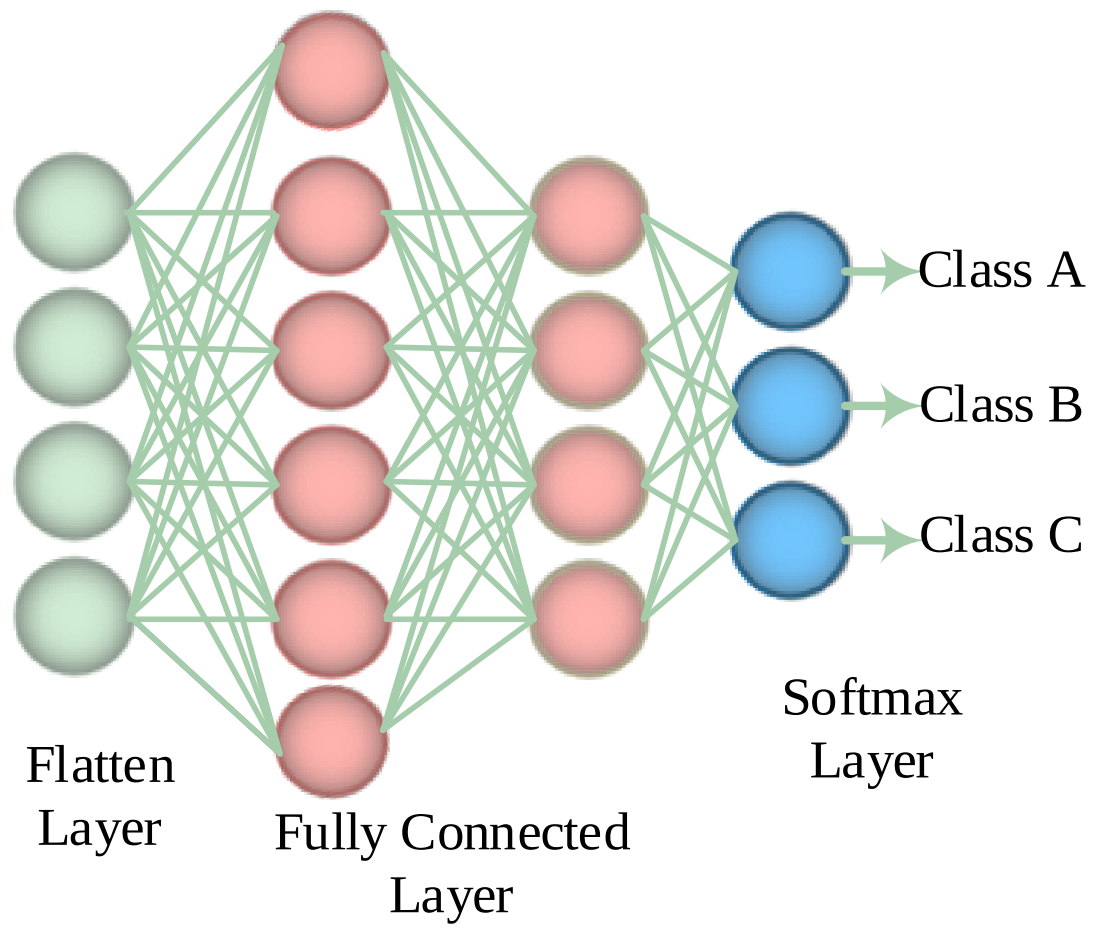

Figure 8: Neurons connect to all neurons in the previous layer, commonly used for classification and regression tasks

## 3. Advancements in CNNs (2015-2025)

Ongoing research in the field of CNNs holds significant potential for further architectural and conceptual advancements. From 2015 to 2025, there was a notable increase in CNN performance and facilitating advancements in more complex applications as illustrated in Figure 2. The efficacy of a CNN's representation varies on its depth, encompassing a hierarchy of features, from simple to complicated abstractions, which is crucial in addressing intricate challenges. Nevertheless, deep CNNs face a significant obstacle known as the vanishing gradient problem, prompting researchers to explore various remedies. Addressing the challenges posed by deep architectures, 2015 marked a milestone with the introduction of pioneering techniques such as information gating between layers, skip connections and cross-layer channel connectivity. These innovations notably enhanced the convergence speed of deep CNNs.

Numerous empirical studies have highlighted the exceptional performance of Cutting-edge deep architectures like VGG, ResNet, and ResNext, particularly in demanding tasks related to recognition and localization. These foundational models (VGG, ResNet, ResNeXt) established the basis for advancements in object detection and segmentation. Key architectures in these domains, including Single Shot Multibox Detector (SSD), Region-based CNN (R-CNN), Faster R-CNN, Mask R-CNN, and Fully Convolutional Network (FCN), extensively build upon the concepts introduced by these foundational models. Diverse innovative detection algorithms, including Feature Pyramid Networks and Cascade R-CNN, have also been influenced by these foundational models., and Libra R-CNN, further evolved and improved these foundational architectures.

In 2016, researchers initiated investigations into the stacking of multiple transformations, considering both depth and parallel configurations. This approach has shown efficacy in addressing complex problems through effective learning. Researchers often fused previously proposed architectures in hybrid forms to improve the performance of deep CNNs. In 2017, there was a notable emphasis on creating versatile blocks that could be integrated at different stages within CNN architectures, introducing the concept of attending to spatial and feature-map (channel) information.

In 2018, channel boosting, as introduced by [74] emerged as a strategy aimed at enhancing CNN performance by capturing distinctive features and leveraging prior knowledge through TL. However, deep and wide architectures posed significant computational and memory challenges in resource-constrained settings. To address this, adaptations such as knowledge distillation, compact network training, and techniques like pruning, quantization, and hashing were implemented. Although achieving impressive performance, deep CNNs face challenges due to high computational demands, limiting deployment on resource-constrained devices. In response, researchers have developed efficient architectures. GoogleNet introduced smaller networks by replacing traditional convolutions with point-wise group convolutions, achieving significant computational efficiencies without compromising accuracy. ShuffleNet further advanced this approach by integrating point-wise group convolutions with a unique channel shuffling technique. This innovative strategy substantially reduces the computational load of the network while preserving accuracy, enabling the deployment of powerful CNNs across a broader spectrum of devices.

Recently, consistent progress has been made in CNN architectures and research endeavors are concentrated on designing innovative blocks that enhance network representations by exploiting feature-map information and manipulating input involving artificial channels. Simultaneously, there is a noticeable trend toward crafting lightweight architectures that maintain performance while accommodating the constraints of resource-limited hardware environments.

## 4. Architectural Innovations in CNN

From 1989 to the present, the domain of CNN architecture has seen a series of significant advancements. CNN advancements can be broadly classified into three key areas: parameter optimization, regularization techniques, and, notably, architectural innovations. While the first two areas are essential for ensuring stable and efficient training, the most substantial performance improvements in CNNs have arisen from strategic modifications to processing units and the introduction of novel architectural components. This emphasis on network design has enabled CNNs to unlock entirely new capabilities. Many of these innovations in CNN architectures have centered on addressing two critical aspects: augmenting network depth and harnessing spatial characteristics. This study structured various CNNs into various categories: spatial exploitation, depth augmentation, multi-path architectures, width expansion, feature-map exploitation, and attention-based CNNs as shown in Figure 9. Additionally, the paper synthesizes architectural specifics of leading CNN models, including parameters and performance on standard datasets. To facilitate further exploration, the paper provides a curated list of online resources related to CNN architectures, datasets, and implementation platforms. Lastly, for a comprehensive understanding, the study evaluates the strengths and weaknesses of various architectures within each category.

### 4.1. Simple-deeper models

In 2012, a pivotal advancement in object recognition occurred with the introduction of AlexNet. This innovative model, drawing inspiration from LeNet, introduced a deep architecture by stacking five convolutional layers sequentially and employing a greater number of filters per layer. This substantial increase in depth and complexity compared to earlier models, such as LeNet, enabled AlexNet to achieve remarkable performance gains. The success of AlexNet played a critical role in showcasing the potential of CNNs across diverse computer vision tasks and catalyzing the widespread adoption of CNNs in the field.

The success of AlexNet in computer vision, achieved through CNN operations, led to the development of several straightforward yet deep models. These "simple-deeper" models, characterized by their emphasis on stacked convolutional layers and activation functions to improve accuracy, aimed to build upon AlexNet's achievements. One notable example is VGG [11], which expanded upon the achievements of AlexNet by demonstrating the effectiveness of increased depth. Instead of using large filters like AlexNet, VGG employed multiple stacked 3x3 filters. This approach, combined with adding more layers, enabled VGG to capture more intricate patterns. VGG's success underscored the potential of deeper architectures using smaller filters to enhance CNN performance. However, scaling up both depth and width in neural networks involves a trade-off. While larger networks can offer greater power, they also pose challenges. The significant increase in parameter count requires substantial computational resources, which can restrict their deployment on devices with limited capabilities.

### 4.2. Spatial Exploitation-based CNNs

CNNs exhibit a vast array of parameters and hyperparameters, encompassing critical aspects like weights, biases, layer count, processing units (neurons), filter dimensions, stride, activation functions, and learning rates, among others, as chronicled in the works of [75] and [76] and confirmed in contemporary literature. The essence of convolutional operations lies in their ability to consider the local context of input pixels, offering a versatile canvas for exploring diverse levels of correlation by employing filters of varying dimensions.

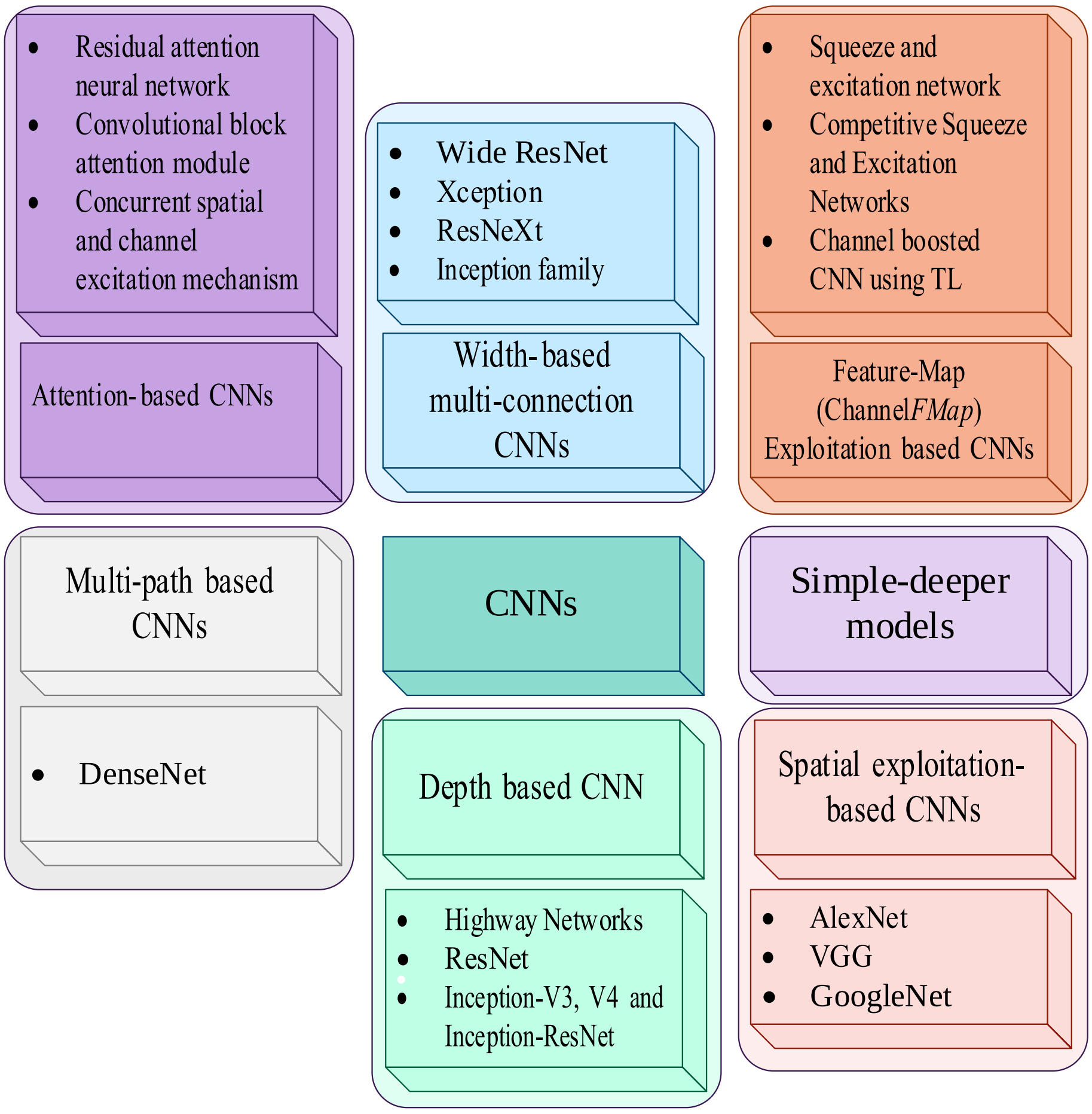

Figure 9: Various CNNs into various categories: spatial exploitation, depth augmentation, multi-path architectures, width expansion, feature-map exploitation, and attention-based CNNs.

These distinct filter sizes function as containers for discrete levels of detail; conventionally, smaller filters demonstrate proficiency in capturing intricate, fine-grained information, while their larger counterparts specialize in extracting broader, coarser features. This valuable insight catalyzed research endeavors in the early 2000s, leading to the strategic utilization of spatial filters to elevate CNN performance and delve into the intricate relationship between spatial filters and the learning dynamics of the network. Pioneering studies from this era conveyed that meticulous filter adjustments empowered CNNs to exhibit robust performance, effectively encompassing both coarse and fine-grained nuances.

#### 4.2.1. AlexNet

AlexNet stands as a significant milestone in the evolution of CNNs, signifying a breakthrough in image classification and recognition tasks [13], [77]. This pioneering architecture substantially augmented CNN’s capacity for learning, achieving this by venturing deeper into its layers and implementing various strategies for parameter optimization. The fundamental architectural design of AlexNet is vividly portrayed in Figure 10 and Figure 11. Notably, during the early 2000s, hardware limitations imposed constrictions on the potential scale of deep CNN architectures [13].

To fully unlock the formidable representational capabilities of deep CNNs, AlexNet was trained in parallel on two NVIDIA GTX 580 GPUs, ingeniously bypassing these hardware constraints. In the case of AlexNet, the network's depth was expanded from the prior 5 layers seen in LeNet to a more substantial 8 layers, thereby extending CNN applicability to a broader spectrum of image categories. While greater depth generally enhances generalization across various image resolutions, it also introduces the challenge of overfitting. Hinton's work [78] gives inspiration by introducing an algorithm to tackle the issue that randomly omitted certain transformational units during training, fostering the acquisition of robust features. Moreover, AlexNet was using the rectified linear unit (ReLU) as a non-saturating activation function, a very important design decision because it helped speed up convergence by at least alleviating in part the vanishing gradient problem [71]. For better generalization and to reduce overfitting, the architecture also used overlapping subsampling along with local response normalization. Among these, larger filter sizes-as in 11 × 11 and 5 × 5-used in initial layers, which had earlier been conventionally avoided, are notable. More importantly, considering the large strides AlexNet took in terms of efficient learning, it has secured its place as a very important milestone within the area of contemporary CNNs. Due to this, there came about a change in research and new inventions in the architecture of CNNs.

#### 4.2.2. VGG

VGG, proposed by Simonyan et al. [35] constitutes a milestone in the area of image recognition CNNs. It introduced a simple but potent design principle- a modular structure of considerable depth, with 19 layers. This was designed to investigate how depth is related to the ability of a network to learn complex features from an image [35]. Unlike prior networks, which favored larger filters, such as 11x11 and 5x5, respectively, VGG innovatively uses a sequence of stacked 3x3 filters. As was shown, research manifested that these smaller filters could achieve comparable performance but with much lower computational complexity introduced by fewer parameters. This insight influenced CNN research a lot and urged the researchers toward using smaller filters.

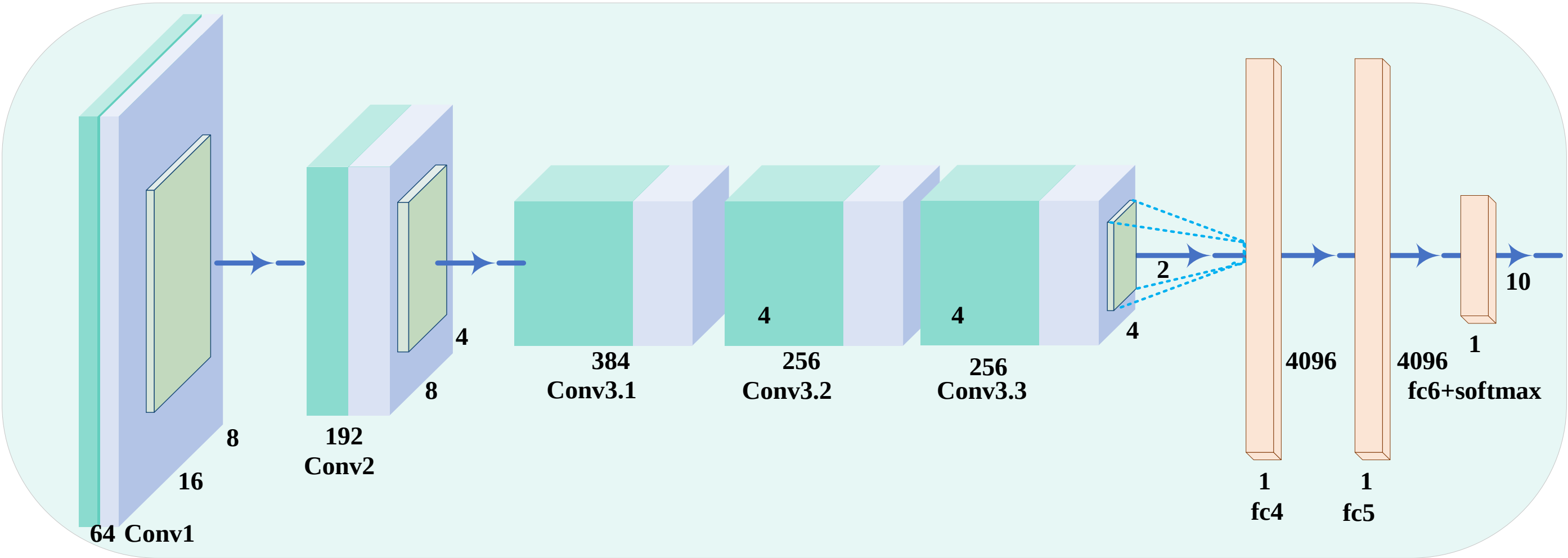

Figure 10: AlexNet: 2012 breakthrough with 5 conv layers, max-pooling, ReLU, and dropout for image classification.

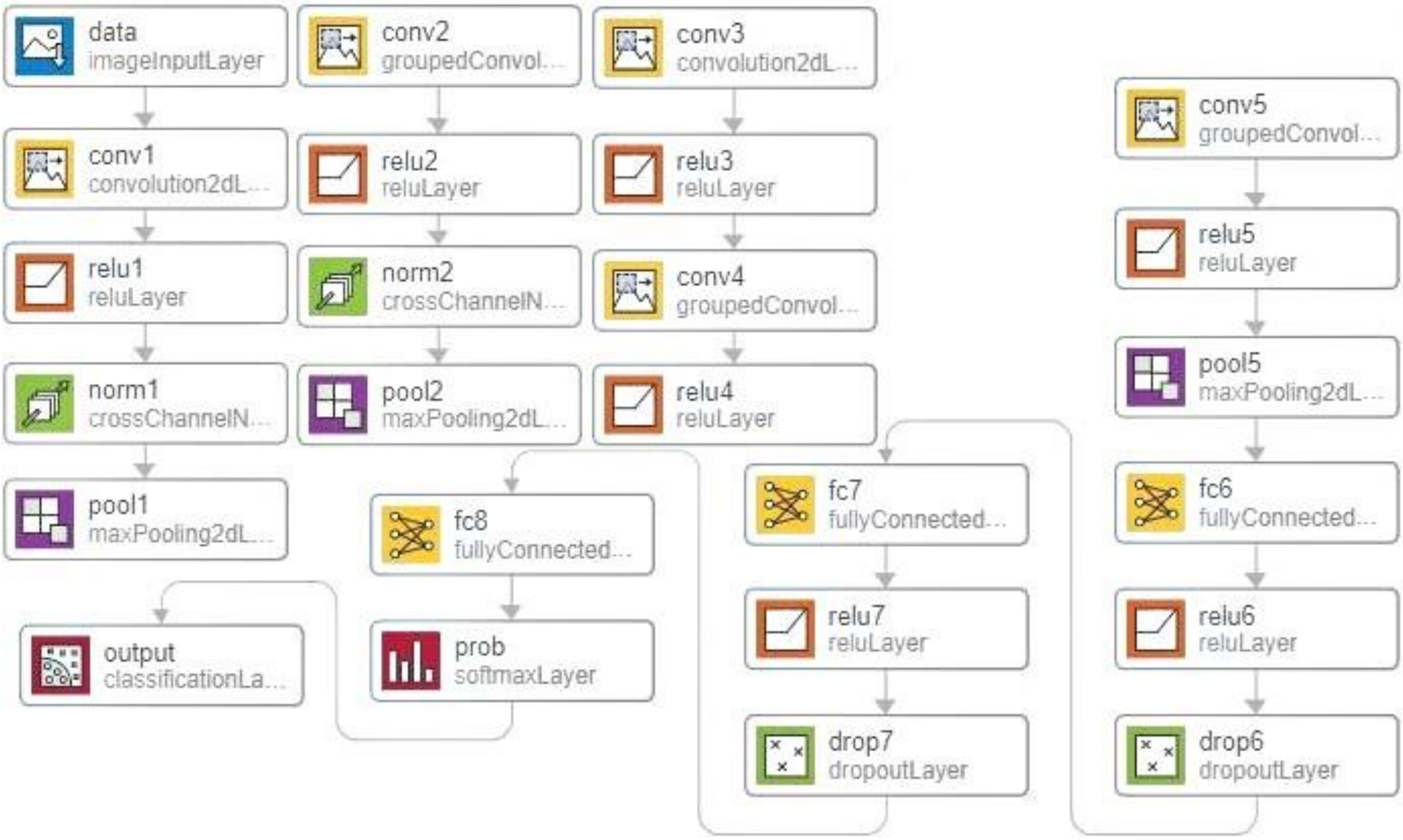

Figure 11: Architecture of AlexNet: A groundbreaking deep CNN was introduced in 2012.

To further manage network complexity, VGG incorporated 1x1 convolutions between regular convolutional layers. These 1x1 filters facilitated the learning of linear combinations of existing feature maps, adding complexity without a drastic increase in parameters. Additionally, VGG utilized max-pooling after convolutional layers and padding techniques to preserve spatial resolution throughout the network as represented in Figure 12, Figure 13and Figure 14. Moreover, VGG excelled in both image classification and localization tasks, achieving second place in the 2014 ILSVRC competition. Its enduring impact stemmed from its simplicity, consistent architecture, and pioneering depth. However, a notable limitation of VGG was its high parameter count (138 million), which rendered it computationally intensive and restricted its deployment on resource-constrained devices.

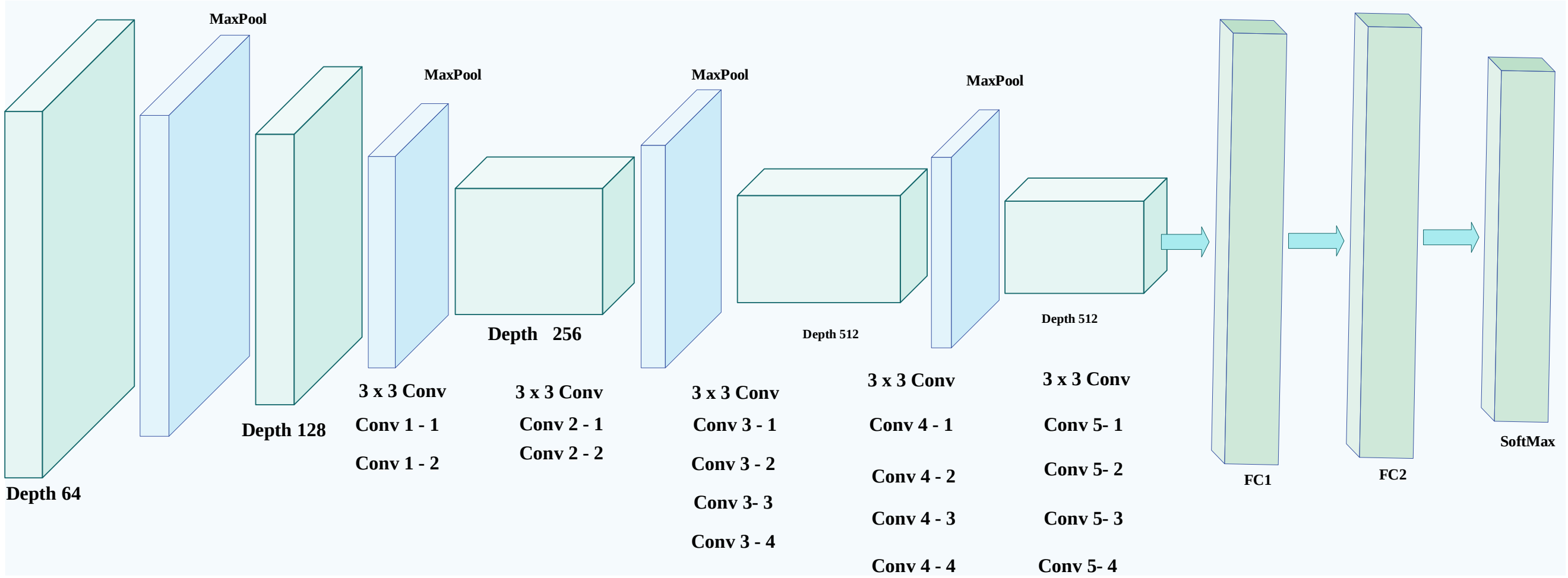

Figure 12: Basic Diagram of VGG: Deep CNN with 3×3 convolution filters for learning complex features.

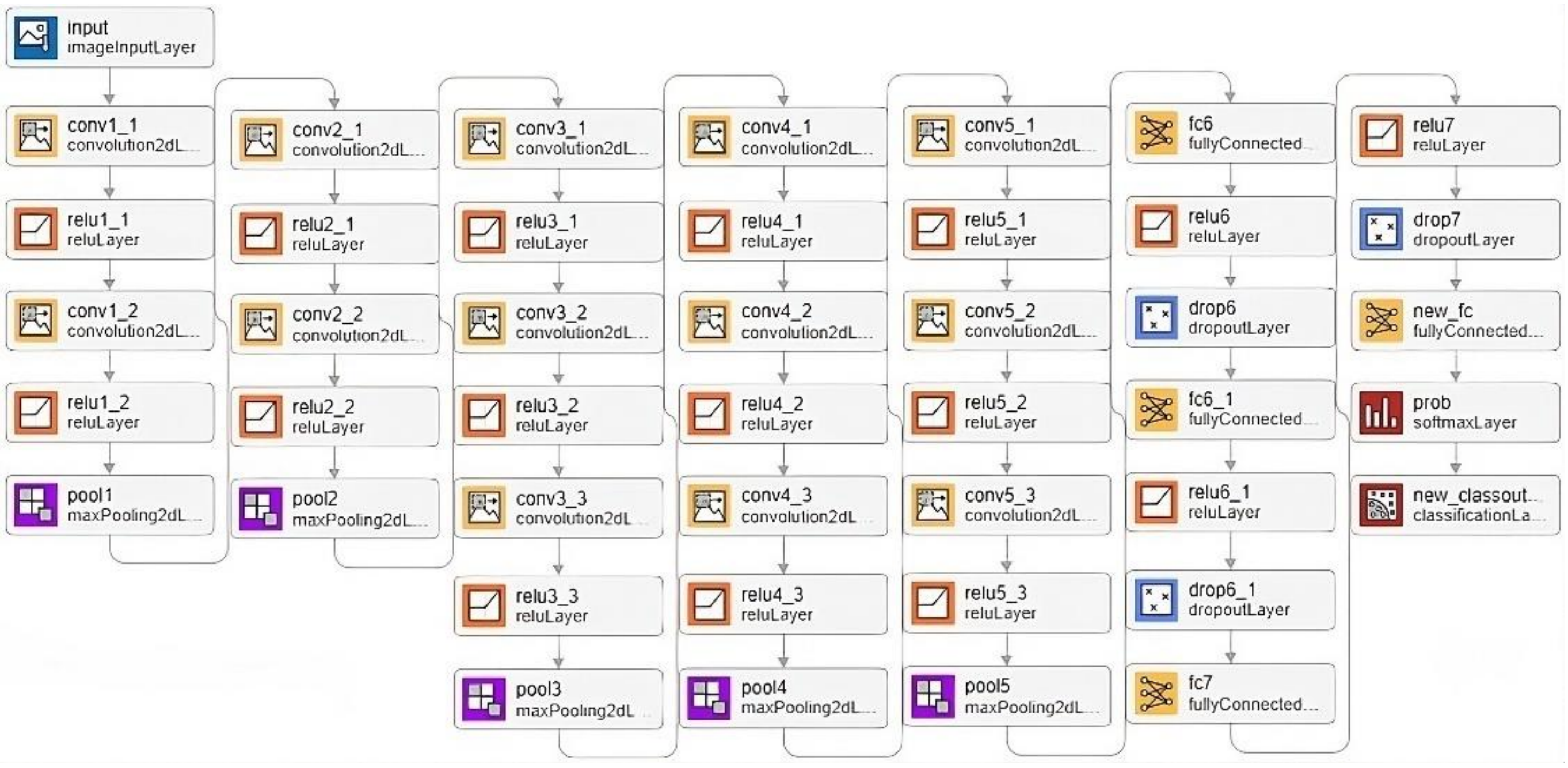

Figure 13: Architecture of VGG-16: A deep CNN with 16 weight layers, including 13 convolutional layers and 3 fully connected layers.

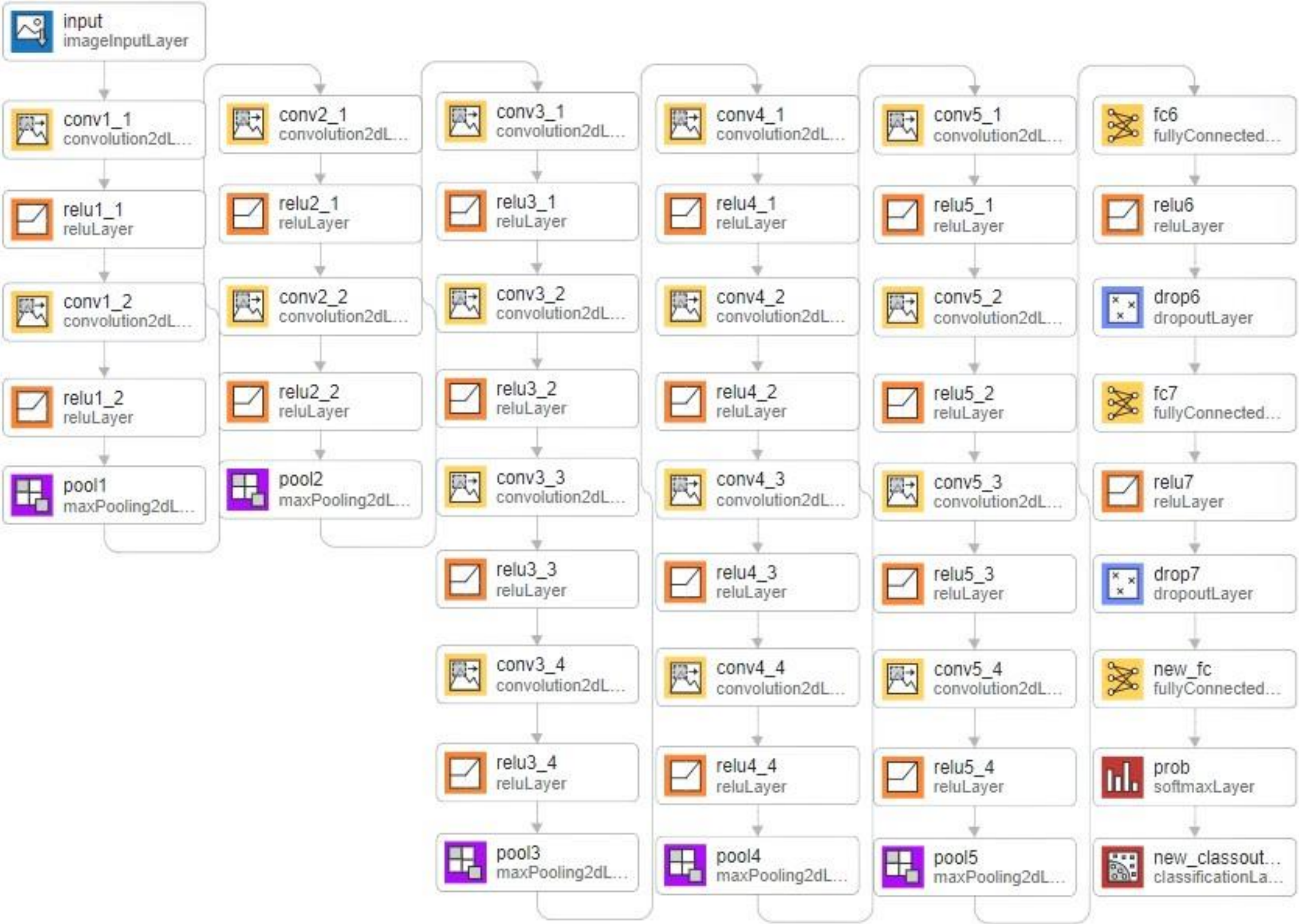

Figure 14: Architecture of VGG-19: 19-layer CNN with 16 convolution layers, 3 fully connected layers, and softmax output.

#### a) Block Models

The block models, as would be quite rightly implied by the name, are the architectural models that would combine function blocks such as Inception modules, Residual learning blocks, Dense blocks, and others. These function blocks are assembled to solve particular challenges in their inside modules to serve a greater performance of the neural network. These all have a lot more layers compared to simpler, shallower architectures, but remarkably remain more compact with fewer parameters. By way of example, DenseNet169 has nearly ten times the number of layers found in VGG19 and occupies just a fraction of VGG19's size and parameter count. However, as the CNNs started to go deeper and wider, problems such as overfitting and computation cost became significant. The Inception [12–14] block resolves this by implementing the Hebbian principle of neurons that fire together, wiring together, and multi-scale processing. This efficiently uses resources and achieves VGG's performance while using much fewer parameters and less computation.

It is designed to resolve a problem referred to as the vanishing gradient, which significantly hinders the effectiveness of the training in deep networks. It introduced "shortcut connections" that let the information directly flow between layers, skipping a few intermediate layers. Such a connection allows the gradients to propagate more effectively; therefore, deeper models can now be enabled for better performance. Densely connected convolutional networks (DenseNet) [17] aim to improve feature propagation. Unlike traditional CNNs, where each layer connects solely to its immediate neighbors, DenseNet uses feature maps from all preceding layers as inputs. This strategy promotes better feature reuse and information flow throughout the network.

### 4.2.3. GoogleNet

GoogleNet also referred to as Inception-V1, claimed the championship title at the 2014 ILSVRC competition. Its principal objective centered on achieving exceptional accuracy while simultaneously minimizing computational overhead [79]. A pioneering innovation introduced by GoogleNet was the inception block, a fundamental concept within CNN architecture. This novel approach involved the integration of multi-scale convolutional transformations using the STM paradigm, as visually represented in Figure 15, Figure 16, and Figure 17. In the realm of GoogleNet, the conventional convolutional layers underwent substitution with compact blocks, reminiscent of the Network in Network (NIN) architecture. These blocks incorporated filters of different sizes ($1 \times 1$, $3 \times 3$, and $5 \times 5$), enabling the capture of spatial information across various scales, containing both fine and coarse-grained details. The application of the STM concept in GoogleNet was way more effective at handling several problems related to learning multiple variations of the same image category even at resolutions different from those it had been trained on.

In light of this, GoogleNet solved the problem by proposing a bottleneck layer with a $1 \times 1$ convolutional filter as a preparation step before applying filters of larger sizes. It also employed sparse connections in which each output feature-map doesn't share a connection with all the input feature-maps. It was one of the ways of effectively conducting computations by removing information redundancy to reduce the computational cost for irrelevant feature-maps. Further, GoogleNet optimized the connection density by incorporating global average pooling in the final layer, and without using a fully connected layer. By optimizing such parameters, the number has been reduced from 138 million initially to a lean 4 million. To further tune network performance, GoogleNet implemented batch normalization and used RmsProp for the optimizer. Another development was the inclusion of auxiliary learners, which further improved convergence rates for the network. However, GoogleNet did exhibit certain limitations. Its heterogeneous topology required intricate customization from one module to another, potentially complicating its application. Furthermore, a noteworthy drawback was the presence of a representation bottleneck that could significantly diminish the feature space in subsequent layers, potentially resulting in the loss of valuable information.

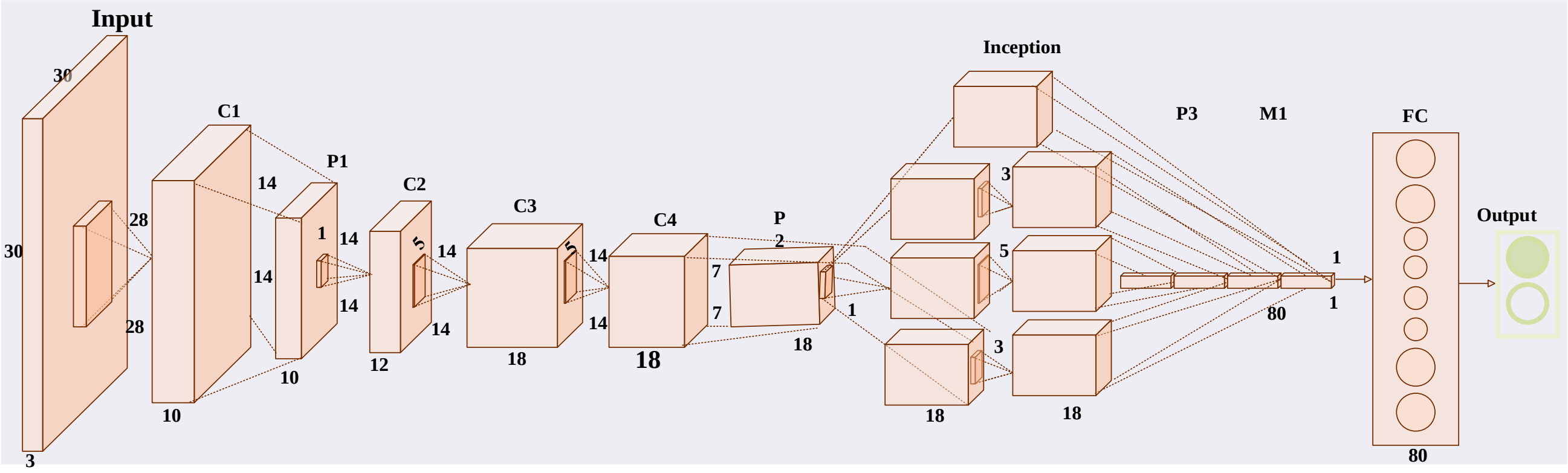

Figure 15: GoogLeNet (Inception): Introduced inception modules for efficient use of different filter sizes within the same layer.

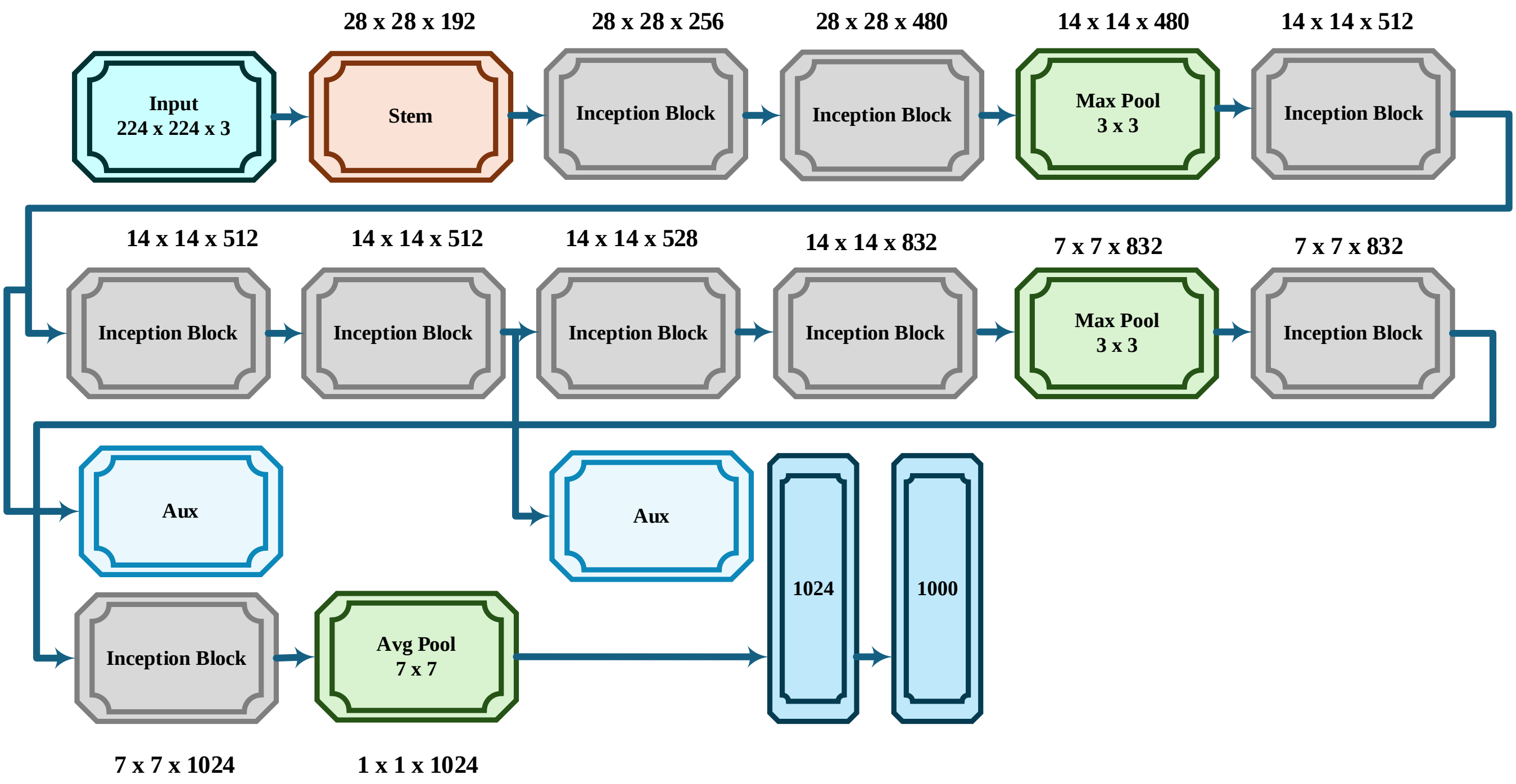

Figure 16: GoogLeNet (Inception V1) CNN Architecture.

### 4.3. Depth-based CNN

Deep CNN architectures operate on the principle that increased depth enables the learning of more complex functions. This occurs through multiple non-linear transformations within the network, resulting in the formation of intricate feature hierarchies [80]. Network depth is essential for successful supervised training.

Theoretical research has demonstrated that deep networks surpass shallower ones in representing certain classes of functions [81]. Although Csáji's 2001 theorem indicates that a single hidden layer can approximate any function, the required number of neurons becomes exponentially large, rendering it computationally impractical. Deeper networks address this issue by preserving the ability to learn complex functions while keeping computational demands manageable. Empirical support for this concept was provided by [80], who illustrated the computational efficiency of deep networks in handling complex tasks[82]. The remarkable performance of deep architectures such as VGG and Inception in the 2014 ILSVRC competition underscores the importance of depth in enhancing a network's learning capacity [35].

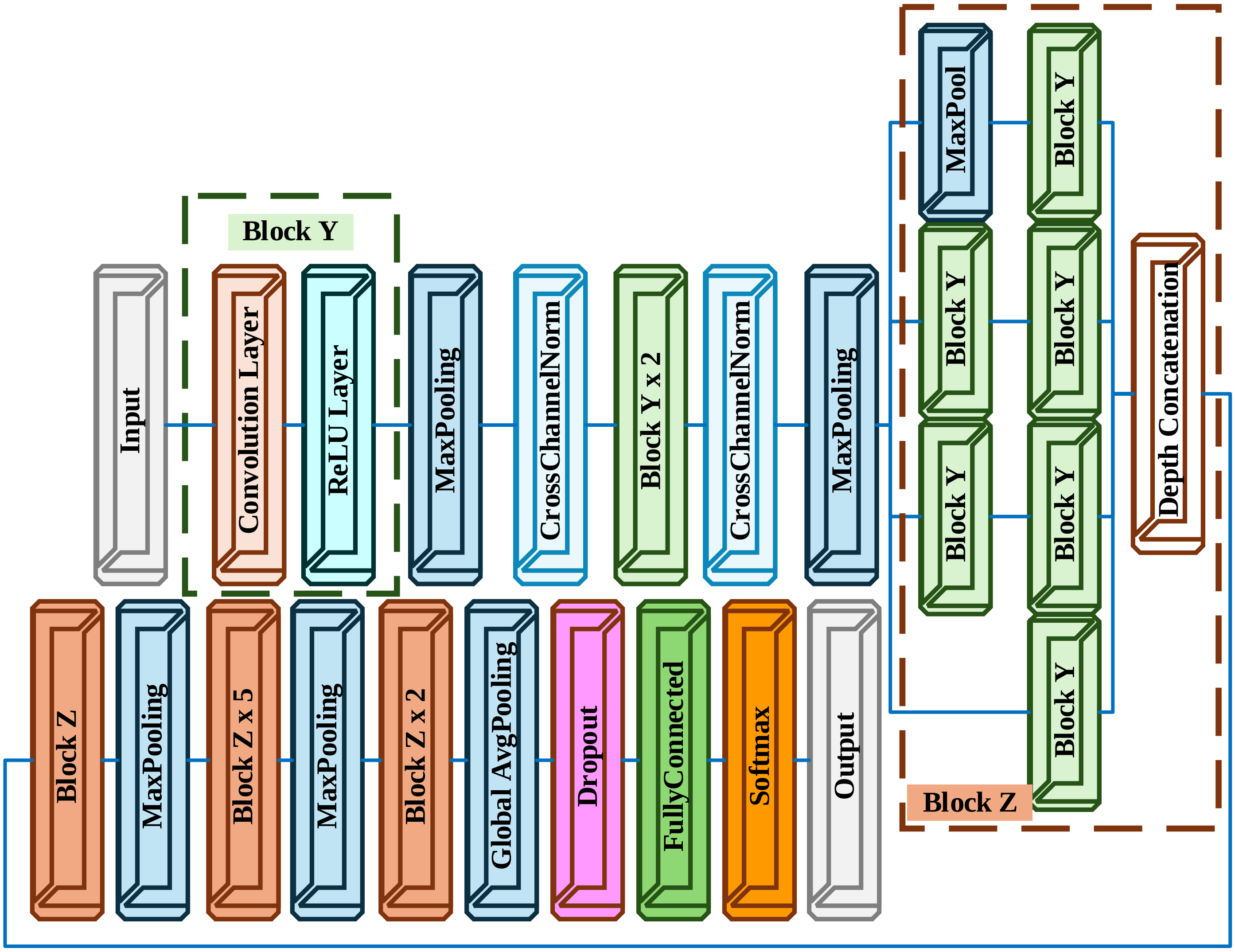

Figure 17: GoogLeNet (Inception V1) introduced inception modules, 1×1 convolutions, and global average pooling.

#### 4.3.1. Highway Networks

Building on the concept that greater depth enhances learning capacity in CNNs, Highway Networks emerged as an innovative deep CNN architecture [83], [84]. A primary challenge for deep networks involves slow training and convergence speeds [85]. Highway Networks address this issue by leveraging depth to learn more sophisticated feature representations and introducing a novel cross-layer connectivity mechanism, as depicted in Figure 18. This mechanism, part of the multi-path architecture category, facilitates faster training of deep networks.

Studies indicate that adding layers beyond a certain threshold in traditional networks can degrade performance [86]. However, Highway Networks achieve significantly faster convergence, even with extreme depths of up to 900 layers. A detailed explanation of Highway Networks' cross-layer connectivity can be found in equations (7) and (8).

$$\boldsymbol{F}_{l+1}^{k} = g_c(\boldsymbol{F}_{l}^{k}, \boldsymbol{k}_l) \cdot g_{t_g}\left(\boldsymbol{F}_{l}^{k},\ {}^{tg}\boldsymbol{k}_l\right) \cdot g_{c_g}(\boldsymbol{F}_{l}^{k},\ {}^{cg}\boldsymbol{k}_1) \qquad (7)$$

$$g_{c_g}(\boldsymbol{F}_{l}^{k},\ {}^{cg}\boldsymbol{k}_l) = 1 - g_{l_g}\left(\boldsymbol{F}_{l}^{k},\ {}^{tg}\boldsymbol{k}_l\right) \qquad (8)$$

In equation (7) & (8), $g_c(\mathbf{F}_l^k, \mathbf{k}_l)$ represents the operation of the $l^{th}$ hidden layer. The gates $t_g$ and $c_g$ regulate the information flow between layers. When the $t_g$ gate is open ($t_g = 1$), the transformed input advances to the next layer. Conversely, when $t_g = 0$, the $c_g$ gate establishes a direct information pathway, allowing the input $\mathbf{F}_l^k$ from the $l^{th}$ layer to pass to the $l + 1$ layer without any transformation.

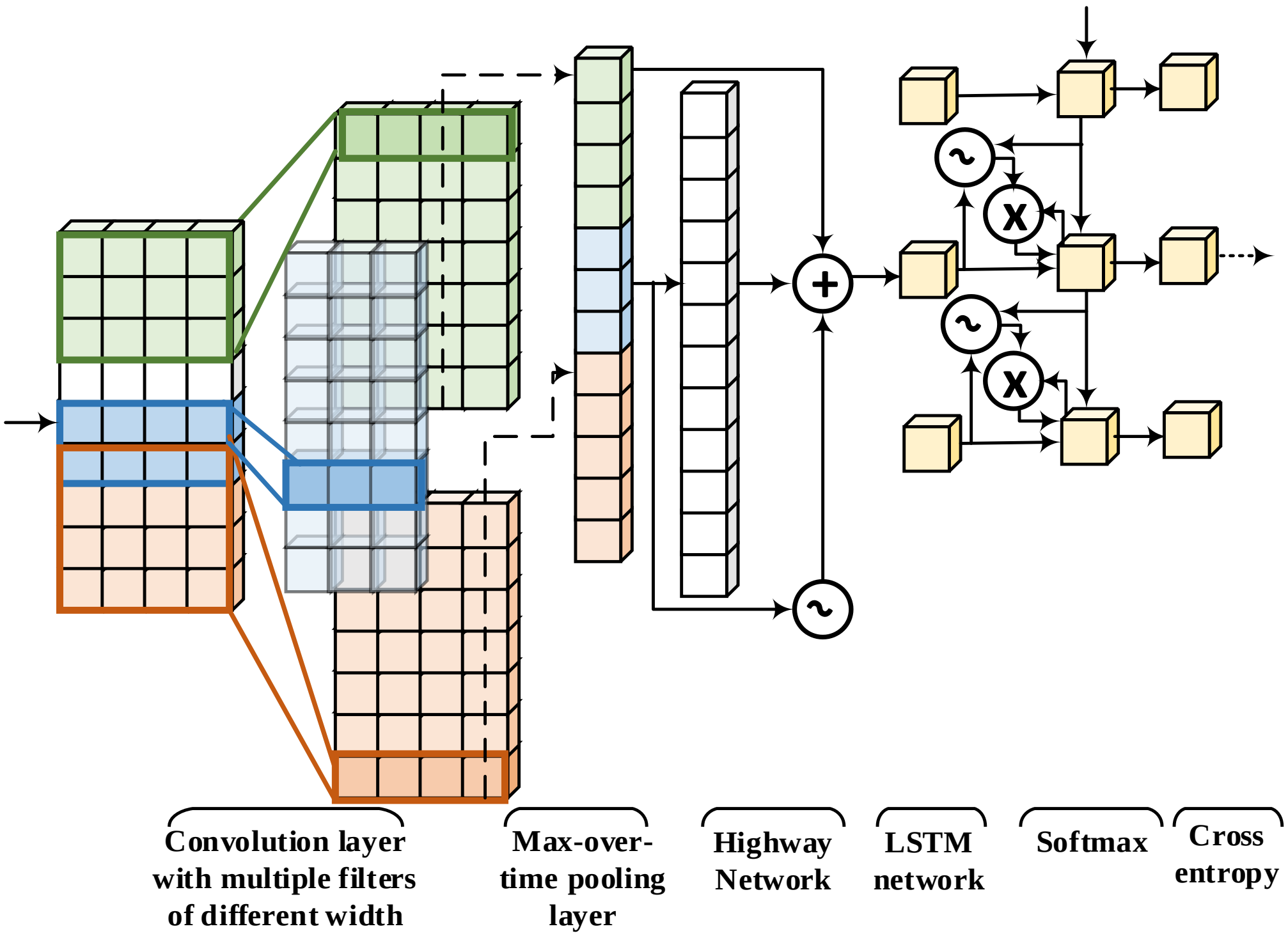

Figure 18: Highway Network: Gating Mechanisms for Neural Networks

#### 4.3.2. ResNet

ResNet, introduced in [66], marks a significant milestone in DL. It transformed CNN architecture by introducing the concept of residual learning. This method entails adding direct connections between layers within the network, facilitating easier information flow, and addressing the vanishing gradient problem that often impedes training in deep networks. Additionally, ResNet employs an efficient training methodology tailored for deep networks. Like Highway Networks, ResNet belongs to the category of multi-path CNNs due to the presence of these shortcut connections.

ResNet's pioneering contribution was the development of a 152-layer deep CNN architecture. This significant achievement led to its triumph in the prestigious 2015 ILSVRC competition and expanded the limits of CNN's depth. Remarkably, despite being 20 times deeper than AlexNet and 8 times deeper than VGG, ResNet maintained lower computational complexity compared to earlier models. Figure 19, Figure 20, Figure 21, Figure 22, Figure 23, Figure 24, and Figure 25 illustrate the structure of ResNet's residual block, the crucial component that facilitates this efficiency [13], [35]. Empirical research conducted by [37] demonstrated that ResNet, with configurations of 50/101/152 layers, outperformed the 34-layer plain Net in image classification tasks. Furthermore, ResNet displayed a notable 28% improvement in performance on the renowned COCO image recognition benchmark dataset [73]. These accomplishments underscored the pivotal role of representational depth in a myriad (numerous) visual recognition tasks, thereby solidifying ResNet's profound impact in the field as represented in equations (9-11).

$$\boldsymbol{F}_{m+1}^{k'} = g_c\left(\boldsymbol{F}_{l\to m}^{k}, \boldsymbol{k}_{l\to m}\right) + \boldsymbol{F}_l^k \; m \geq l \tag{9}$$

$$\boldsymbol{F}_{m+1}^{k} = g_a\left(\boldsymbol{F}_{m+1}^{k'}\right) \tag{10}$$

$$g_c\left(\boldsymbol{F}_{l\to m}^{k}, \boldsymbol{k}_{l\to m}\right) = \boldsymbol{F}_{m+1}^{k'} - \boldsymbol{F}_l^k \tag{11}$$

In this context, $g_c\left(\mathbf{F}_{l\to m}^{k}, \mathbf{k}_{l\to m}\right)$ represents a transformed signal, while $\mathbf{F}_l^k$ is the input to the $l^{th}$ layer. $\mathbf{k}_{l\to m}$ indicates the $k^{th}$ processing unit (kernel), and $l \to m$ signifies that the residual block may include one or multiple hidden layers. The initial input $\mathbf{F}_l^k$ is combined with the transformed signal $\left(g_c\left(\mathbf{F}_{l\to m}^{k}, \mathbf{k}_{l\to m}\right)\right)$ via a bypass pathway, resulting in an accumulated output $\mathbf{F}_{m+1}^{k'}$, which is then passed to the subsequent layer after applying the activation function $g_a(.)$. Additionally, $\mathbf{F}_{m+1}^{k'} - \mathbf{F}_l^k$ provides residual information used for reference-based weight optimization. A key feature of ResNet is its reference-based residual learning framework, which facilitates the optimization of residual functions and enhances accuracy with increased network depth.

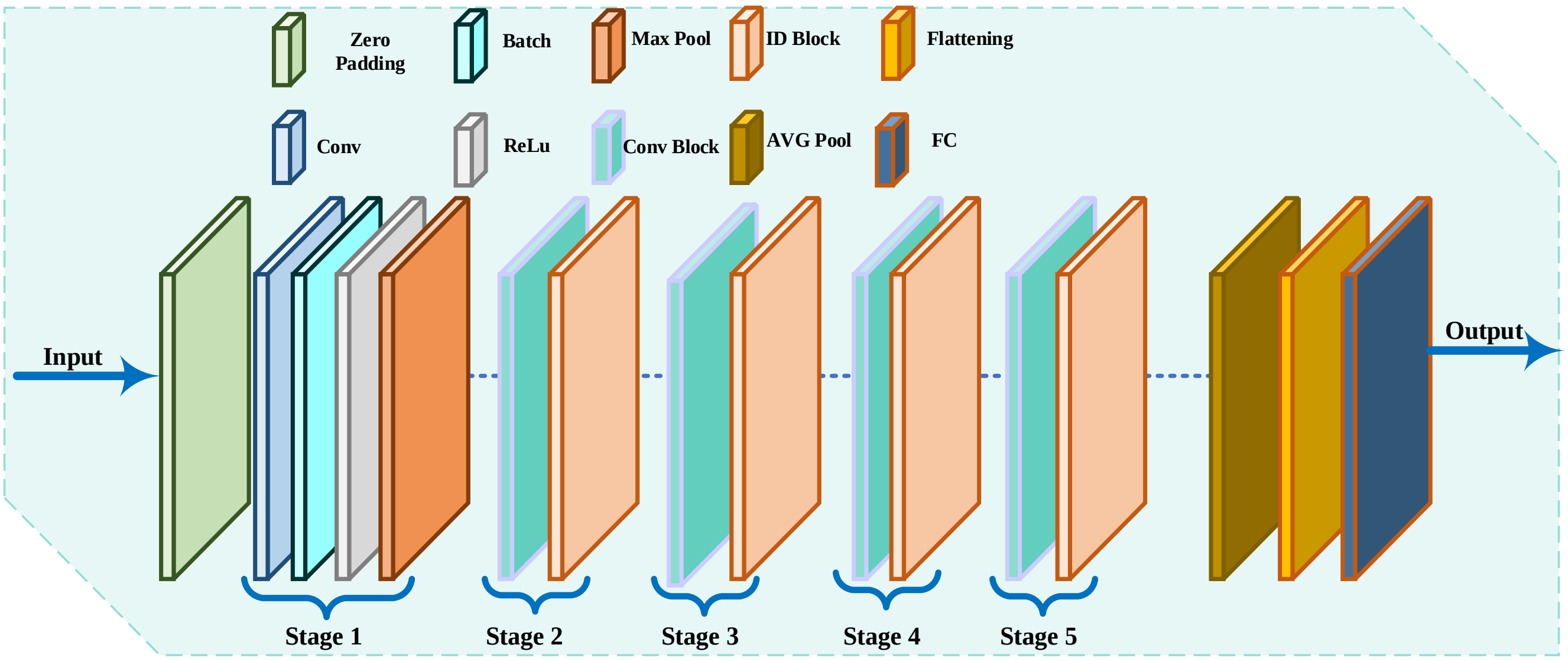

Figure 19: ResNet Architecture: Skip Connections for DL.

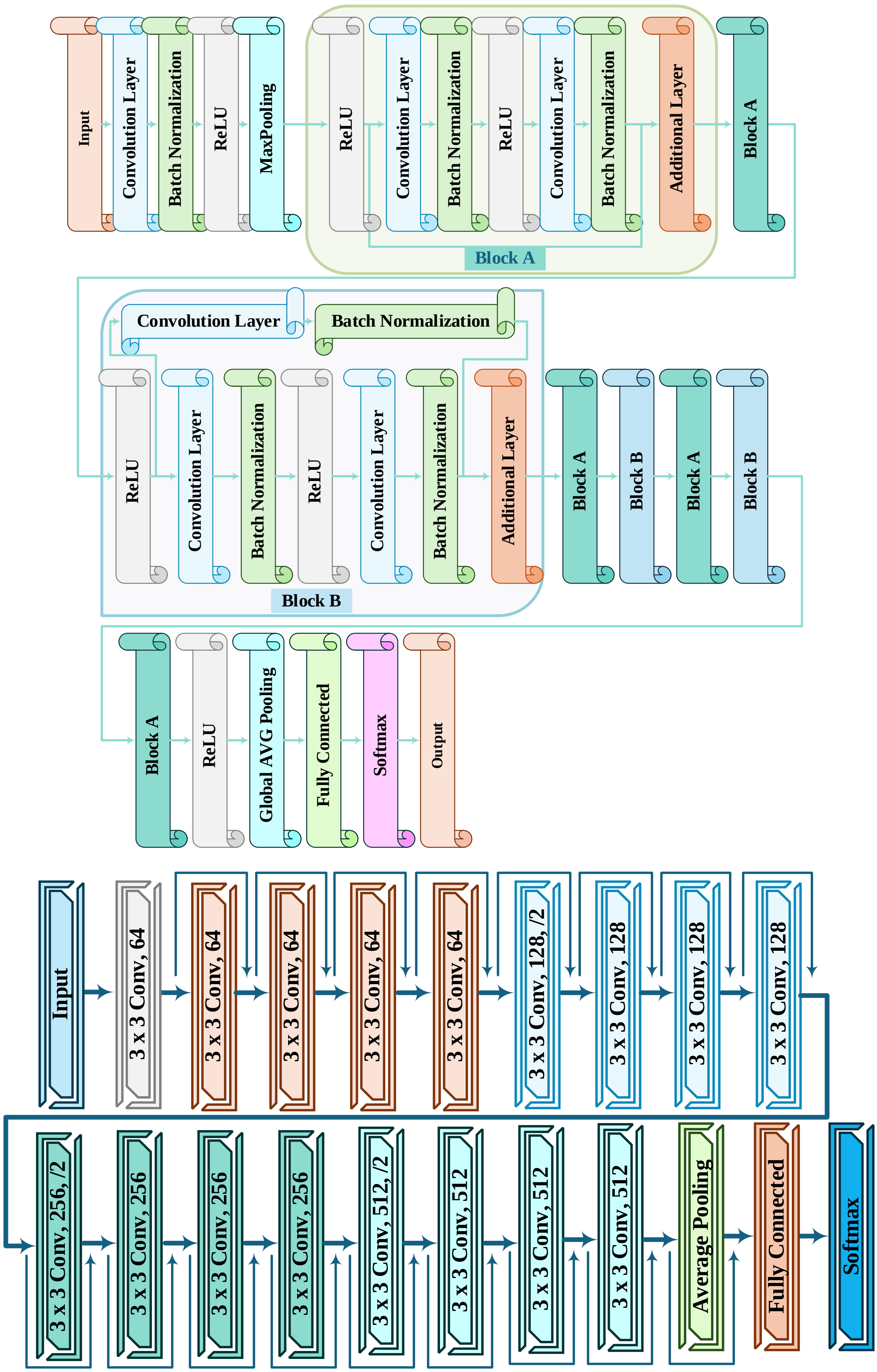

Figure 20: ResNet18 a variant of the Residual Network (ResNet) architecture, which was introduced to address the vanishing gradient problem in deep neural networks.

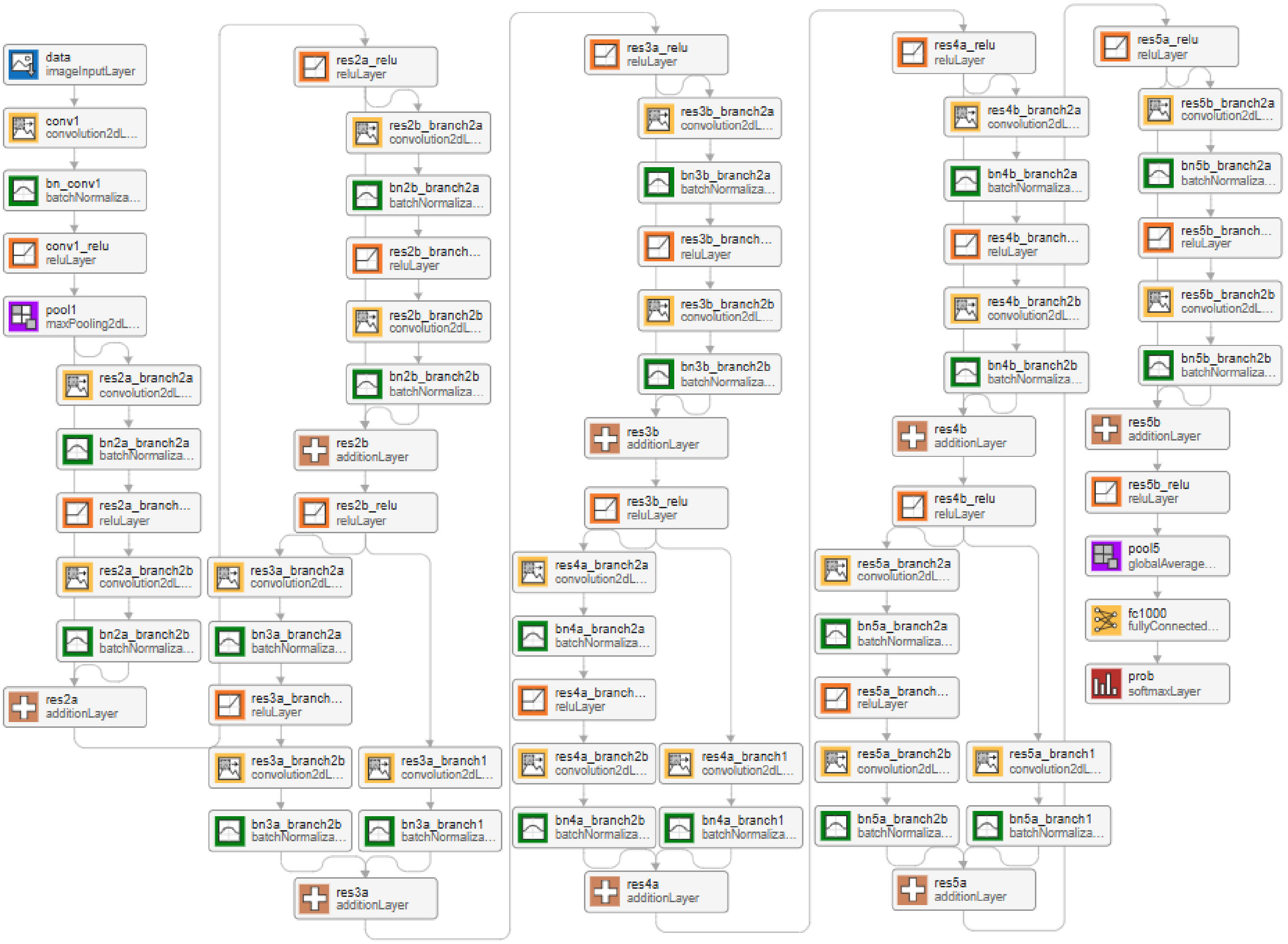

Figure 21: ResNet-18: Neural Networks with Identity Shortcut Connections.

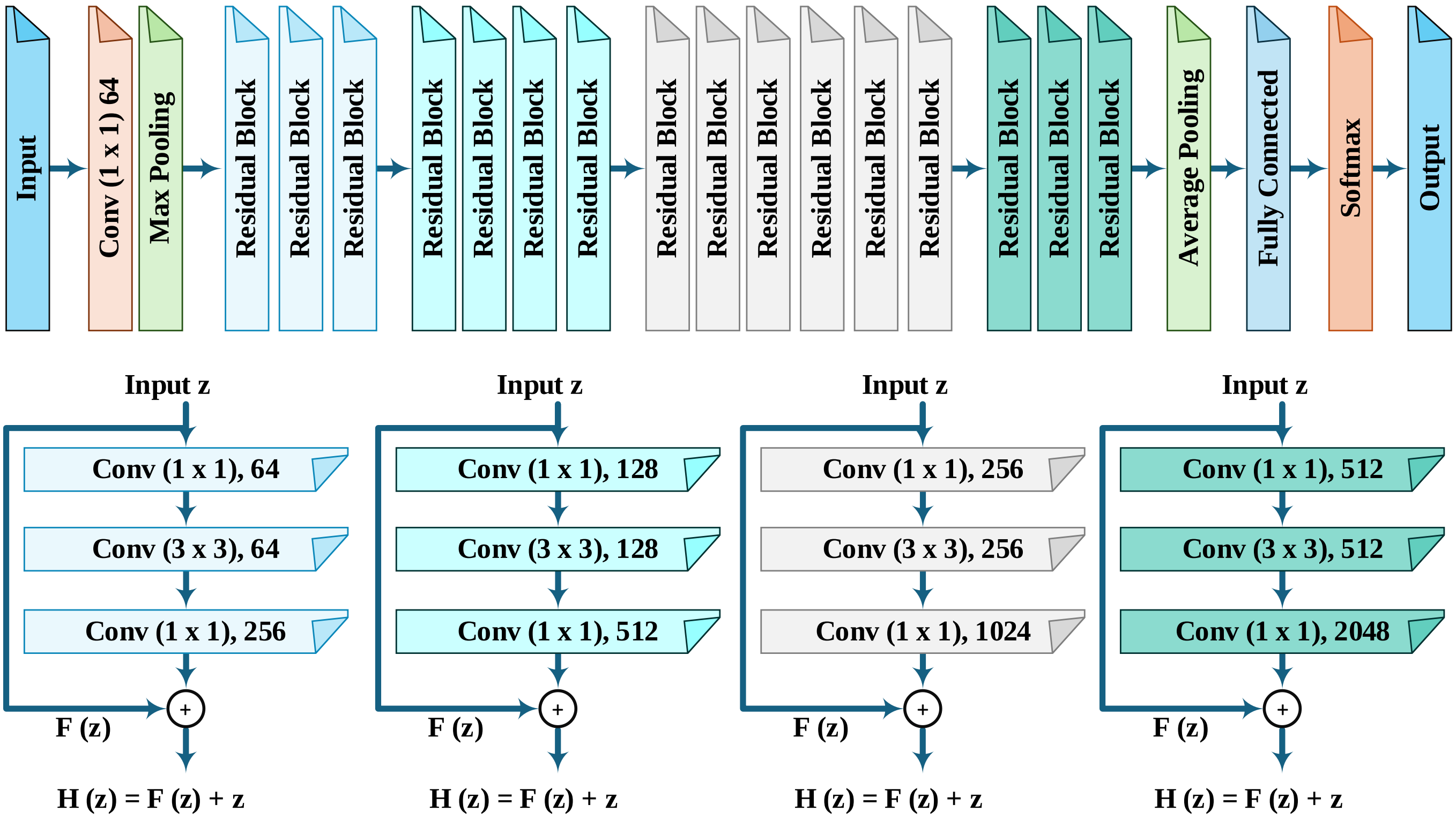

Figure 22: ResNet50, deep CNN architecture that was developed by Microsoft Research in 2015.

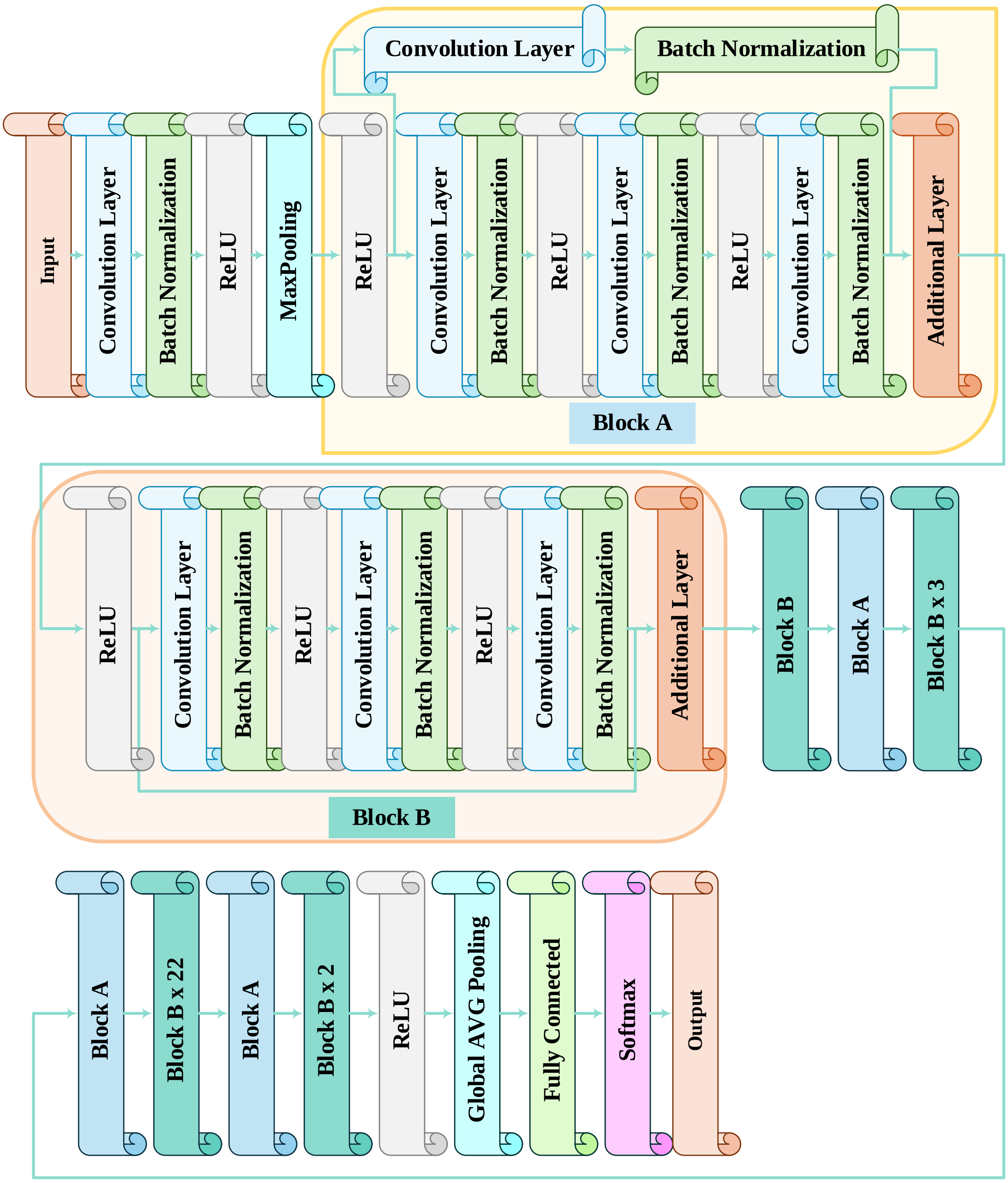

Figure 23: ResNet101: More powerful ResNet variant.

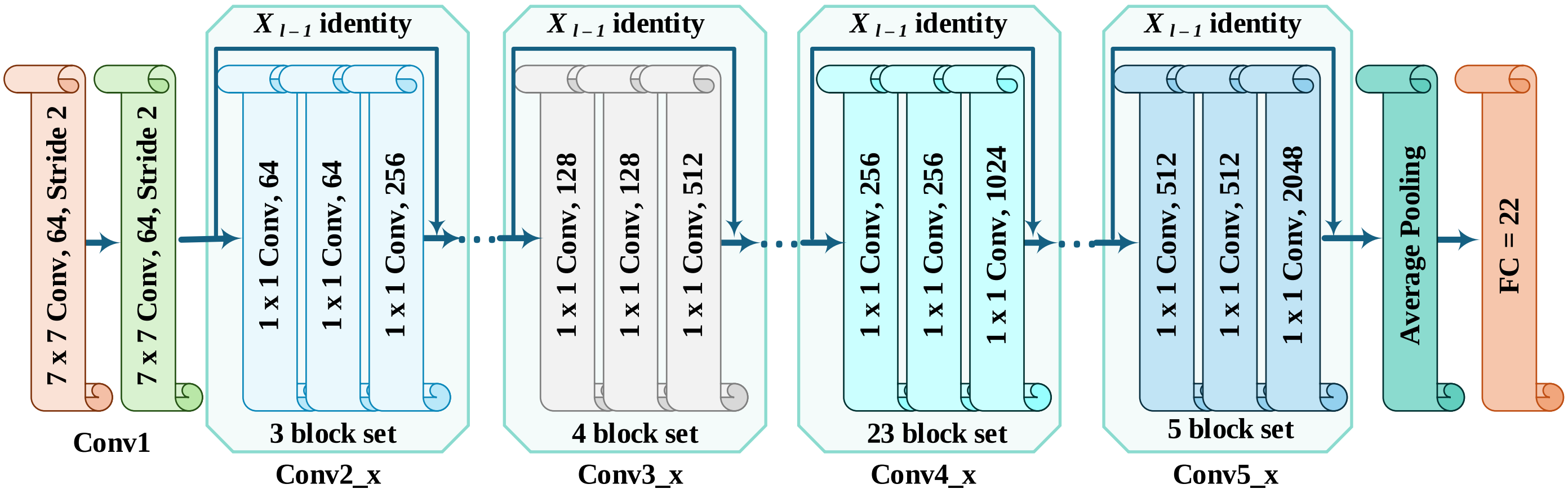

Figure 24: ResNet-101 is a CNN network with 101 layers.

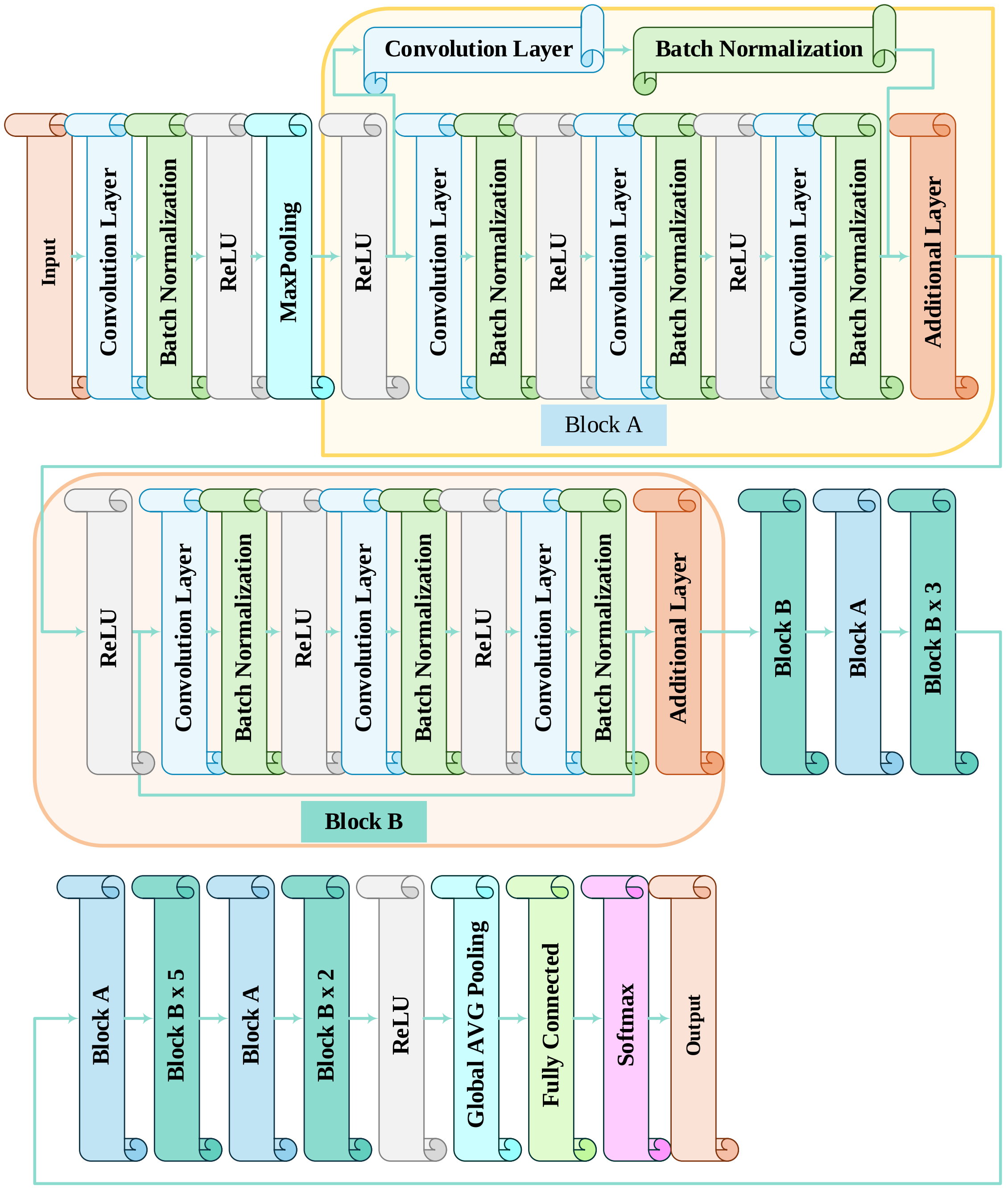

Figure 25: ResNet50: Groundbreaking architecture with residual connections.

#### 4.3.3. Inception-V3, V4 and Inception-ResNet

Inception-V3, V4, and Inception-ResNet are advanced iterations of the Inception-V1 and V2 architectures originally introduced [23]. The central aim behind the development of Inception-V3 was to alleviate the computational burden associated with deep networks while upholding their capacity for generalization, as shown in Figure 27 and Figure 28. To achieve this, [87] implemented a series of strategic modifications. They adopted an innovative strategy by substituting larger filters (such as 5x5 and 7x7) with a combination of smaller asymmetric filters (1x7 and 1x5) and adding a 1x1 convolutional layer before the larger filters as shown in Figure 26. This 1x1 layer served as a bottleneck, decreasing dimensionality and computational complexity. The concurrent application of a $1 \times 1$ convolutional layer alongside a larger filter transformed the conventional convolution operation into something resembling cross-channel correlation [36].

In the context of Inception-V3 utilizes a sophisticated technique for feature extraction by employing 1x1

convolutions to project the input data into several lower-dimensional spaces. This enables the network to efficiently process the data using standard convolutions (3x3 or 5x5) within these reduced dimensions. Inception-ResNet enhances this approach by integrating the strengths of Inception blocks and the residual connections introduced by ResNet [23]. It replaces the filter concatenation in the original Inception blocks with residual connections. Studies indicate that this combination allows Inception-ResNet to achieve greater depth and width compared to Inception-V4 while maintaining similar generalization capacity—the ability to perform well on unseen data.

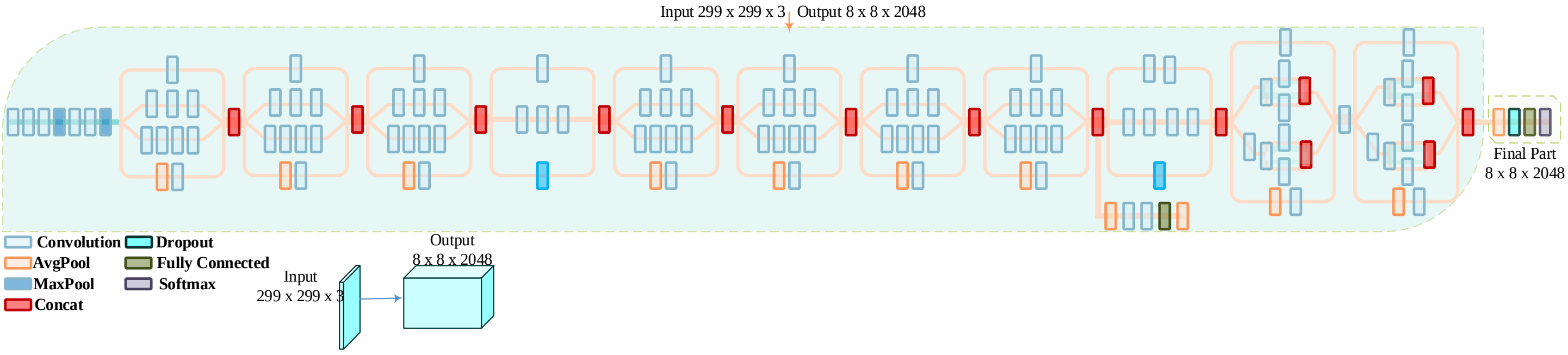

Figure 26: Inception Architecture: Parallel Convolutional Operations.

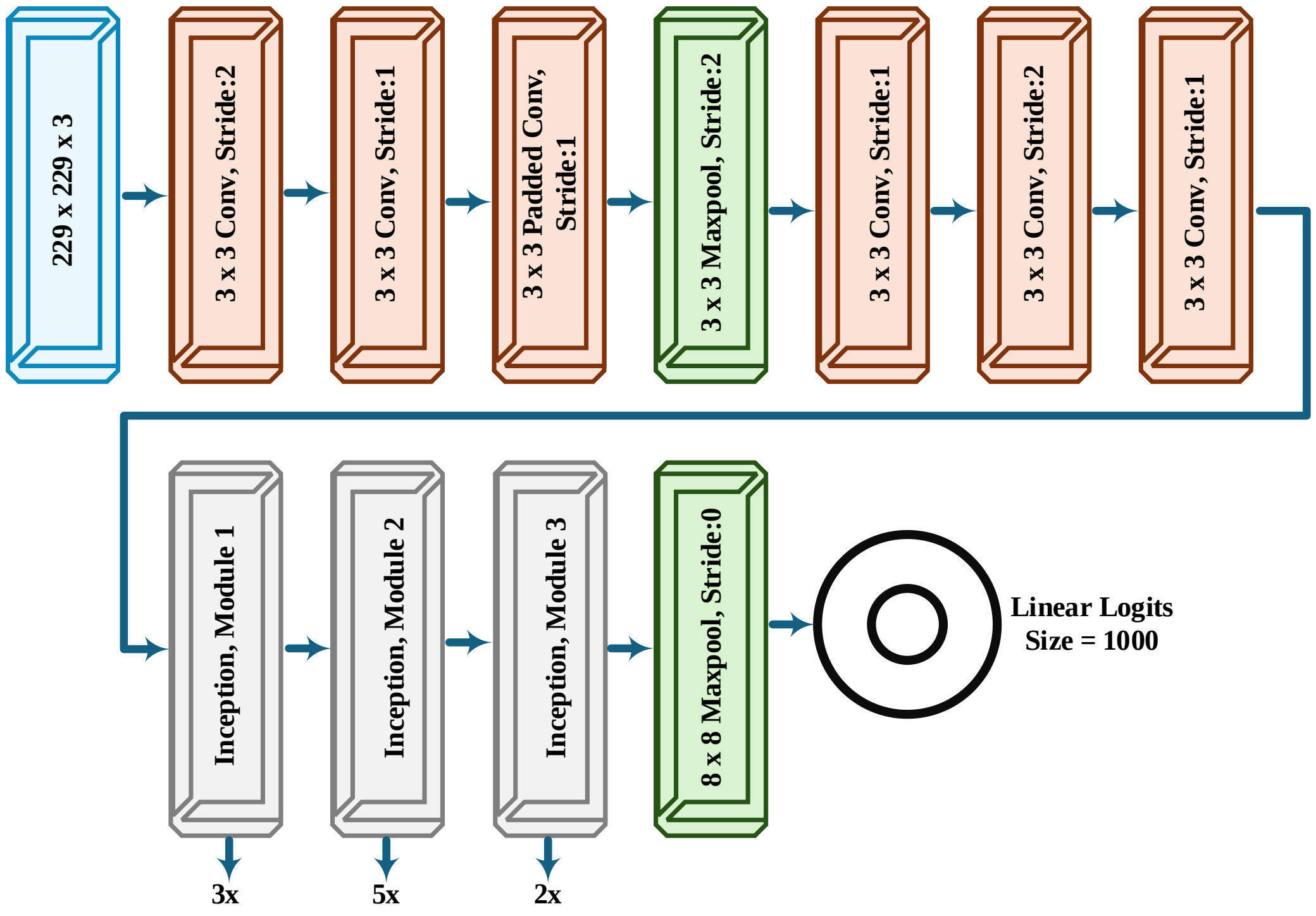

Figure 27: InceptionV3 is a CNN network with 48 deep layers and classifies into 1000 different object categories.

### 4.4. Multi-path based CNNs

Training very deep neural networks presents significant challenges in contemporary research. These networks, despite their adeptness at handling complex tasks, often encounter issues such as performance degradation and vanishing or exploding gradients, distinct from the problems associated with overfitting [88]. The vanishing gradient problem, particularly severe, not only compromises performance on unseen data-test errors but also impedes effective learning during training [88].

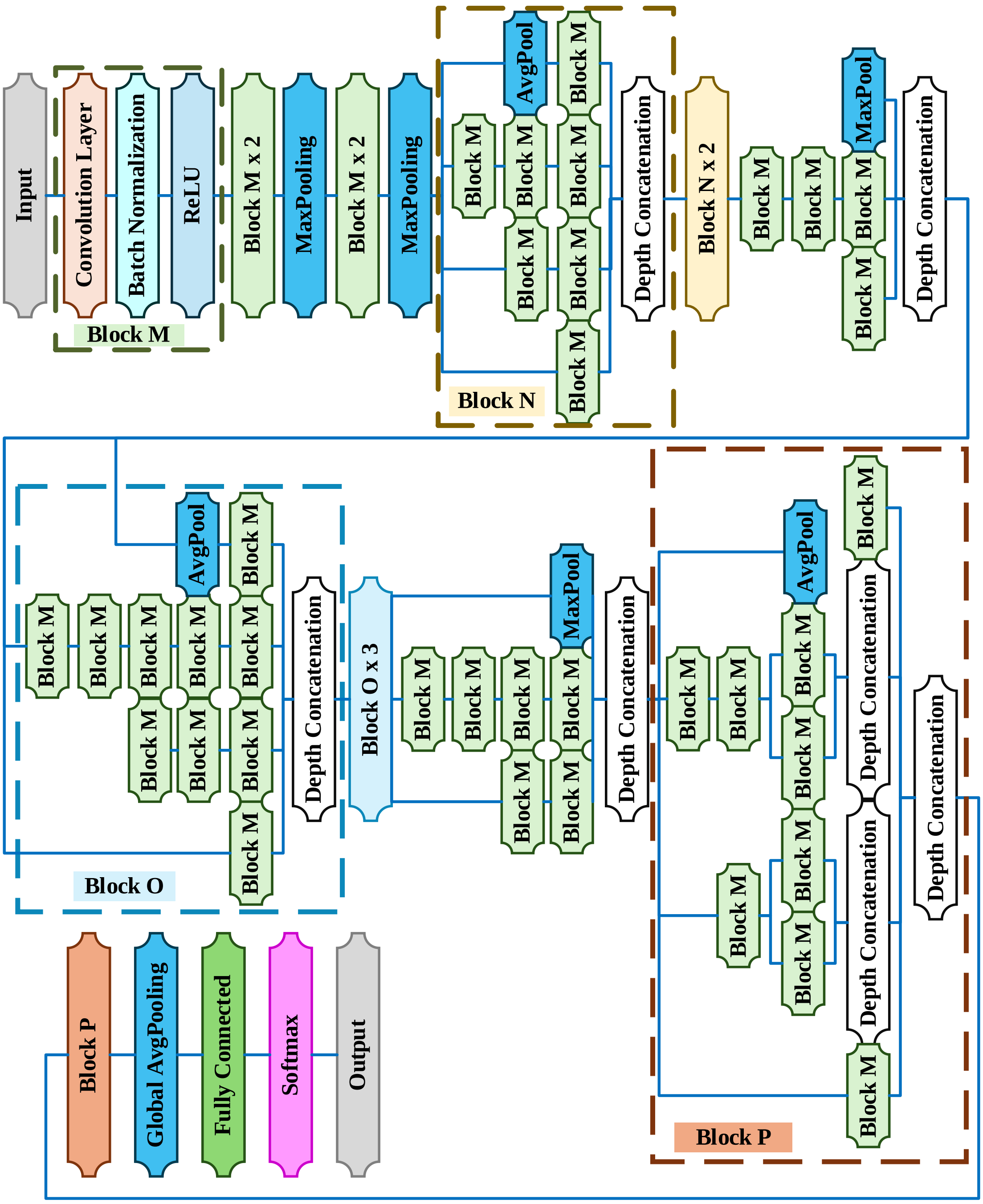

Figure 28: Inception-V3: Uses inception modules for image classification.

In addressing these, there are novel strategies under consideration: adding connections across layers [89], [90]. A very fundamental strategy consists of the use of multi-path connections. These indeed establish varied paths between layers, often bypassing intermediate layers. Such a design would allow for a better flow of specialized information within the network and hence may result in an improved overall performance.

The other strong approach is to use cross-layer connections, where the whole network is divided into small blocks with prescribed connections between these blocks. It tries to avoid problems of vanishing gradients since a better gradient flow to the previous layers in the network can be allowed. The different techniques serving this purpose, such as zero-padding, extra links-interjections-projection-based connections, dropout, and special shortcut connections, such as skip and 1×1 connections, make such cross-layer links feasible. These are a must for the enhancement of training efficiency and performance deep neural networks can reach and have enabled more powerful and effective models to solve ever-more difficult tasks in machine learning (ML) and artificial intelligence.

#### 4.4.1. DenseNet

The problem of vanishing gradients, which can hinder training in deep networks, is addressed with DenseNet [83]. While ResNet utilizes shortcut connections to maintain information flow, some layers in ResNet might have minimal impact. DenseNet overcomes this by introducing a distinct form of cross-layer connectivity. In contrast to ResNet's shortcut connections, DenseNet uses feed-forward connections where each layer directly supplies its feature maps to all subsequent layers (explained in Equations. 12 and 13) Figure 29, Figure 30, and Figure 31. This method ensures that every preceding layer contributes to each subsequent layer, potentially alleviating the vanishing gradient problem and enhancing information flow throughout the network.

$$\boldsymbol{F}_2^k = g_c(l_c\,, \boldsymbol{k}_1) \tag{12}$$

$$\boldsymbol{F}_l^k = g_k(\boldsymbol{F}_l^k, \ldots, \boldsymbol{F}_{l-1}^k) \tag{13}$$

Here, $\mathbf{F}_2^k$ and $\mathbf{F}_l^k$ represent the resulting feature maps of the 1st and $l-1^{th}$ transformation layers, and $\mathrm{g}_k(.)$ is a function that enables cross-layer connectivity by concatenating information from preceding layers before assigning it to the new transformation layer $l$. DenseNet distinguishes itself from traditional CNNs by establishing direct connections between all preceding layers and each subsequent layer $\frac{l(l+1)}{2}$, contrasting with typical CNNs where connections exist only between a layer and its immediate predecessor. This dense connectivity promotes depthwise convolution, allowing information from earlier layers to influence later ones. Another notable feature of DenseNet is its use of feature concatenation rather than addition. This approach explicitly separates newly learned information from each layer and preserved information from previous layers. While advantageous, it increases parameter complexity as the number of feature maps expands. Furthermore, to mitigate potential issues with information flow and overfitting, DenseNet provides each layer direct access to gradients from the loss function. This direct gradient access serves as a form of regularization, particularly beneficial for tasks with limited training data.

### 4.5. Width-based multi-connection CNNs

Between 2012 and 2015, neural network research predominantly concentrated on unlocking the potential of network depth and exploring the effectiveness of multi-path connections as a means to govern network behavior [83]. However, a pivotal revelation surfaced in 2019 through the work of [91], highlighting the paramount importance of network width. They revealed that a multilayer perceptron could outperform a single-layer perceptron in its ability to map complex functions, achieved by harnessing multiple processing units within a layer in parallel. This insight firmly establishes network width as a fundamental parameter that shapes learning principles in conjunction with network depth.

Recent research underscores the growing importance of network width, particularly in architectures utilizing the ReLU activation function [92]. These studies indicate that ReLU-based networks require adequate width to maintain their capability to learn any function (universal approximation property), especially with increasing depth. Moreover, studies have revealed limitations in approximating certain continuous functions even with very deep networks if the network's width (number of filters) does not exceed the input data's dimension [82]. While increasing the number of layers (depth) allows CNNs to acquire diverse feature representations, it does not guarantee enhanced learning capacity. Deep architectures face the challenge that some layers may struggle to capture meaningful features. Consequently, researchers have moved beyond simply constructing deeper and narrower networks, instead exploring thinner designs (fewer layers) but wider (more filters per layer).

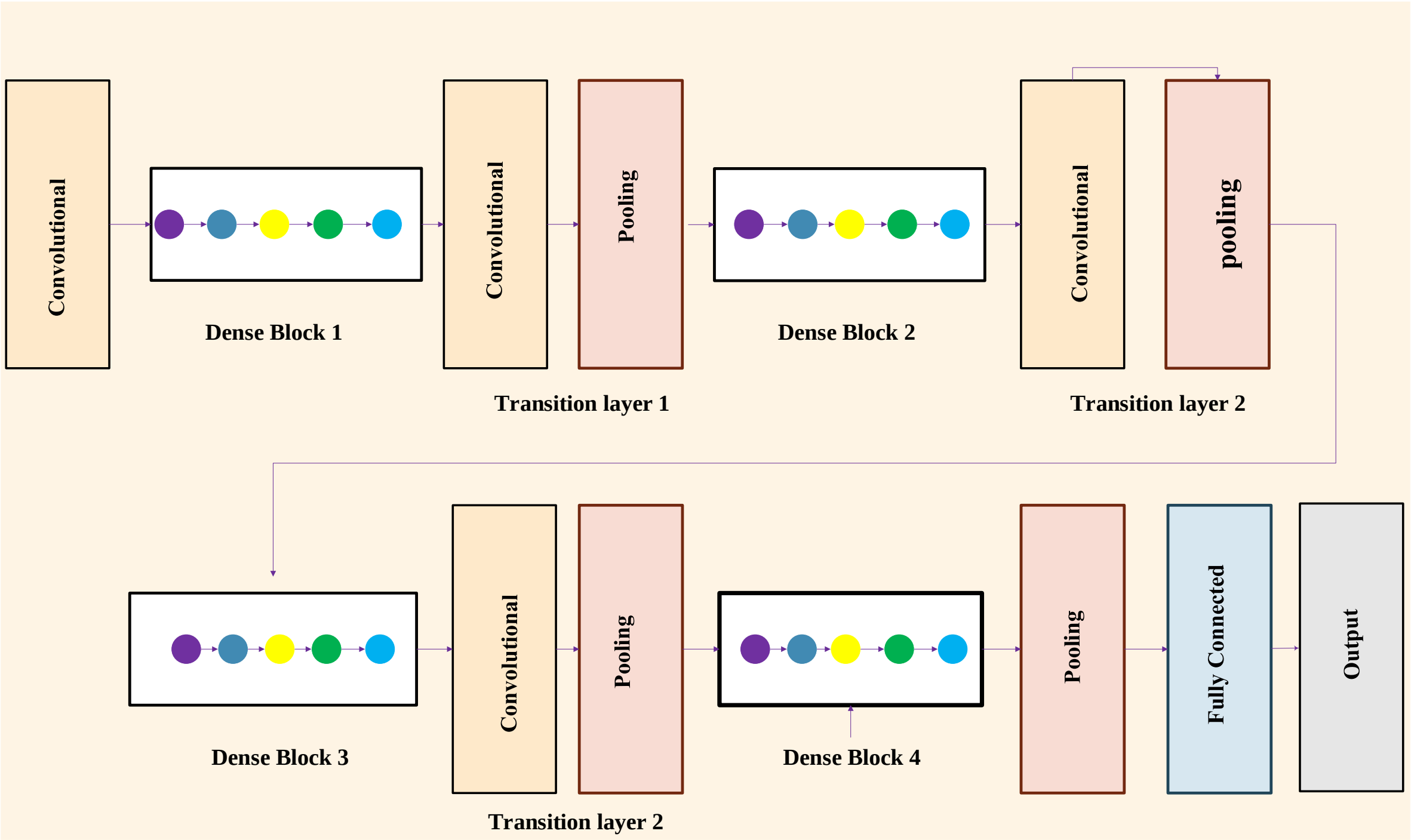

Figure 29: DenseNet is a CNN architecture that uses dense connectivity between layers to boost the performance of networks.

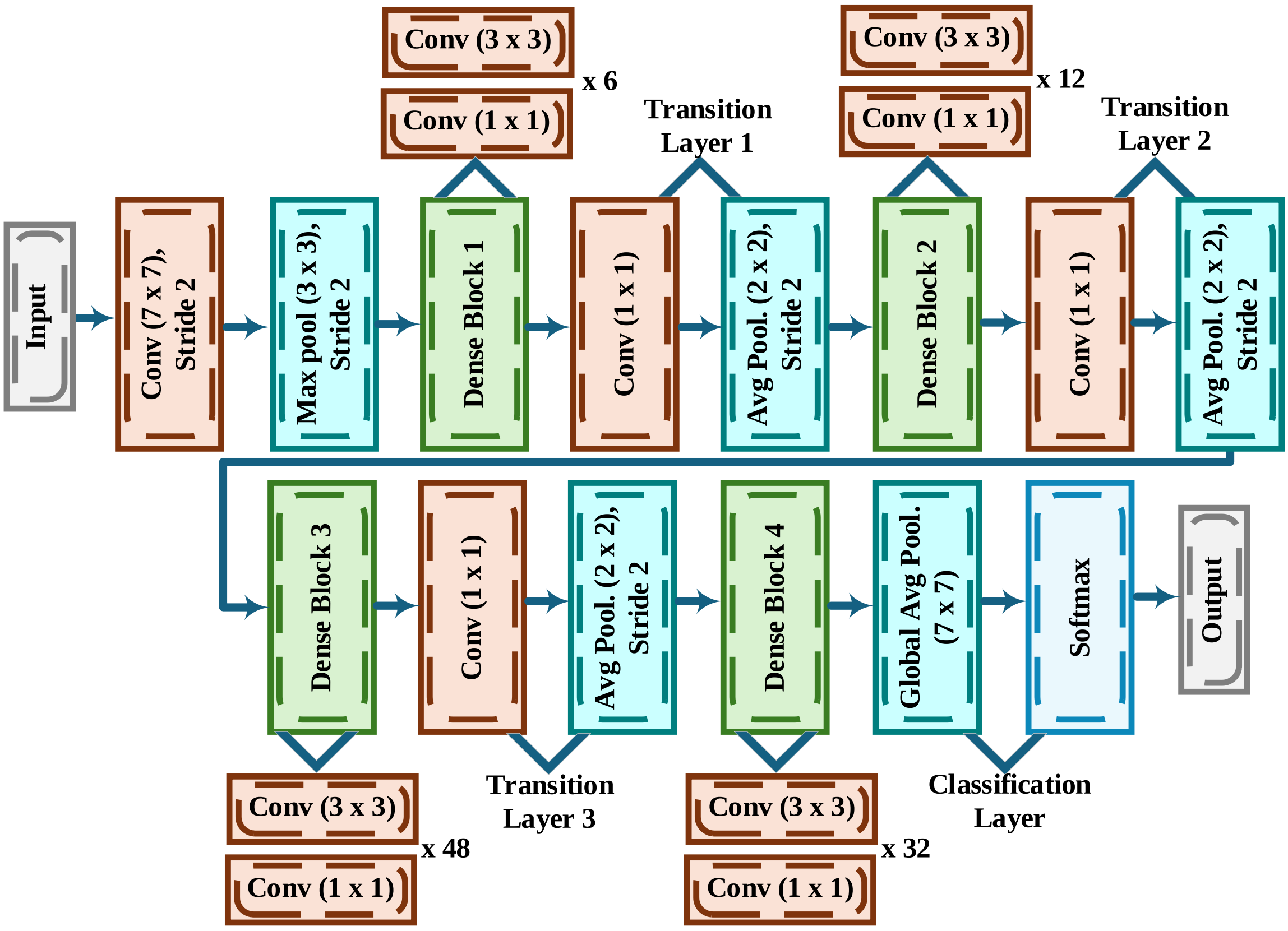

Figure 30: DenseNet201 Layered Architecture.

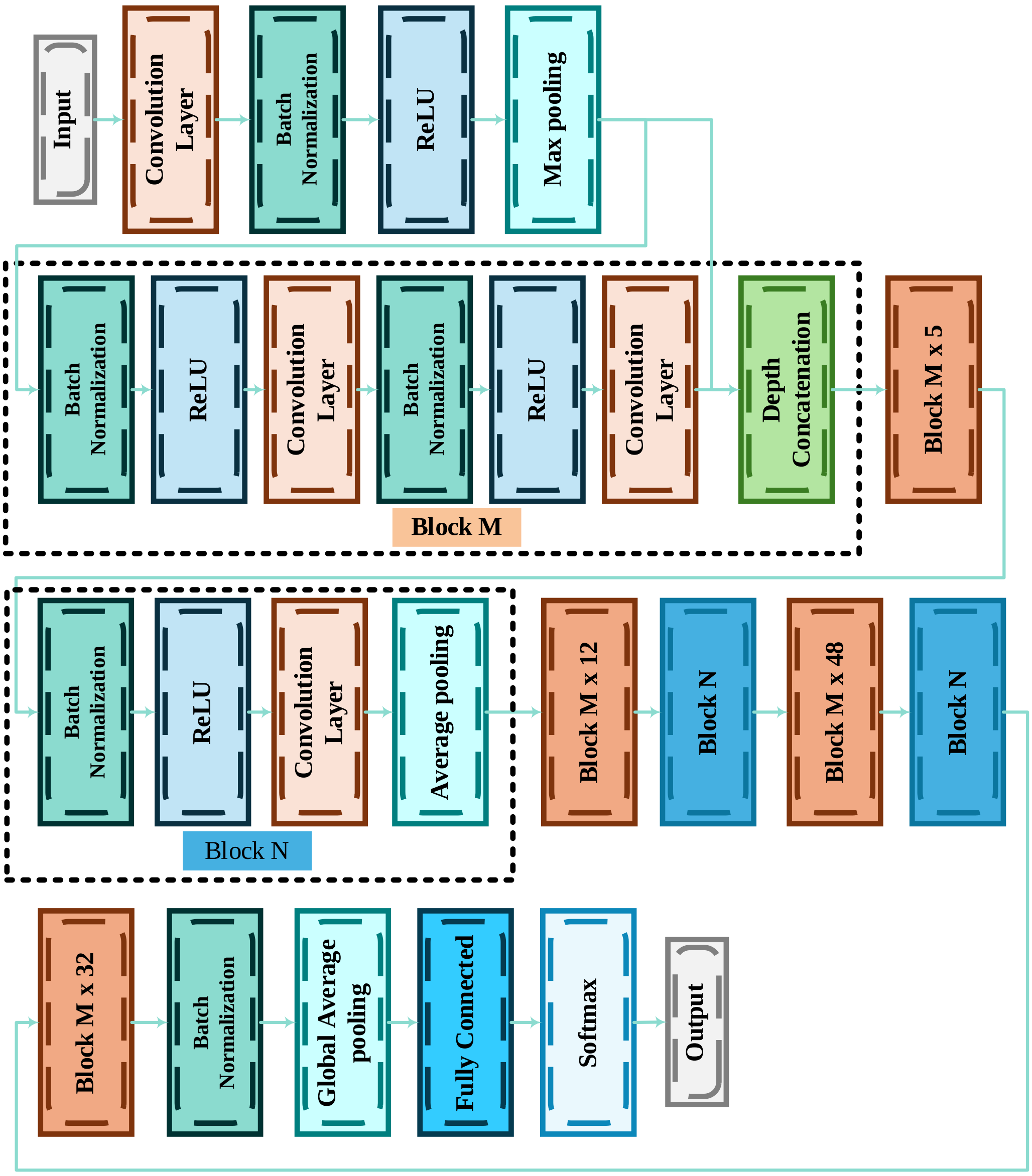

Figure 31: DenseNet201: Dense connectivity for feature reuse.

### 4.5.1. Wide ResNet

Deep residual networks, while potent, can exhibit a phenomenon known as feature reuse, where certain parts of the network may contribute minimally to the learning process. Research indicates that the primary learning efficacy stems from the residual units themselves, with depth playing a secondary role [83]. Wide ResNet addresses this issue by prioritizing network width over depth. It expands upon the standard ResNet framework by introducing a parameter "k" that controls the network's width (number of filters per layer). Studies have demonstrated that increasing the width of network layers can be more effective for improving performance compared to simply augmenting the network depth [93]. Additionally, the residual connections in Inception ResNet facilitate faster convergence during training as illustrated in Figure 32 and Figure 33. This approach diverges from traditional deep networks, which emphasize increasing depth exclusively [38]. While deep networks offer advantages, they can be slow to train and vulnerable to issues like vanishing gradients. Wide ResNet mitigates feature reuse concerns without encountering these drawbacks.

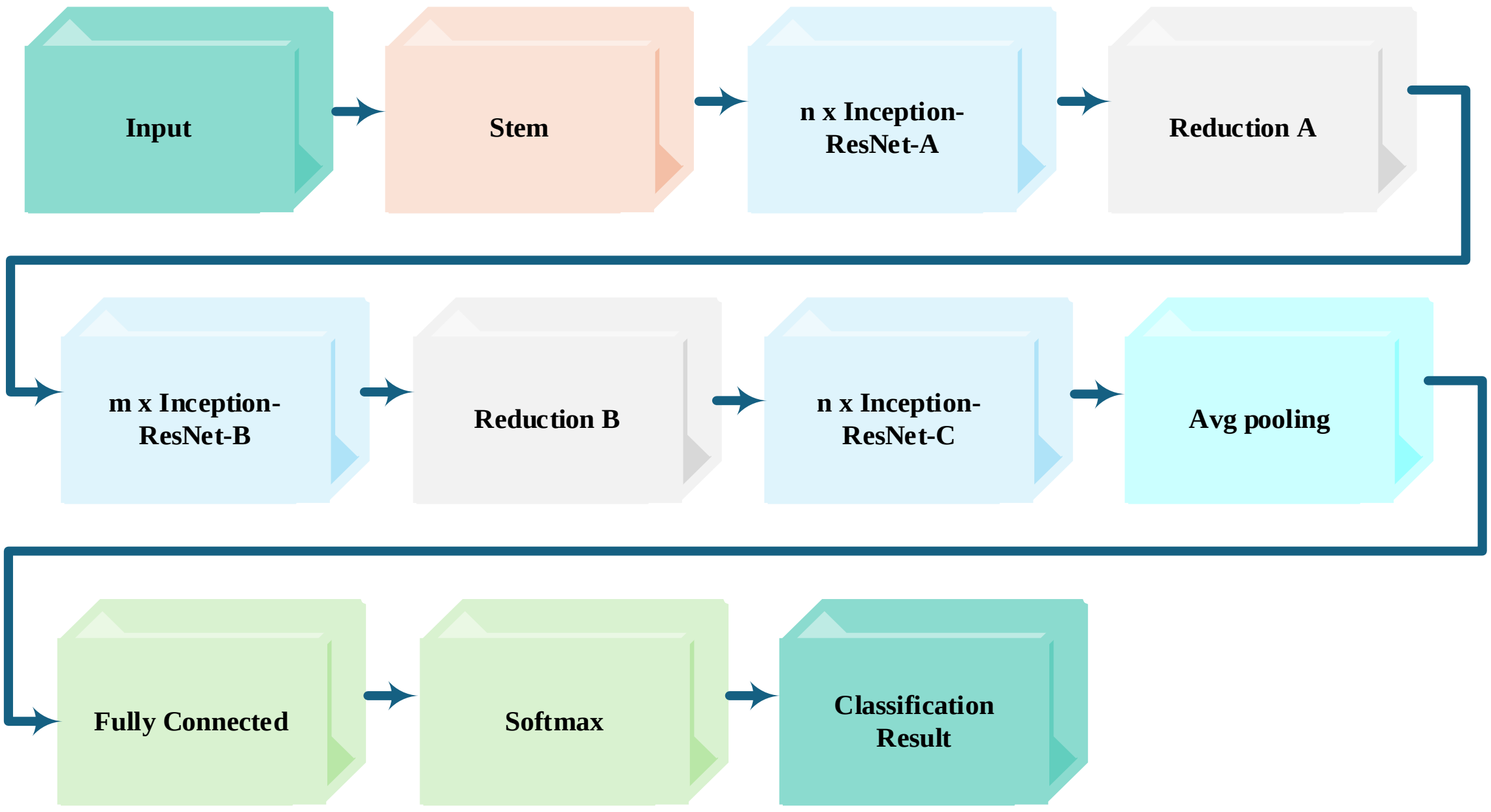

Figure 32: Inception-ResNet-v2 CNN architecture which integrates residual connections to improve its performance.

### 4.5.2. Xception

Xception represents a notable advancement in the Inception architecture by leveraging depthwise separable convolutions, a technique introduced by [94]. This method provides a more efficient approach to feature extraction from images. Xception enhances the original Inception block by employing depthwise separable convolutions consistently throughout the network. Instead of utilizing various filter sizes (1x1, 5x5, 3x3), Xception uses a standardized 3x3 depthwise convolution for spatial filtering, followed by a 1x1 pointwise convolution to reduce the number of channels. This approach maintains efficiency while effectively capturing informative features, as illustrated in Figure 34, Figure 35, and Figure 36. Xception enhances computational efficiency with depthwise separable convolutions by splitting the process into two steps which are depthwise convolution and pointwise convolution. The depthwise convolution processes each input channel with a separate filter, capturing spatial information individually. The pointwise convolution then establishes cross-channel correlations, enhancing the data representation. This method streamlines computation without reducing the total number of parameters, ultimately boosting Xception's learning capacity and performance.

Xception achieves substantial computational efficiency by separating spatial information processing from feature extraction within the convolution operation (detailed in equations 14 and 15). Initially, it uses 1x1 convolutions to project the data into lower-dimensional spaces. Subsequently, it applies multiple 3x3 depthwise separable convolutions to efficiently capture spatial information, with the number of these convolutions controlled by a parameter 'n'.

$$f_{l+1}^{k}(p,q) = \sum_{x,y} f_l^k(x,y).\, e_l^k(u,v) \tag{14}$$

$$\boldsymbol{F}_{l+2}^{k} = g_c(\boldsymbol{F}_{l+1}^{k}, k_{l+1}) \tag{15}$$

Equation (14) explains the specifics of depthwise separable convolutions used in Xception. The term $\mathbf{k}_l$ represents the $k^{th}$ kernel (filter) in the $l^{th}$ layer, which has a depth of one, indicating that it processes information from only one channel of the input feature map. This kernel conducts a spatial convolution across the $k^{th}$ feature map, denoted by $\mathbf{F}_l^k$. The spatial location within this feature map is indicated by $(x, y)$, while $(u, v)$ represents the spatial location within the kernel itself. A crucial aspect of depthwise separable convolutions is the alignment between the number of kernels K and the number of input feature maps. Unlike traditional convolutions, Xception's depthwise separable convolutions ensure that K directly corresponds to the number of input feature maps, since each kernel operates on a single channel. The equation also introduces $f_l^k$, which represents the $k^{th}$ kernel in the 1x1 pointwise convolution step for the $l^{th}$ layer. This kernel performs a convolution across the entire depth of the output feature maps generated by the preceding depthwise convolution $k_{l+1}$. $[\mathbf{F}^1_{l+1}, \ldots, \mathbf{F}^k_{l+1}, \ldots, \mathbf{F}^k_{l+1}]$ of the $l^{th}$ layer these output maps then serve as the input for the $l+1^{th}$ layer in the Xception architecture.

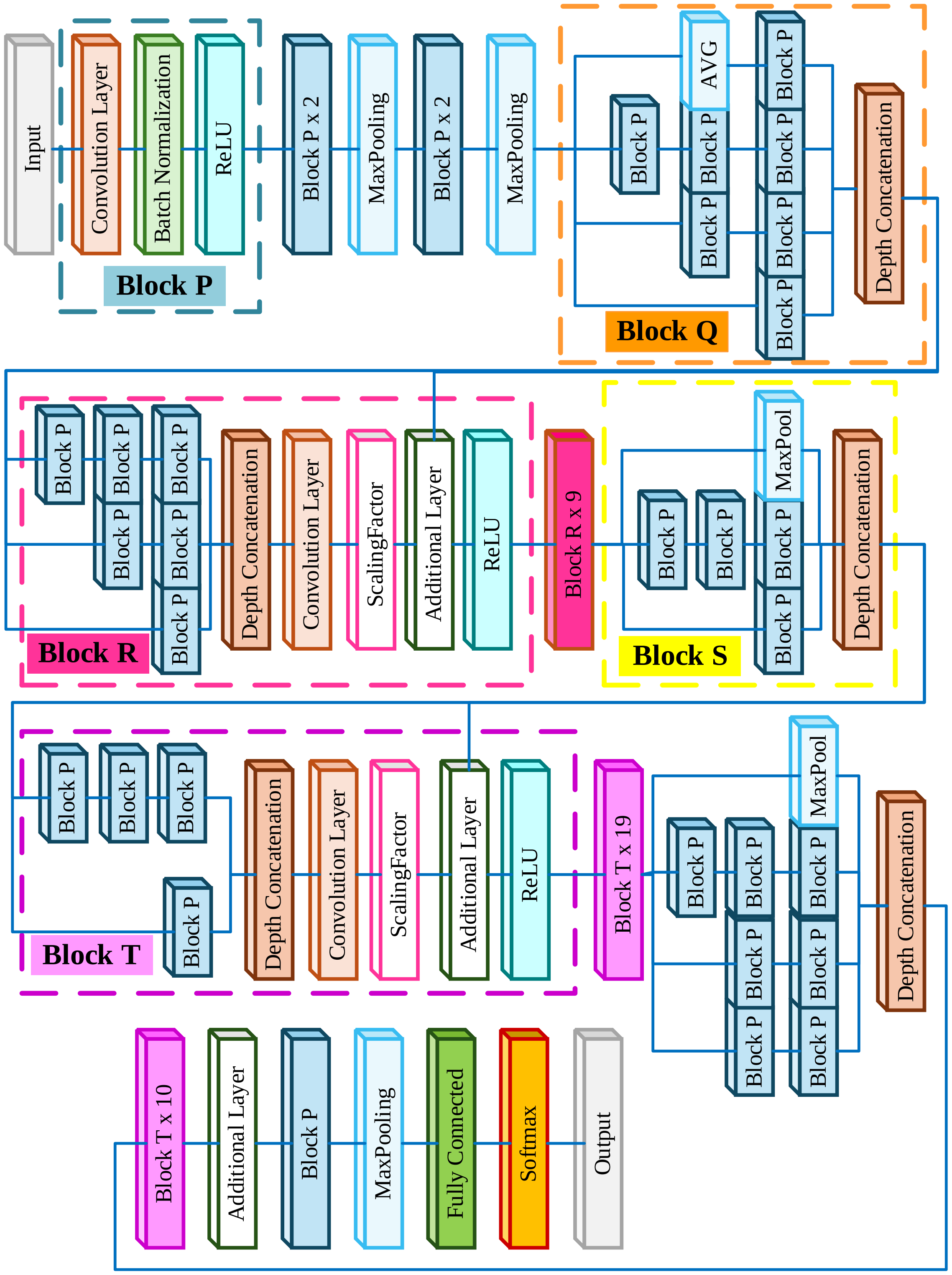

Figure 33: Inception-ResNetV2: Combines Inception and ResNet.

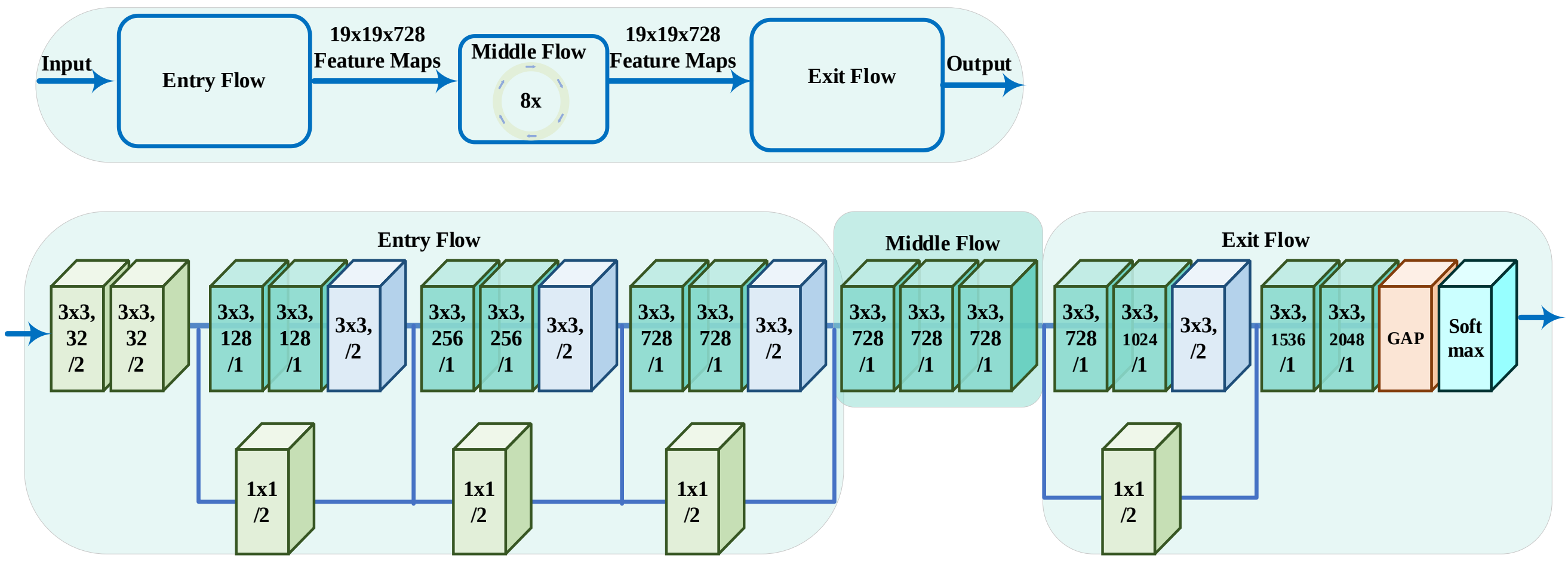

Figure 34: Xception architecture: Depthwise separable convolutions enhance performance.

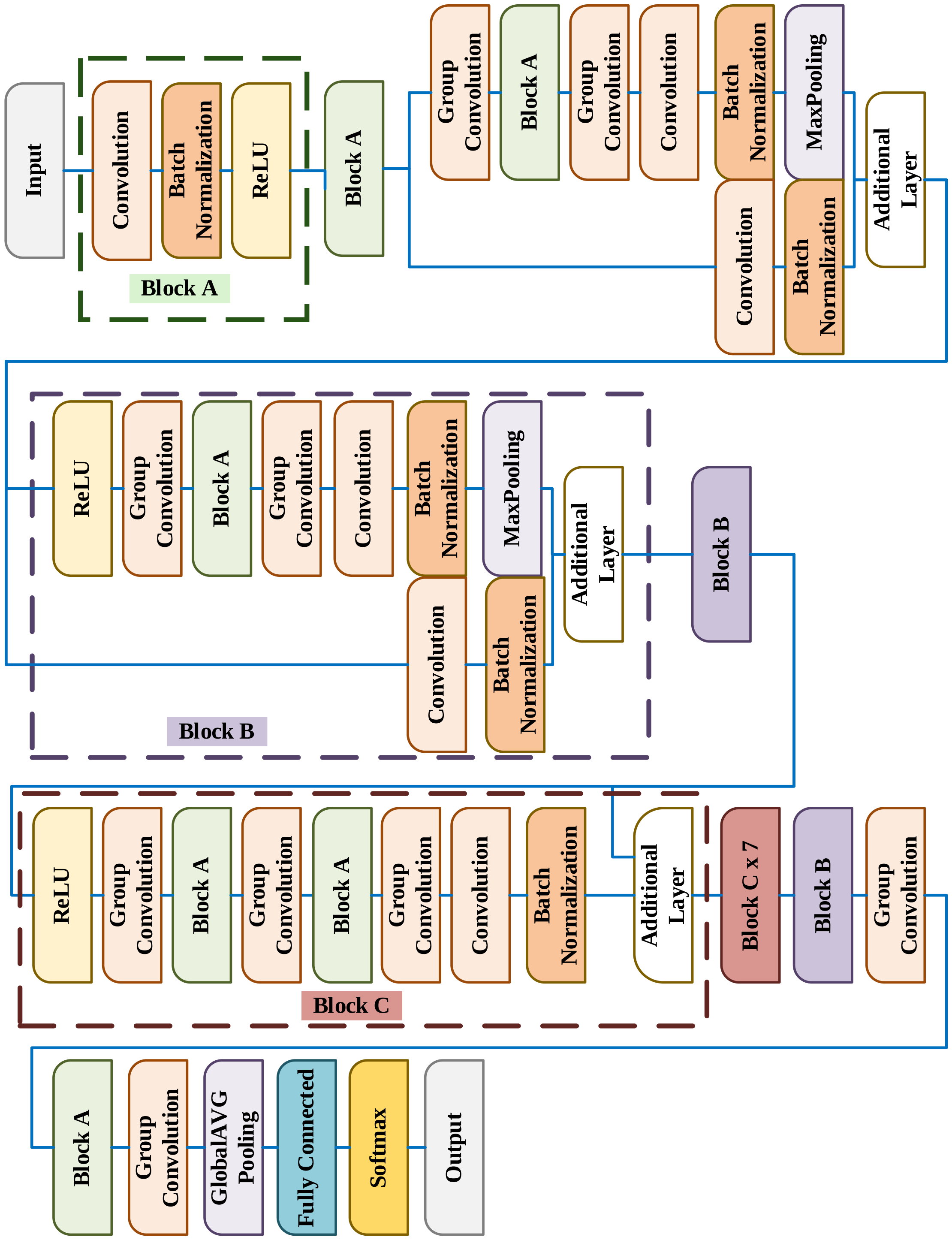

Figure 35: Xception: Extreme Inception with depthwise separable convolutions.

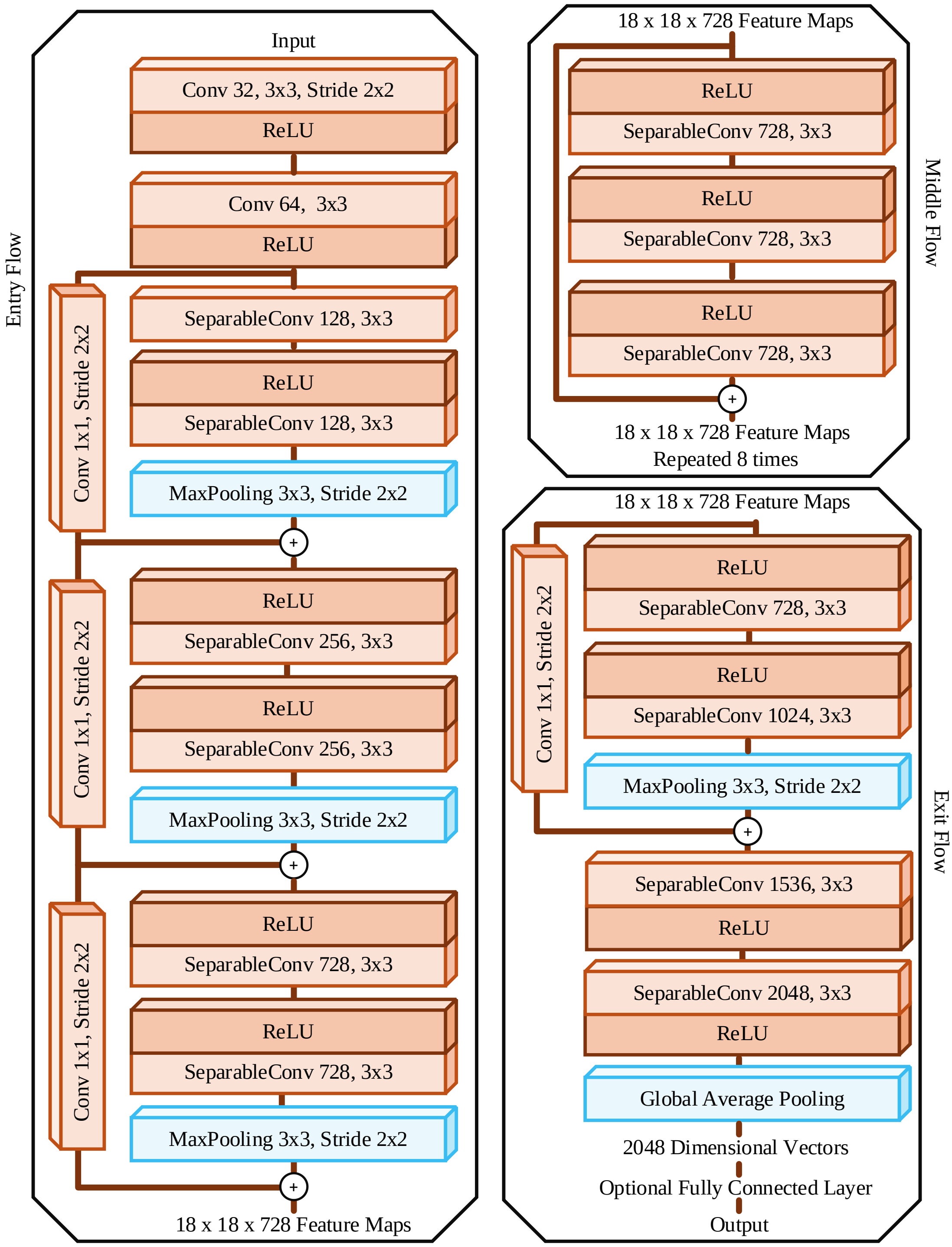

Figure 36: Xception involves Depthwise Separable Convolutions.

### 4.5.3. ResNeXt

ResNeXt, also recognized as the Aggregated Residual Transform Network, stands as a remarkable evolution that builds upon the foundations of the Inception Network [41]. [41] and their collaborators ingeniously harnessed the principles of division, transformation, and consolidation by introducing the innovative concept of "cardinality"[70]. Cardinality introduces an additional dimension, signifying the magnitude of transformations within the network [95]. The Inception network not only amplifies the learning capabilities of traditional CNNs but also optimizes the utilization of network resources. While the Inception network boasts power, its complexity requires adjustments at each layer due to its utilization of various filter sizes (3x3, 5x5, and 1x1) within specific architectural segments. In contrast, ResNeXt simplifies the design inspired by VGG's deep structure and refines GoogleNet by uniformly employing 3x3 filters across its critical components. Moreover, ResNeXt incorporates residual learning, enhancing the training of deeper and more intricate networks [96]. ResNeXt introduces multiple transformations within a divide, transform, and combine module, characterizing these transformations based on the notion of cardinality [41]. Convincingly demonstrated that an increase in cardinality leads to substantial enhancements in performance. To manage the intricacies of ResNeXt, low-level embeddings ($1 \times 1$ filters) are applied prior to the $3 \times 3$ convolution. Additionally, the training process benefits from the incorporation of skip connections [90].

#### 4.5.4. Inception family

The Inception family of CNNs can be classified as one of the approaches that consider network width [70]. In the Inception networks, a variety of filter sizes are utilized within a single layer, leading to improved outputs from intermediate layers. This strategy of incorporating diverse filter sizes proves effective in capturing a wide range of high-level features.

### 4.6. Feature-Map (Channel/FMap) Exploitation based CNNs

CNNs have transformed computer vision by automating the discovery of critical patterns in data [4]. This marks a significant departure from traditional ML approaches that necessitate manual selection of these patterns. CNNs employ specialized filters akin to miniature image scanners, which adjust their internal settings (weights) through learning. These filters become adept at identifying specific features such as edges or shapes within images. By continuously refining these filters, CNNs learn to prioritize the most relevant patterns for tasks like object classification, image segmentation, or identifying specific elements in an image [97].

One of the strong points of CNNs is hierarchical feature extraction. Through multiple convolutional layers, the network learns progressively complex data representations. Each layer will produce multiple feature maps, otherwise called channels, which capture different aspects of the image [97]. Not all these feature maps are equally important regarding the performed task. Some might carry irrelevant information or noise, and if not dealt with properly may lead to overfitting. Thus, careful selection or manipulation of feature maps can improve performance significantly in a network.

#### 4.6.1. Squeeze and excitation network

The Squeeze and Excitation Network (SE-Net) introduces an innovative component known as the SE block, which enhances object recognition by identifying crucial features (channels) within an image [97]. SE-Net demonstrates significant improvements on the ImageNet dataset due to the functionality of the SE block. This block's versatility permits its integration into various CNN architectures prior to the convolution layer. It functions through two primary stages: squeeze and excitation. During the squeeze stage, convolution kernels, which typically concentrate on localized information within their receptive field, might overlook broader contextual relationships across the image. The squeeze operation addresses this by capturing a global perspective, suppressing spatial information, and generating statistics for each feature map, effectively summarizing its content into feature-map motifs.

The excitation stage builds upon the global understanding achieved during the squeeze operation. It refines this information, accentuating significant features while downplaying less important ones. Consequently, the SE block emphasizes the most pertinent channels for object recognition within the image. This is represented by the function $g_{sq}(.)$ in equation (16).

$$\boldsymbol{s}_l^k = g_{sq}(\boldsymbol{F}_l^k) = \frac{1}{P\ X\ Q}\sum_{p,q} f_l^k(p,q) \qquad (16)$$

The SE block operates in two phases: squeeze and excitation. In the squeeze phase, for every feature map (channel) of the layer, a global feature descriptor $\boldsymbol{s}_l^k$ is computed. This is supposed to capture the essence of each channel by discarding the spatial information. Here, P × Q represents the original dimensions of a single feature map $\boldsymbol{F}_l^k$. These squeezed feature maps $\boldsymbol{s}_l^k = [s_l^1, \dots, s_l^k]$ are then fed to the excitation stage $g_{ex}(.)$. This stage uses a gating mechanism to capture dependencies among these various feature maps. In computer vision, this is how machines learn objects in an image technique that helps them to focus on important features. This is achieved in two steps. First, a light AI algorithm assesses the data, giving scores, or weights, to various parts of the image or feature maps. These scores serve as a kind of flag of importance, pointing to important information that is needed for object recognition, somewhat like highlighting key details in a picture. Then, taking those scores emphasizes important areas while diminishing the role of less relevant areas. This refinement sharpens the system's focus and selects distinctive information that is necessary for object recognition. Some of the detailed technical aspects are given in Equation of the original study.

$$\boldsymbol{y}_{l+1}^k = g_{ex}(\boldsymbol{S}_l^k) = g_{s_g}(w_2, g_t(\boldsymbol{S}_l^k, w_1)) \qquad (17)$$

Here, in equation (17), $\mathbf{y}_{l+1}^k$ represents the weight applied to the input feature map $\mathbf{F}_{l+1}^k$ of the next layer, i.e., $(l+1)$. The functions $g_t(.)$ and $g_{s_g}(.)$ apply ReLU-based nonlinear transformations and a sigmoid gate, respectively. $\mathbf{Y}_{l+1}^k = [y_{l+1}^1, \dots, y_{l+1}^k]$ signifies the weights for K number of convolved feature maps which are gain normalized before being fed into the next, i.e., the $(l + 1^{th})$ layer.

The excitation operation introduces motif-wise dependencies through a gating mechanism. Feature-map weights are determined by a two-layer feed-forward neural network, which controls model complexity and aids in generalization. The output undergoes ReLU activation to introduce non-linearity. The SE block utilizes a gating mechanism with sigmoid activation to model non-linear feature-map responses and assign weights based on relevance. This recalibration process dynamically adjusts feature maps by multiplying the convolved input with motif responses, illustrated in Figure 37 and Figure 38.

### 4.6.2. Competitive Squeeze and Excitation Networks

The Competitive Inner-Imaging Squeeze and Excitation for Residual (CMPE-SE) Network, introduced by Hu [98], expands upon the SE block concept originally proposed by Hu et al. [97]. SE blocks enhance the learning capabilities of deep residual networks by strategically assigning weights to feature maps based on their relevance in object differentiation. While SE-Nets are effective, their reliance on residual information within the ResNet architecture to determine feature map weights can result in redundant information and potentially limit the SE block's effectiveness.

To overcome this challenge, Hu's team proposed generating separate statistics for feature maps derived from both residual and identity mappings within the ResNet framework. This approach involves using global average pooling to gain a comprehensive understanding of feature maps. The CMPE-SE block then introduces a competitive mechanism that analyzes these statistics not only within residual feature maps but also between residual and identity feature maps. This competitive process fosters a refined weighting scheme for the feature maps by encouraging competition and enhancing the discernment of crucial features. While the specific mathematical details of the CMPE-SE block are not discussed here, its core innovation lies in introducing competition between residual and identity information to optimize the weighting of feature maps effectively.

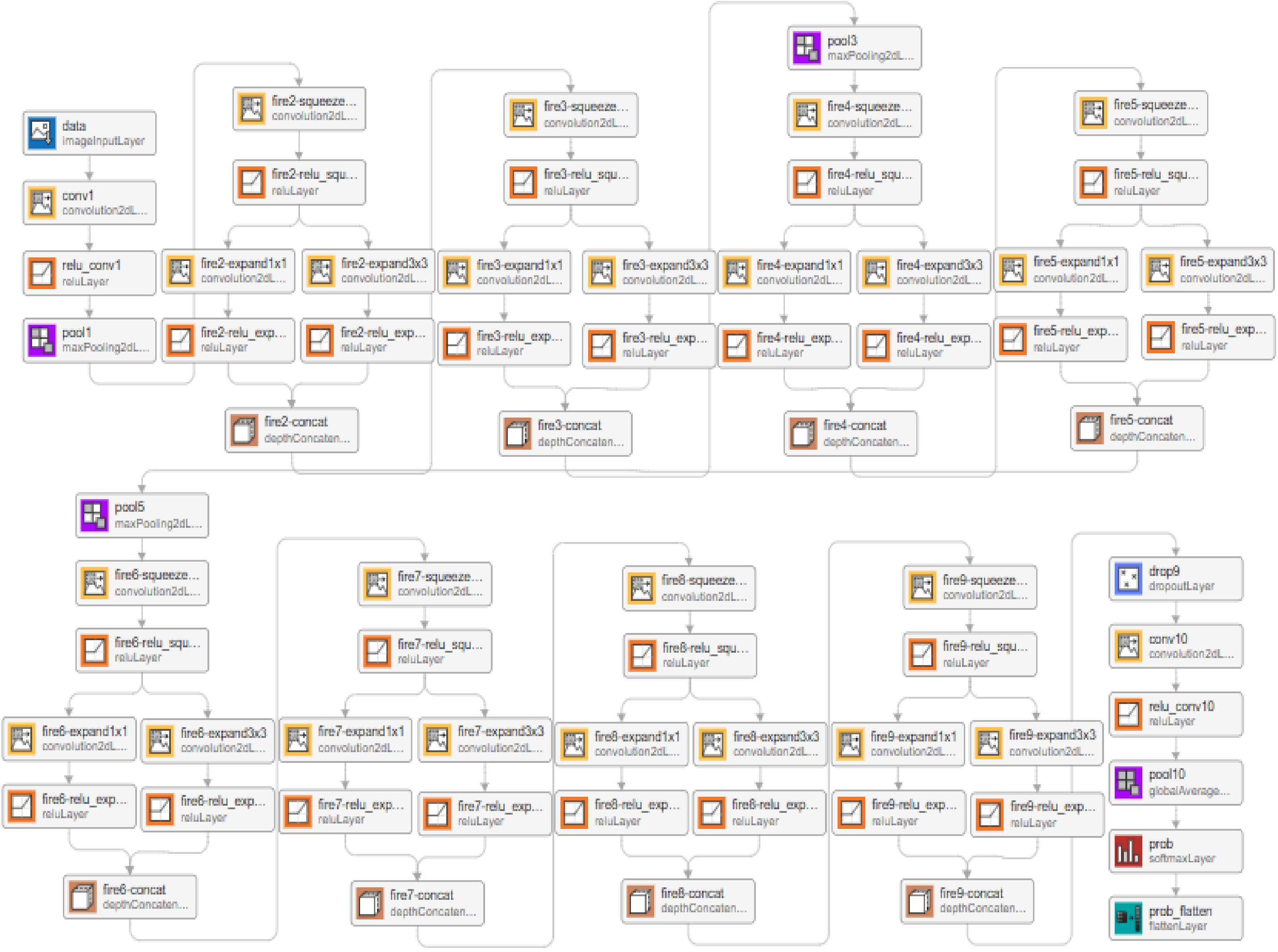

Figure 37: SqueezeNet: Reducing parameters with fire modules.

$$S_l^k, S_{m+1}^k = g_{sq}(F_l^k), g_{sq}, (F_{m+1}^{k'}) \tag{18}$$

$$Y_{m+1}^k = g_{ex}(g_k(S_l^k, S_{m+1}^k)) \tag{19}$$

$$F_{m+1}^k = Y_{m+1}^k . F_{m+1}^{k'} \tag{20}$$

In this equation, $\mathbf{F}_l^k$ and $\mathbf{F}_{m+1}^{k'}$ denote the individuality and residual mappings of the input $\mathbf{F}_l^k$ , respectively. The SE block (Equation. 18) involves applying the squeeze operation $g_{sq}(.)$ to both the residual and identity feature maps. Their outputs from this operation collectively serve as inputs to the excitation operation $g_{ex}(.)$. The operation $g_k(.)$ denotes the concatenation process.

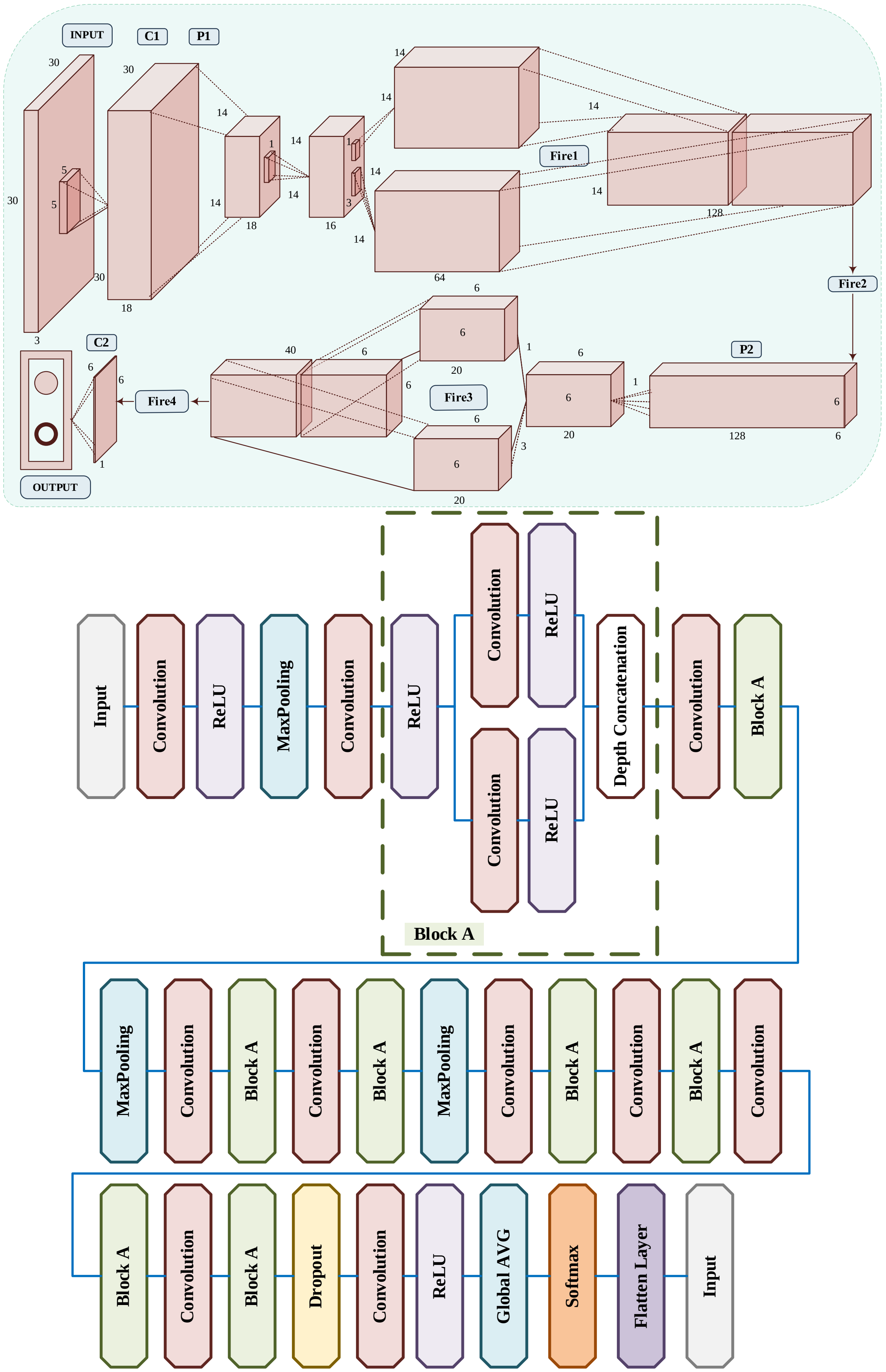

Figure 38: The SqueezeNet and Excitation block illustrate the process of computing masks to recalibrate feature maps.

This two-step process first involves creating masks using a mathematical operation (such as Equation 19) to identify key elements. These masks are then combined with the original data (multiplied, as in Equation 20) to emphasize the most crucial parts. This method aims to enhance the system's learning during training by focusing on essential features and the underlying details within each data piece, or feature map, of a residual block. Consequently, the training process gains a better understanding of the connections between features, leading to a more optimized system.

**a) Channel*(Input)* exploitation- based CNNs**

The portrayal of images significantly influences the effectiveness of image processing algorithms, encompassing a spectrum that includes traditional methods as well as approaches rooted in DL. An effective image representation should be capable of succinctly capturing the fundamental characteristics of an image. In the realm of computer vision tasks, conventional filters have long been employed to extract diverse layers of information from a single image [99]. These distinct representations are subsequently utilized as inputs for models to augment their performance [100]. Nevertheless, within the domain of CNNs, the efficacy of feature learning is intrinsically linked to the quality of the input representation. If the input data lacks diversity and fails to encapsulate distinctive class-related information, it can hinder CNN's role as a discriminative tool. In response to this challenge, the concept of "channel boosting" has emerged, involving the expansion of input channel dimensions through auxiliary learners integrated into CNNs. This innovation is aimed at bolstering the network's representational prowess [14].

### 4.6.3. Channel boosted CNN using TL

The Channel Boosted CNN (CB-CNN), introduced by [74], presents an innovative strategy to enhance the representational capacity of CNNs. This approach focuses on augmenting the number of input channels supplied to the network. Furthermore, CB-CNN achieves this by utilizing deep generative models to generate artificial channels, also known as auxiliary channels. These newly created channels are then incorporated alongside the original input channels. Essentially, CB-CNN enriches the network's information intake by introducing new channels that complement the existing data. For detailed mathematical formulations, please refer to Equations (21 and 22). However, the key concept is that CB-CNN employs generative models to produce informative auxiliary channels, thereby expanding the network's capability to capture intricate data representations.

$$I_B = g_k(l_c, [\boldsymbol{A}_1, \dots, \boldsymbol{A}_m]) \quad (21)$$

$$\boldsymbol{F}_l^k = g_c(I_B, K_l) \quad (22)$$

In Equation (21), $I_C$ signifies the original input channels, where $\mathbf{A}_M$ denotes a synthetic channel generated by the M[th] auxiliary learner. The function $g_k(.)$ acts as a combiner, merging the initial input channels with the auxiliary channels to produce an enhanced input denoted as $I_B$. This refined input is subsequently utilized by the discriminator. Equation (22) describes the generation of the k[th] resulting feature map, $\mathbf{F}_l^k$, achieved by convolving the enriched input $I_B$ with the k[th] kernel of the l[th] layer.

The empirical evidence presented by [80] underscores the crucial role of data representation in shaping classifier performance, as different representations can reveal diverse facets of information [101]. To enhance data representation, [74] collaborators leveraged the potential of TL and deep generative learners [102]. Generative learners, such as Autoencoders (AEs) in the CB-CNN framework, aim to capture the underlying distribution of data during the learning process, unraveling the factors that explain variations in the data. The application of inductive TL is distinctive, as it constructs an augmented input representation by enhancing the learned data distribution with the prime channel space (input channels).

The CB-CNN by [74] enhances a CNN's representational capacity through a "channel-boosting" phase, which incorporates additional channels alongside the original input. A key feature of CB-CNN is its use of TL in two stages: data generation and discrimination. In the data generation stage, deep generative models create auxiliary channels by learning patterns from existing data, generating new, complementary information. In the discrimination stage, these auxiliary channels are processed by a deep CNN that employs TL from the pre-trained generative models, improving class differentiation. CB-CNN also integrates multiple deep learners, with the generative models acting as auxiliary learners alongside the main CNN, enhancing the network's ability to learn complex representations. While initial studies focused on incorporating the channel-boosting block at the beginning of the network, [14] suggest that auxiliary channels can be introduced at any layer within the architecture, offering the potential for further refinement. CB-CNN has shown promising results with medical image datasets, indicating its potential value for medical image analysis tasks.

### 4.7. Attention-based CNNs

The cognitive mechanism known as attention is crucial for recognizing and understanding complex visual scenes. Human vision, like neural networks, processes information at various levels of detail through a focused approach. Rather than passively absorbing everything at once, we gather visual information through a series of targeted glimpses. This mechanism prioritizes specific regions of a scene based on their relevance to the task or context, allowing us to focus on important details while considering multiple interpretations of objects within those areas. This selective processing not only enhances object identification but also helps us grasp the broader context and relationships between them, enabling effective navigation and comprehension of the visual world [103], [104]. A similar concept of interpretability has been injected into RNN and LSTM networks as well [103]. The RNN and LSTM networks make use of attention modules to predict further data. The modules grant weights to newer samples based on how these were occurring at previous steps. Further refinement of the concept of attention has been done by the research community for its usage in CNNs, which allows for better encoding of features without bypassing computation-related bottlenecks. It will add an attention mechanism to CNNs so that they can recognize objects intelligently, even in scenes with heavy clutter in the background.

#### 4.7.1. Residual Attention Neural Network

It proposes an attention mechanism specially designed for object recognition, enhancing feature representation in CNNs [66]. Unlike the traditional CNNs that have complex representations of pixels, Residual Attention Neural (RAN) embeds an attention module into its residual block architecture. This module essentially consists of two branches: trunk and mask. The attention module of RAN synergizes bottom-up and top-down learning strategies. Whereas bottom-up processing in the network learns simple features from input images, top-down information refines these features relative to the overall object context. The integration of this information together in the single forward pass optimizes the efficiency of RAN such that top-down feedback can guide improved feature extraction.

While the concept of integrating these learning strategies is not entirely novel, previous research with Restricted Boltzmann Machines [105] and Deep Boltzmann Machines [105], as explored by [106] has demonstrated the efficacy of top-down information in network optimization and feature learning. RAN advances these principles by integrating both strategies within a unified module, thereby achieving more efficient and context-aware feature representation for tasks in object recognition. In the context of RAN, the attention module produces object-aware soft masks, represented as $g_{sm}(.)$, at each layer for the input feature-map $\mathbf{F}_l^k$ [107]. These soft masks are described by Equation. (23), recalibrate the output of the trunk branch $g_{tm}(\mathbf{F}_l^k)$, effectively serving as control gates for individual neuron outputs:

$$g_{am}\left(\boldsymbol{F}_l^k\right) = g_{sm}(\boldsymbol{F}_l^k).g_{tm}(\boldsymbol{F}_l^k) \tag{23}$$

While dealing with images containing clutter, noise, or complex scenes, several CNN architectures face problems. RAN overcomes this problem by embedding several attention modules within its hierarchical architecture [66]. Since the nature of RAN is hierarchical, the network dynamically can give different weights on each feature map concerning information channel importance at different layers. Multi-level features analysis allows RAN to filter out irrelevant features and give more emphasis on the features that are important for a recognition task. Besides, RAN introduces residual units to enable deep hierarchical learning; the deeper the network goes, the more complex features are extracted. Furthermore, RAN introduces three kinds of attention levels-mixed, channel, and spatial-which model object-aware features in different levels of image details to enhance the capability of comprehensive scene understanding [14].

#### 4.7.2. Convolutional block attention module

Recently, attention mechanisms have become an essential component that helps enhance feature learning within CNNs, a fact widely discussed in current works, such as RAN and SE-Net [66], [97]. In the work from 2018, the authors continued this direction to propose the Convolutional Block Attention Module (CBAM) [108], which works similarly to SE-Net by implementing an attention mechanism that could help optimize CNN performance [108]. SE-Net deals mainly with deciding how much importance each feature map is toward an image classification task and may not consider the spatial contextual information, which plays a significant role in object detection. CBAM remediates this with the attention mechanism in two consecutive steps: Channel Attention and Spatial Attention (CBAM). CBAM determines the significance of each feature map or channel by integrating max pooling to extract the most salient object features with global average pooling to determine the importance of the feature map as a whole. These are then further refined by Spatial Attention, which incorporates the spatial information using both average and max pooling with the spatial dimension. This two-stage strategy amplifies the network's ability to focus on important image regions while amplifying informative feature maps.Another important advantage of CBAM is its efficiency. With the sequential application of channel and spatial attention, bolstered by pooling, comes a compact 3-D attention map that reduces model parameters and computations. Being of simple design, it easily plugs into any existing CNN architecture, hence very practical for being versatile in boosting the performance of a network.

#### 4.7.3. Concurrent spatial and channel excitation mechanism

The study presented a new approach that combined spatial information with feature-map information, thus becoming more applicable for segmentation tasks [97]. First, it introduced the cSE Module: Squeezing Spatial and Exciting Feature-Map Information, inspired by the SE-block concept of scaling feature maps relevant to object detection. Next, it introduced the sSE Module: Squeezing Feature-Map Information and Exciting Spatial Information, which favors spatial locality over feature-map data to improve segmentation. Finally, the cSE Module: Concurrently Squeezing and Exciting Both Spatial and Feature-Map data selectively highlights object-specific feature-maps using both spatial and feature-map information [98]. Their study used an Autoencoder-based CNN that combined the different modules for segmentation after the encoder and decoder layers.

#### a) Lightweight models

The growing complexity of CNNs has driven research toward lightweight models. Although traditional block-based methods have reduced model size and parameters, deploying them on resource-constrained systems, such as embedded devices, remains difficult. To overcome this, researchers are designing highly efficient architectures optimized for such environments. MobileNet [18] employs depthwise separable convolutions to streamline operations. These convolutions split the computationally intensive standard convolutions into two distinct steps: depthwise convolutions, which process each channel independently, followed by pointwise convolutions that reduce the number of channels as depicted in Figure 39, Figure 40, and Figure 41. In contrast, ShuffleNet [19] reduces complexity through pointwise group convolutions, offering a more efficient alternative to dense 1x1 convolutions commonly used in CNNs as represented in Figure 42 and Figure 44. ShuffleNet mitigates the limitations of group convolutions through a channel shuffle operation, enhancing cross-channel information flow. Meanwhile, MnasNet [20], shown in Figure 43, employs automated search to identify optimal architectures for specific platforms. By addressing latency constraints and leveraging a hierarchical search space, MnasNet balances accuracy and efficiency, demonstrating effective strategies for lightweight CNNs on resource-limited devices.

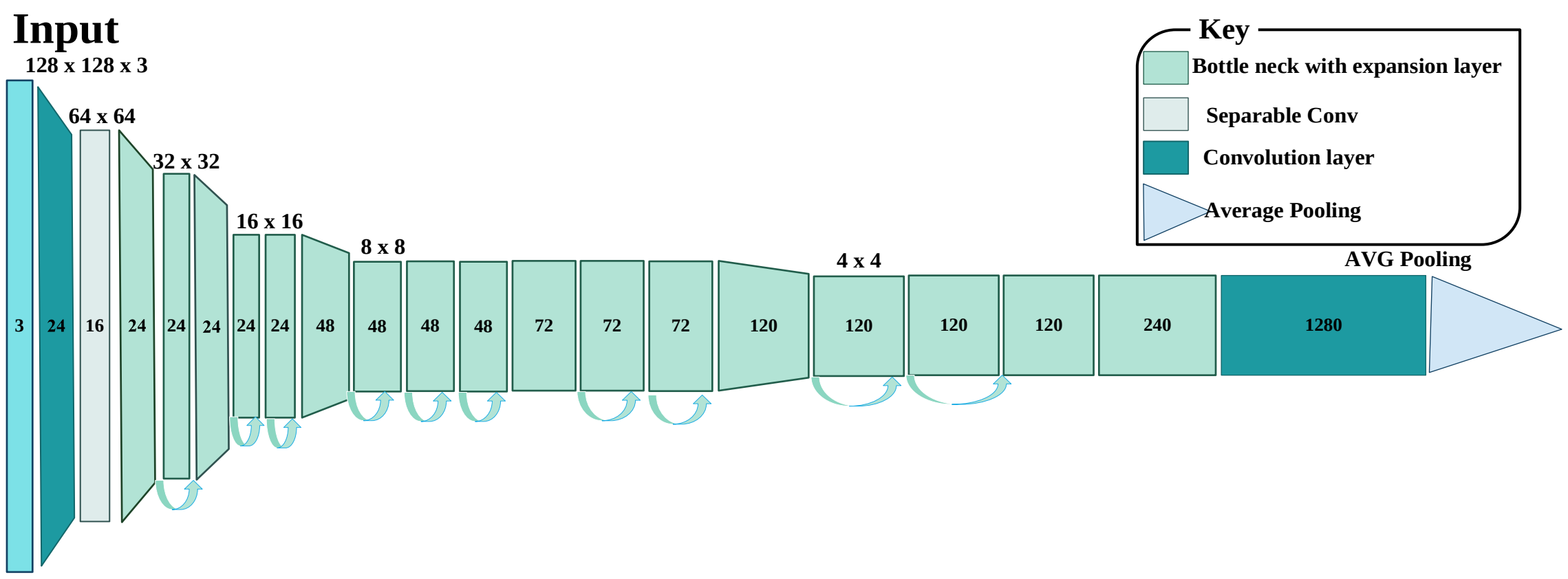

Figure 39: MobileNet V2: Optimized for mobile and embedded with inverted residuals and linear bottlenecks.

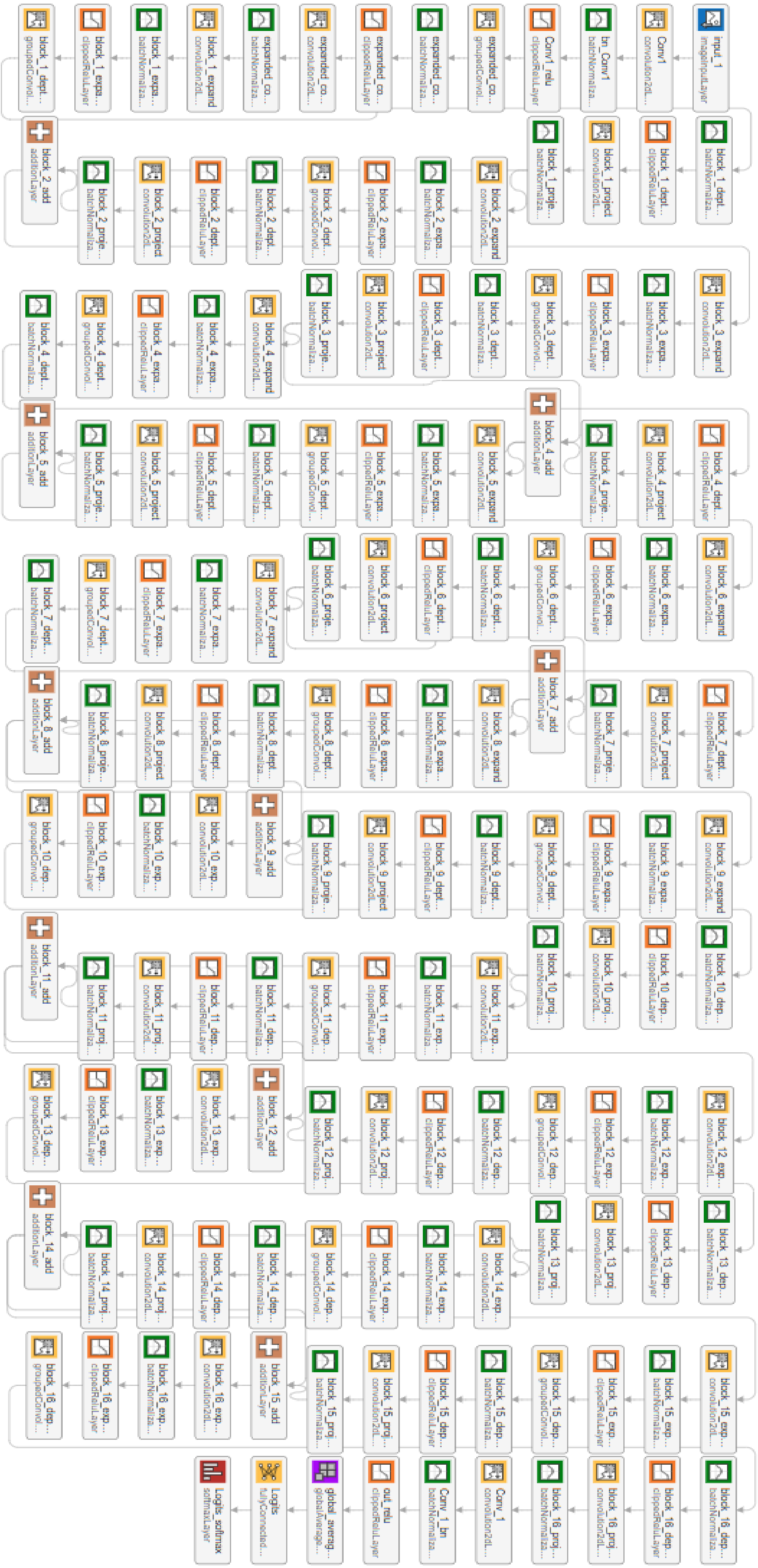

Figure 40: MobileNet-V2: Designed for mobile and embedded devices.

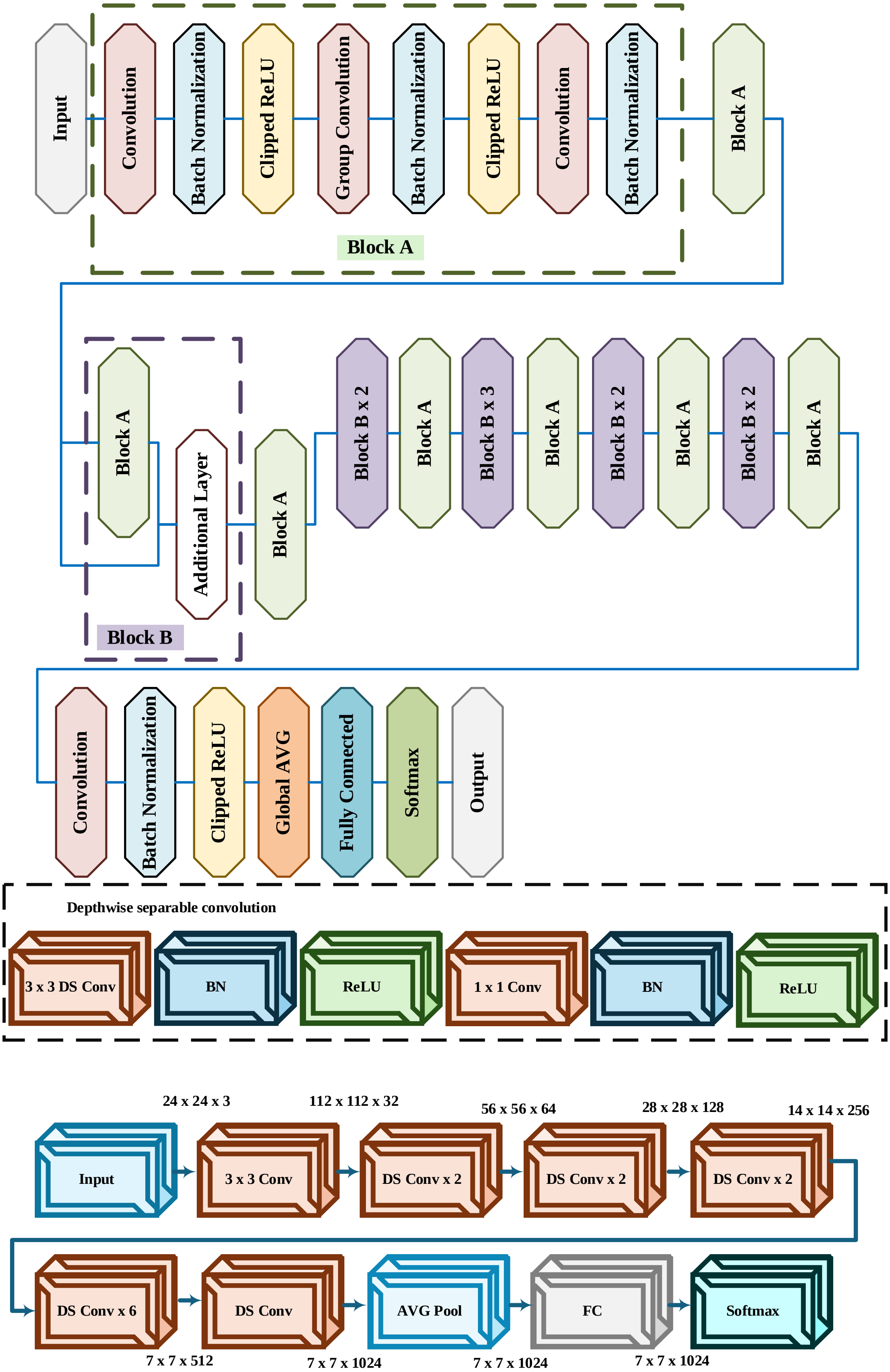

Figure 41: MobileNet architecture.

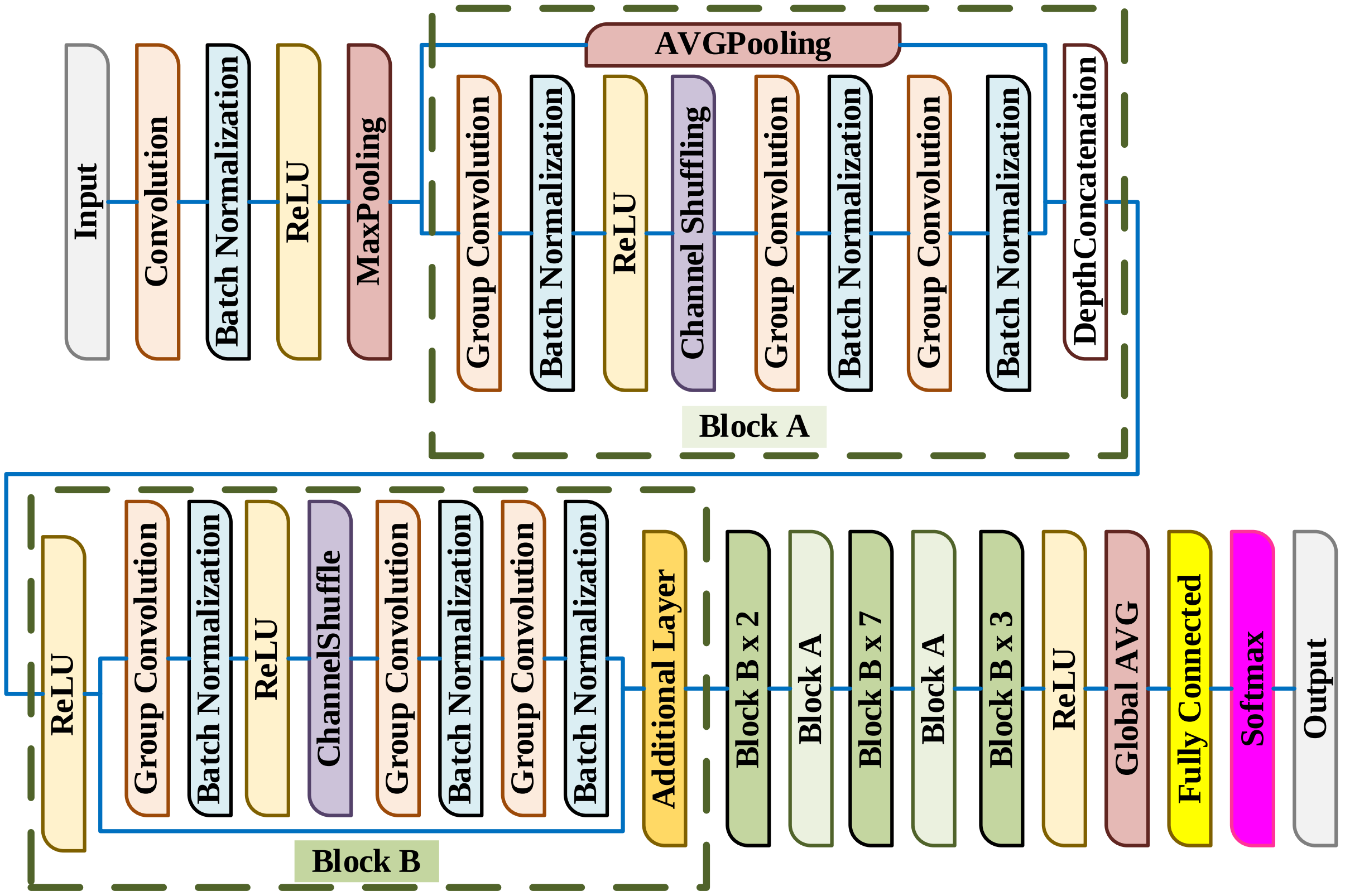

Figure 42: ShuffleNet: Efficient architecture for mobile and edge devices.

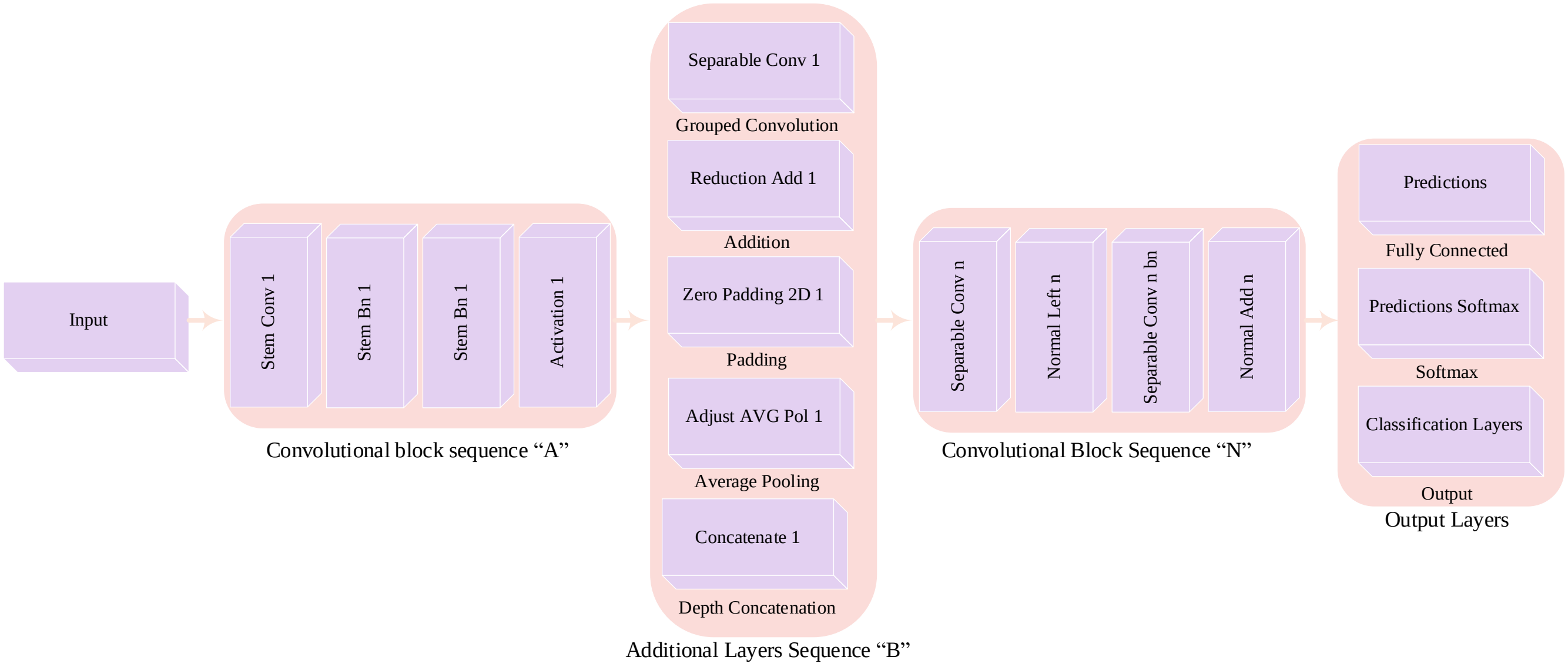

Figure 43: NasNet-Mobile: Optimized NAS model for mobile devices.

## 5. Applications of CNNs

CNNs excel in diverse ML tasks such as object identification, pattern recognition, categorization, predictive modeling, and image segmentation [109], [110]. However, their effectiveness heavily depends on the availability of substantial training data. CNNs have achieved significant success in various domains including traffic sign recognition, medical image segmentation, NLP, face detection, and object identification in natural images. These achievements are particularly notable in domains with ample annotated data [40].

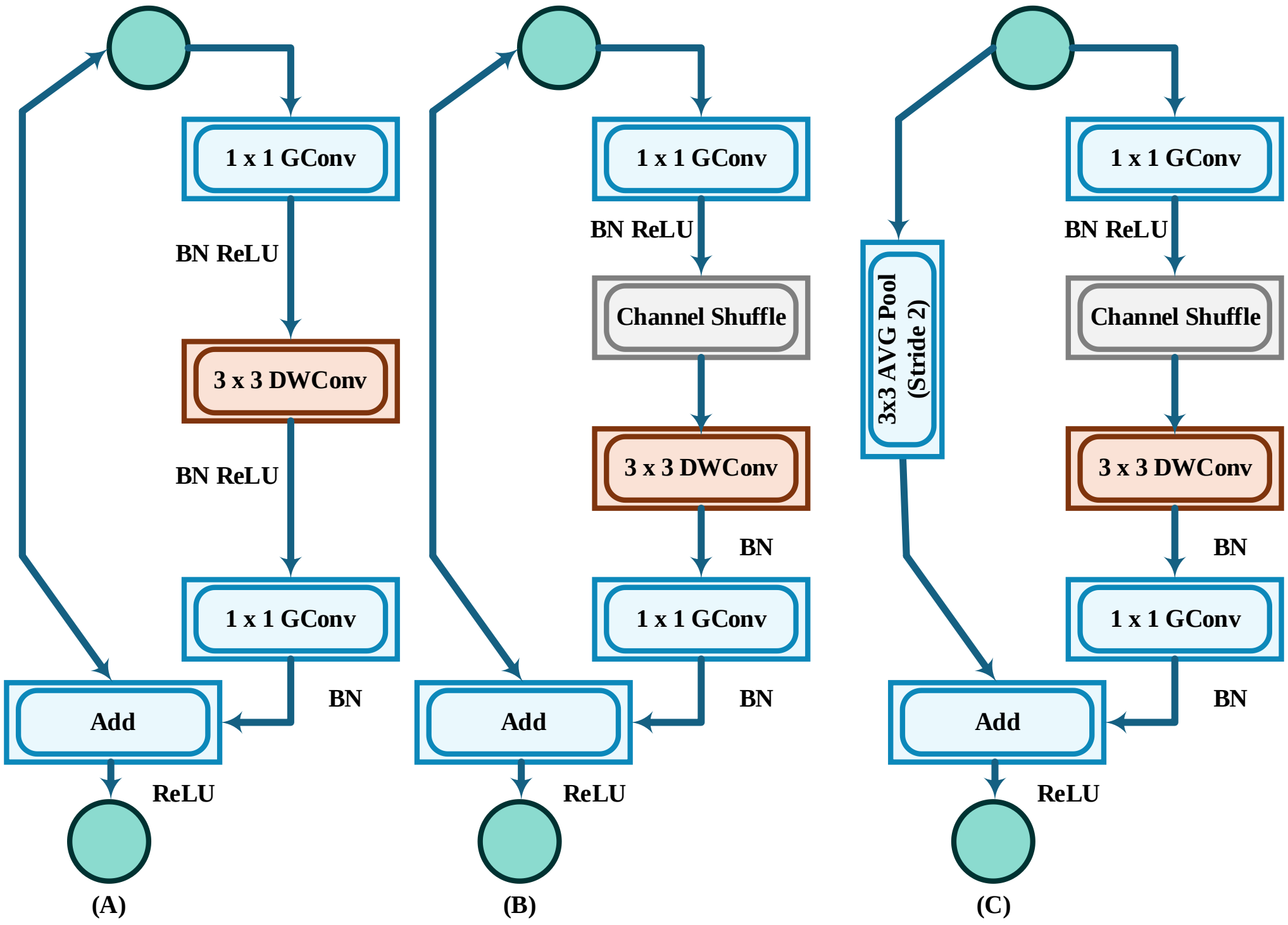

Figure 44: ShuffleNet is a CNN architecture that is extremely computation-efficient.

## 5.1 Computer Vision

Computer vision enables computers to interpret images and videos, facilitating tasks like facial recognition, body movement tracking, and action identification. However, challenges such as varying lighting conditions, head positions, and expressions complicate face recognition. Despite these challenges, deep CNNs have been successful in detecting faces [111]. By detecting the different parts of the body, pose estimation identifies whether a person is standing, jumping, and so forth. The methods for pose estimation range from the use of CNNs in locating these parts to step-by-step CNN structures with heatmaps [2]. For action recognition, combining 3D CNNs with LSTM networks captures the actions effectively and interprets them accordingly [7]. Another work improved action recognition using a 3D CNN system that was supervised to train [112].

### 5.1.1. Face Recognition

Face recognition faces challenges such as lighting, pose, and facial expression. Farfade, 2015, presented a very resilient Deep CNN for face detection featuring varied angles and occlusions. Zhang, 2016, discussed a multitasking cascaded approach for CNNs where the detection results showed higher performance compared to the other current techniques.

### 5.1.2. Pose Estimation

Pose estimation, which is difficult because of diverse body poses, was addressed by using the cascade-based deep neural network approach [113]. The basic steps involve initial heat map detection in cascaded architecture, followed by regression over detected heat maps to improve accuracy.

#### 5.1.3. Action Recognition

In activity detection systems, action recognition involves identifying activities along with their translations and distortions. A system comprising 3D CNN with LSTM was proposed for successfully identifying activities from video frames, outperforming other algorithms [7]. Another system for action recognition based on 3D CNN utilized multi-channel input frames [112].

### 5.2. Natural Language Processing

New opportunities for NLP based on CNNs are researched by scientists. Dynamic CNN with k-max pooling modeled the relations of words in a sentence effectively and improved text understanding. Another approach employed a CNN-based architecture to solve several NLP tasks like chunking, named entity recognition, and role labeling while keeping semantic coherence. The next wave of research has used CNNs for multilingual sentence-matching tasks judging the similarity of sentences across languages to help with large-scale machine translation and cross-lingual information retrieval. These studies underline the fact that CNNs have more and more powerful roles in NLP, which might improve language understanding and enhance human-computer interaction.

#### 5.2.1. Statistical Language Modeling

Statistical language modeling uses incomplete word sequences as input with the CNN architecture of genCNN and ByteNet. These use convolutional and dilated CNNs to model sequential data effectively. The architecture provides hierarchical representations that capture long-range information, hence giving better results in language tasks.

#### 5.2.2. Text Classification

Effective local relations in sentences are achieved using CNN architectures in NLP for text classification. From single-layer to deep architecture models, the use of a convolutional layer is followed for sentence modeling. Variable-length sequences are dealt with adaptively through techniques such as k-max pooling and dynamic k-max pooling to improve accuracy in classification despite sequence variations.

### 5.3. Object Detection and Segmentation

In computer vision, object detection aims to identify and classify multiple objects within an image. R-CNN was a pioneering method that significantly advanced the field. Building on this, "Fast R-CNN" introduced a fully connected CNN, streamlining detection processes. Another region-based CNN approach demonstrated impressive precision in object detection on the PASCAL VOC dataset. A multi-region-based CNN architecture was advanced to enhance feature extraction and image semantics understanding [39]. CNNs also excel in image segmentation tasks, identifying objects, and delineating their boundaries. Leading architectures in this field include FCN, SegNet, Mask R-CNN, and U-Net. These models perform well in both semantic segmentation, labeling image content (e.g., sky, car), and instance-based segmentation, identifying and outlining individual object instances. This demonstrates CNNs' versatility in addressing various computer vision challenges, from object detection to precise image segmentation. Furthermore, advancements in object detection include the introduction of R-CNN for rapid object detection improvement [114]. Fully connected CNNs based on object locations were proposed to enhance detection accuracy [115]. A multi-region-based deep CNN approach was developed, demonstrating high accuracy on benchmark datasets [39].

### 5.4. Image Classification

CNNs excel in image classification, particularly in medical image analysis for cancer diagnosis using histopathological images. Recent advancements include CNN-based methods for breast cancer diagnosis, focusing on non-mitosis instances and data augmentation to address class imbalance challenges. In traffic sign analysis, one study used CNNs for effective classification using the German benchmark dataset.

### 5.5. Image Recognition and Classification

CNNs are integral to image recognition and classification tasks, utilizing 2D convolutions to discern images accurately. These models excel in detecting edges, textures, and shapes essential for classification tasks like Optical Character Recognition (OCR), animal species classification, and handwritten digit recognition. CNNs have demonstrated exceptional performance in competitions such as ImageNet.

### 5.6. Object Detection and Localization

Object detection uses 2D and 3D convolutions to locate and identify various items inside a picture. CNNs with region proposal mechanisms and anchor-based methods excel in detecting objects of various scales and aspect ratios. Accurate object localization is facilitated by pooling layers and convolutional sliding windows, finding applications in surveillance, robotics, and autonomous vehicles.

### 5.7. Speech Recognition

CNNs are increasingly effective in speech recognition. A study introduced CNN-based system for speaker-independent speech recognition, significantly reducing error rates compared to previous methods. Studies also explore diverse CNN architectures, experimenting with weight sharing and pre-training phases for robust performance in phone and vocabulary recognition tasks. CNNs are also applied in speech emotion recognition, combining CNNs with LSTM for improved emotion recognition from speech.

### 5.8. Video Processing

CNNs excel in video processing by handling spatial (image content) and temporal (sequence of images) data, making them suitable for tasks such as action recognition and event detection. Researchers have explored various CNN architectures for video analysis. For example, a system was developed using CNNs and Temporal Action Graphs to identify specific moments within videos, such as scene changes [116]. Similarly, combining 3D CNNs with Long Short-Term Memory (LSTM) networks has achieved high accuracy in action recognition [66]. Moreover, CNNs are effective in detecting critical events in videos; for instance, CNNs have been utilized to identify smoke and fire in video footage, potentially aiding real-time fire alarm systems [117]. Action recognition has also advanced through a three-stream CNN framework that extracts features from different motion aspects within a video [118], and the effectiveness of combining CNNs with bi-directional LSTMs for action recognition has been demonstrated [119]. These studies highlight CNNs' versatility in extracting meaningful information from videos, driving advancements in video analysis and computer vision applications.

### 5.9. Low-Resolution Images

Researchers have examined CNNs for enhancing image quality and resolution. The LR-CNN architecture demonstrates effectiveness in low-resolution image classification [120], while CNNs combined with LSTM networks enable action recognition in inherently low-resolution thermal images [121]. These investigations underscore the adaptability of CNNs in addressing image quality and resolution challenges across various computer vision tasks.

### 5.10. Resource-Limited Systems

Despite computational demands, CNNs find applications in embedded systems for tasks like license plate recognition and mobile applications. CNNs have been optimized for FPGA platforms to enhance efficiency in resource-constrained settings [122]. MobileNet, ShuffleNet, and ANTNets are CNN architectures tailored for mobile device applications. A real-time system for detecting driver drowsiness using MobileNet with SSD was designed, making it suitable for smart devices [123].

### 5.11. 1D Data Analysis

CNNs are effective in analyzing 1D data. One study utilized 1D-CNN with RNNs for intrusion detection in network traffic, demonstrating superior performance on the KDDCup 99 dataset. In structural damage detection, another study employed 1D-CNNs for real-time extraction of damage-sensitive features from accelerated signals. 1D-CNNs were applied in biomedical data for cardiac irregularity detection, achieving high accuracy on the MIT-BIH Arrhythmia database [124].

### 5.12. Cryptocurrency Analysis

#### 5.12.1. Sentiment Analysis for Cryptocurrency Price Prediction

Research into accurate cryptocurrency price prediction has focused on various ML algorithms, particularly LSTM models, which have shown promising results. Comparative studies highlight the effectiveness of LSTMs in capturing the volatile nature of cryptocurrency markets. Beyond LSTMs, researchers have developed hybrid models combining LSTMs with Recurrent Neural Networks (RNNs) and Gated Recurrent Units (GRUs), demonstrating enhanced capabilities. Attention mechanisms, such as those in SAM-LSTMs, have been used to focus on crucial financial news information for Bitcoin price prediction. Moreover, multi-scale analysis that integrates several algorithms into one has been tried out because it captures the price movements in higher detail across various time horizons. Besides deep reinforcement learning, DFFNNs, and CNNs have been used for the same task. Some have even taken into consideration such external impacts on cryptocurrency price volatility as COVID-19 epidemics. It is indicated that still, efforts are continuing regarding the improvement of cryptocurrency price prediction with a variety of DL methods.

### 5.12.2. Cryptocurrency Portfolio Construction

DL methods have been extensively used for constructing cryptocurrency portfolios, among which DRL models are widely used. LSTM models have shown their performance in risk management and learning temporal information from historical transactions. The volatility prediction in cryptocurrency portfolios has also been considered with hybrid models such as LSTM combined with GARCH. Some of the methods utilized for sophisticated financial portfolio management, including cryptocurrencies, are deep factorization machines and spiking neural networks. Among the DRL models, the utilization of DDPG has already outperformed conventional ML methods to build up and manage cryptocurrency portfolios. In the case of external events analysis, for example, the COVID-19 pandemic, on cryptocurrency portfolios, event analysis uses DL models. Despite these models' great successes, challenges remain regarding high volatility and limited historical data. Future research probably will be directed at integrating external factors such as macroeconomic variables and the incorporation of state-of-the-art DL models for better portfolio construction.

### 5.13. CNN Applications in Medical Images

It includes the analysis of diagnostic images, including MRIs, CT scans, and X-rays, in medical imaging analysis. Advanced techniques involving 3D convolutions and dilated convolutions are used to analyze the given images. 3D convolutions operate on 3D medical data, aiding in capturing the spatial relationships that are critical in performing various tasks like tumor detection, organ identification, and disease classification. Dilated convolutions enhance feature extraction and comprehension of image content for the accurate detection of abnormalities. These convolution techniques have transformed medical image analysis into a clinically useful modality. CNNs, through these techniques, have successfully segmented a multitude of anatomical structures within medical images regarding the brain, heart, breast, eye, and fetal tissues.

## 6. Challenges and Future Directions

Serious challenges also exist with DL models, mostly requiring collaboration across disciplines: data collection and preprocessing methods, algorithm improvement, training with fairness, decision interpretability methods, and making models safe against attack. Collaboration with domain experts and communities usually serves as both a beneficiary of and a prerequisite for the full unleashing of DL. The following are some of the fundamental challenges in DL.

### 6.1. Data Availability and Quality

Generally, DL models require a large amount of labeled training data for optimal performance. However, collecting high-quality labeled data can be expensive, time-consuming, and sometimes impossible for many specialized domains or sensitive information, such as cybersecurity. While data augmentation techniques can artificially increase the available data, the generation of sufficient training data remains a big challenge. Other challenges include overfitting, which includes overconnecting or closely relating the model to particular training instances. This type of overfitness harms the generalization capability of the model to newer data. The complexity of these models should therefore be balanced, while regularization techniques are also used to ensure robust generalization performance. Researchers are actively trying to find ways to increase data efficiency in DL. Few-shot learning, active learning, and semi-supervised learning are being worked upon to enable the models to learn from a few data or utilize unlabeled data together with labeled data.

### 6.2. Ethics and Fairness

First and foremost, of utmost importance is to ensure that DL is done in a manner that is non-discriminatory and ethical. DL models, normally trained on very large datasets, contain biases within the dataset which, if propagated, ensure continued imbalance within an already unfair society. DL influences more and more areas of hiring, loan approval, criminal justice, and so on, these ethical considerations raise urgent needs. Making DL ethical includes several key steps: first, techniques have to be developed for the detection and mitigation of biases in both training data and models themselves. Secondly, it will create transparent policies and standards that will enable the deployment of a nondiscriminatory AI and not only its development. Guarantees of nondiscrimination should be assured during the design and training of DL models to treat individuals fairly. A multidisciplinary approach to such issues allows us to responsibly harness DL for social good and in developing and benefiting all from DL.

### 6.3. Interpretability and Explainability

While DL systems have become much better in the past few years, it remains difficult to understand how decisions are made in their deepest parts. The inner workings of a neural network become increasingly difficult to understand as neural networks become large and complex; these networks are pretty much black boxes for most people. This lack of interpretability leads to a lack of trust, making wide acceptance impossible in domains where accountability and oversight are fundamental, such as in health and finance. The balance can be drawn out between high performance and interpretability. Increasing the interpretability and explainability of the model will help various stakeholders understand why such a prediction has been made. This ushers in transparency, and trust, enabling informed decisions based on model reasoning, and accountability in deployment.

### 6.4. Robustness to Adversarial Attacks

Decision boundary vulnerabilities are one of the major threats to DL models using adversarial attacks. An attacker might subtly alter the input data, usually imperceptible to human inspection, into making bad predictions with a model. This sensitivity shows that there is something very crucial that lacks real-world practical application: robustness. The success of adversarial attacks is calling into question the reliability of DL systems, particularly within high-stakes domains. It will require state-of-the-art defense mechanisms and intrinsic robustness enshrined within model development to counteract this. Any consideration of security needs to be placed front and center in design and training if the emergent models are to resist these attacks and the data against corruption. It is an issue of trust and, correspondingly, wider adoption for DL across a wide array of domains.

### 6.5. Catastrophic Forgetting

Catastrophic interference, now more commonly referred to as 'catastrophic forgetting', is a profound DL problem wherein a model forgets what it has learned after the training of more recent data takes place. This is damaging practically because a model can forget to perform well on a task that it has previously mastered. It is essentially embedded like deep neural networks, which comprise a huge number of parameters with complex internal representations. Training on new data often causes unintentional changes in network weights and connections, thus probably affecting knowledge previously learned. The need has become imperative for models that could reduce the impact of this challenge and preserve knowledge across a wide range of diverse learning scenarios.

### 6.6. Safe Learning

Safety-first methods play an increasingly important role in designing and training DL models to make the models reliable and robust across their life cycles to minimize accidents, errors, and unforeseen breakdowns when deployed into applications.

The slack safety of DL models in safety-critical applications, like autonomous robots or medical diagnosis, could result in disastrous outcomes. Their solutions also relate to the integration of risk estimation techniques, management of data uncertainties, and anomaly detection mechanisms that spot unusual behaviors to avert possible failures. While much progress has been made, safe DL remains a dynamic field, with active research on how AI will be responsibly and reliably used across diverse areas.

### 6.7. TL and Adaptation

Utilization of pre-trained knowledge by DL is effective; however, applying this to new tasks is quite challenging. Training over large-scale datasets enables a model to grasp interesting features; transferring knowledge across different data distributions, semantics, and context is not so straightforward. Careful strategies are required when adapting such pre-trained models for specific applications. This might be achieved either through fine-tuning with new data, domain adaptation techniques, or by devising architectures different from pre-training that can handle various inputs with semantic variations. The key is always the balance between leveraging pre-trained knowledge and tuning the model for better representation of the new data to capture relevant information, making the transferred representations accurate and retaining nuances of the particular task. The full utilization of AI in diverse domains requires the mastering of the complexities of TL.

### 6.8. Challenges Specific to Deep CNNs

Deep CNNs have always exhibited outstanding performances for datasets with time series or grid-like structures. However, these models face a variety of issues including:

**6.8.1. Interpretability and Explanation:** Deep CNNs are often considered mysterious black boxes, where the process of explanation and interpretation of their outputs is hard.

**6.8.2. Sensitivity to Noise:** Noisy image data of CNNs lead to higher misclassification rates. Even a slight random noise added to an input image can easily result in different classifications between the original and the slightly perturbed versions.

**6.8.3. Feature Visualization:** For some tasks, this is an important part: knowing the intrinsic features extracted by deep CNNs before classification. Techniques such as feature visualization within CNNs may be useful.

**6.8.4. Data Dependence:** Another big weakness of Deep CNNs, when compared to human learning, is that these are bound by supervised learning. While in the case of humans, it would suffice to learn from a few examples, for effective training Deep CNNs require masses of labeled data. This can increase the complexity in practical cases where gathering such large volumes of labeled data may be bothersome and costly.

**6.8.5. Hyper-parameter Sensitivity:** Hyper-parameters play a major role in CNN which has a huge influence on the performance of the model. Even slight changes in those hyper-parameters have drastic effects on the accuracy and performance of CNN. This inflicts the importance of selection and optimization strategies for finding the best hyper-parameters for the task at hand.

**6.8.6. Hardware Requirements:** The training process of CNN is highly consuming in hardware resources, especially GPUs. Recently, there has been a growing demand for the efficient deployment of CNNs into embedded and smart devices.

**6.8.7. Estimation of Object Pose and Location:** Estimating the pose, orientation, and location of an object in a vision-related task may be comparatively difficult for CNNs. Data augmentation will increase the diversity of internal representations learned by CNNs, which may improve their performance.

### 6.9. Additional Areas of Exploration for CNNs

#### 6.9.1. Ensemble Learning:

Ensemble learning combines multiple CNN architectures into a much-improved ability for the generalization of any model with high accuracy across different image categories. In this regard, ensemble learning, due to diversity in CNNs, extracts broader semantic representations from data and improves the robustness and accuracy of the model. Other techniques include batch normalization, dropout, and new types of activation functions, which have been critical for regularizing training and avoiding overfitting in CNNs.

#### 6.9.2. Generative Learning:

CNNs have shown promise as generative learners for tasks such as image segmentation. If one could effectively leverage the generative capabilities of CNNs at the stages of feature extraction, the representation capability of the model would increase significantly. For future research, one might want to investigate new ways of enhancing the learning of CNN by incorporating informative feature maps obtained via auxiliary learners in its intermediate processing steps.

### 6.9.3. Attention Mechanisms

Biological inspiration for attention mechanisms comes from the human visual system. Future research efforts in attention mechanisms will continue to be one of the leading ways to extract information from images with contextual relationships preserved. It is expected that in the further stages of learning, special attention will be directed toward preserving the spatial significance of the objects and keeping their distinguishing features.

### 6.9.4. Memory and Computational Efficiency

Increasing the network size of a CNN generally improves its learning capability but leads to problems regarding memory use and computational resources. These problems will be solved by hardware technology advancement. ASICs, FPGAs, and Eyeriss are some of the hardware accelerators proposed for reducing execution time and power consumption. Research in approximation models oriented to hardware also needs attention.

### 6.9.5. Hyperparameter Optimization

Deep CNNs have a large number of hyperparameters regarding options of activation functions, kernel sizes, numbers of neurons per layer, and layer order. Genetic algorithms perform the hyperparameter optimization both randomly and directly, with iterations improving previous results.

### 6.9.6. Pipeline Parallelism

Pipeline parallelism, on the other hand, can help tackle hardware limitations and further accelerate the training processes of complex CNNs. This technique improves scalability in training processes. Model parallelism in training can be easily implemented using GPipe, Google's distributed ML library. This can significantly speed up the training of large models and lead to higher performance overall with minimal changes in hyperparameters. Pipeline parallelism is used for developers and researchers to overcome the limitations of hardware and realize more from large CNNs.

### 6.9.7. Cloud Computing

With the increased complexity and computational needs of CNNs, cloud-based computing platforms are increasingly playing a vital role in their development process. Conventionally, deep and wide CNNs have been hindered in performance by the processing powers of single devices. This is where cloud computing becomes a very attractive alternative since it offers vast computational powers for data set processing. That translates into significant strides in efficiency that enable researchers and developers to train and deploy the CNN models in an economically viable way. These come with scalability, high-processing speeds, flexibility, etc., which suit the changing demand for CNN applications, and their large providers include Amazon, Microsoft, Google, and IBM.

### 6.9.8. Sequential Data

Though conventionally applied to image recognition, CNNs have lately been quite successfully adapted to sequential data. Such adaptation has been achieved by transforming 1-dimensional sequence data into a 2-dimensional format appropriate to be processed by CNNs. Indeed, the current popularity of 1D-CNNs in tasks involving sequential data is because they are well capable of capturing the important features in such data while retaining computational efficiency. This makes them particularly useful across various applications reliant on sequential data.

#### 6.9.9. High-Energy Physics

Recently, CERN physicists have leveraged CNNs to analyze particle collision data. Further growth in the domain of DL and especially the use of deep CNNs in high-energy physics, is expected to continue.

## 7. Fast Processing Techniques for CNNs

As CV and ML tasks grow more complex, deep neural networks increase in sophistication, necessitating substantial training data to prevent overfitting. The large volume of training data introduces challenges, particularly in training networks within reasonable timeframes. This section explores several rapid processing methods for CNNs.

### 7.1. Fast Fourier Transform

Researchers have explored using Fast Fourier Transforms (FFTs) for convolutions in the frequency domain, as detailed by [125]. One advantage is the ability to reuse Fourier transforms of filters across multiple images in a mini-batch, reducing computational load by avoiding redundant calculations. Recycling Fourier transforms of output gradients during backpropagation also boosts efficiency for both filters and input images. Efficiency is further enhanced by directly summing over input channels in the Fourier domain, minimizing the need for inverse transforms to once per output channel per image. GPU libraries like cuDNN and fbfft accelerate training and testing when employing FFTs for convolutions.

However, using FFTs requires additional memory to store feature maps in the Fourier domain, especially with stride parameters greater than 1. While effective for speeding up training and testing, specialized approaches are necessary for architectures like ResNet and GoogleNet, which use small convolutional filters. Winograd's minimal filtering algorithms improve inference efficiency by reducing channels in the transform space before applying the inverse transform.

### 7.2. Structured Transforms

Neural networks benefit from exploiting inherent redundancy within their structure. Low-rank matrix factorization is a key technique in achieving this, decomposing matrices into smaller ones to reduce storage complexity from $O(mn)$ to $O(r(m + n))$ (where m and n are matrix dimensions, and r is rank). This factorization also improves time complexity.

Low-rank matrix factorization was applied to the final weight layers in deep CNNs, leading to faster training times with minimal accuracy compromise [126]. Singular value decomposition was used per CNN layer to significantly reduce model size without accuracy loss [127]. Researchers focused on convolutional filter redundancies, developing approximation techniques to reduce computational complexity during training and inference [128] and [129]. Low-rank factorization was extended to multi-dimensional tensors using Tensor-Train decomposition to reduce parameters in fully connected layers [130]. The Adaptive Fastfood transform approximates matrices with improved space complexity ($O(n)$) and time complexity ($O(n \log n)$) [131]. Circulant matrices were explored for efficient computation and lower time complexity [132], while circulant structures combined with orthogonal Discrete Cosine Transform achieved similar efficiency gains.

### 7.3. Low Precision

CNNs use floating-point numbers for parameter updates, which can be inefficient due to redundant information. Binarized Neural Networks (BNNs) address this by restricting computations to binary values (0s and 1s). Two main approaches exist: full binarization applies binary values to all elements (input activations, weights, and output activations), while partial binarization focuses on one or two components.

The research explored full binarization, achieving high accuracy on MNIST with specialized operations like XNOR and bit counting [133]. XNOR-Net extended this approach to convolutional BNNs on ImageNet, achieving significant accuracy. DoReFa-Net reduced precision in both forward and backward passes while maintaining competitive accuracy. Training fully connected networks and CNNs with full binarization was proposed, showing competitive results across datasets [134].

### 7.4. Weight Compression

Various techniques can reduce parameters in convolutional and fully connected layers of neural networks. Vector quantization (VQ) compresses layers by exploiting parameter redundancy, as demonstrated by [135]. Pruning removes unimportant connections; methods like fine-grained sparsity by [30] and restoration by [136] exist. Coarse-grained methods, e.g., filter removal by [97] and filter merging by [44] also reduce parameters. Hashing techniques, such as HashedNets, group weights into hash buckets for size reduction while maintaining performance. Sparse Convolution introduces sparsity directly into convolutional layer weights, achieved through kernel redundancy exploitation [137]. Structured Sparsity Learning optimizes hyperparameters concurrently [138]. Lookup-based CNNs were proposed to encode convolutions via a trained dictionary, achieving high accuracy with fewer iterations than standard CNNs [139].

## 8. Emerging Research Domains

### 8.1. Image Classification

The influence of CNNs was huge, and it became a turning point in computer vision when AlexNet performed very well in the ImageNet competition in 2012. Nowadays, novel models, including ResNet, Inception, Xception, and EfficientNets, yield state-of-the-art performance on benchmark image datasets, reaching human-level performance or even better. This is possible because modern models update the principles of parameter efficiency, layer connection, and information flow for maximum accuracy per computational step. Consequentially, these CNNs are good at classifying various object categories with a very high accuracy. Besides, the application of CNNs extends beyond the static images themselves to the analysis of both spatial and temporal features of videos, thus enabling the differentiation of complicated actions and events, which is of great benefit in numerous applications.

### 8.2. Object Detection

Object detection in images is one of the fundamental challenges in the field of computer vision, and CNNs lie at the center of this attempt. There are mainly two approaches that prevail for object detection: two-stage detectors and one-stage detectors. Two-stage detectors, such as Faster R-CNN, work according to a sequential approach: first, it's the region proposal network proposing regions where objects may exist, and afterward, refinement by using predefined shapes called anchor boxes is done for an accurate classification of the detected object. Meanwhile, one-stage detectors, such as YOLO, perform the task in one go: it predicts both bounding boxes and object class probabilities in a single step without requiring a separate proposal stage.

Both detectors make use of prior knowledge about objects to improve training accuracy. Present efforts for improvement are being made to increase the speed and precision of such models so that complex models might be able to run in real-time. However, the use of these heavyweight models is significantly restricted by device computing capabilities. Given this challenge, research pioneers different techniques such as mobile object detection. These methods involve formulating low-weight CNN architectures that run efficiently on devices with meager computation resources.

### 8.3. Image Segmentation

Image segmentation is the process of assigning specific labels to each pixel, which requires an understanding of the image content in much greater detail. Fully Convolutional Networks or U-Net architectures utilize two important features for this purpose. First, they employ skip connections to directly access information in earlier stages of processing in a higher resolution, thereby preserving critical information on object locations across abstraction in feature analysis. Second, most of these models use the encoder-decoder structure: the encoder tries to extract the essential features from the image while the decoder infers a segmentation map using these features with detailed information to the original image.

The Researchers innovate by borrowing information from images across different scales. Pyramid pooling methods, as in PSPNet and DeepLab models, propose a pyramid architecture where each level captures an increasingly broader field of view of the image, helping to take into account the fine details from the overall contextual information. In addition, refinement methods are designed for high-accuracy tasks such as medical imaging. Techniques like GAN generate segmentations that normally get evaluated on precision by another network. In addition, conditional random fields use contextual information to enable even better refinement of segmentations. Segmentation techniques, on the other hand, show very strong implications in medical imaging. Segmentation CNNs make huge steps forward in the field of medical diagnosis and treatment planning while analyzing organ structures and locating abnormal ones.

### 8.4. ViTs

In conclusion, ViTs are the new paradigm in computer vision. They provide an equal yet robust alternative to the traditional CNNs on diverse image analysis tasks. The success of transformers in NLP inspires them, so they divide an entire image into smaller patches. Each patch is given a numerical representation and studied along with self-attention. This is because self-attention in the model lets it study relationships between diverse regions of the image, besides the adjacency constraints of CNNs. Besides achieving this, it enables the ViTs Transformer to strongly capture the long-range dependencies and contextual subtleties across an entire image. Several models, such as ViT, DeiT, and Visual BERT, have reported promising performance concerning tasks such as image classification, especially after pre-training over large datasets. However, active research has started to tune the transformers for efficiency. This optimization is critical for real-world, real-time computer vision applications that need very fast processing.

## 8.5. One-shot, Few-shot, and Zero-shot Learning

One-shot and few-shot learning methods represent techniques that also address the challenge of limited training data by leveraging metric learning and prototypical networks as core building blocks. Metric learning conveys the ability to extract informative features out of a comprehensive set of well-labeled examples and helps robust recognition of completely new concepts with minimal data. Prototypical networks complement this by forming a reliable representation of known concepts, which fosters effective comparison and categorization of objects even with a few instances. These methods also avoid "catastrophic forgetting": a model, when learning new information, may forget previously learned knowledge.

These approaches represent a leap forward in computer vision, where long-tailed learning, which is one of the forms of incremental learning of imbalanced datasets, has only a few training examples for a few object categories. This kind of learning enables the acquisition of new concepts continuously without restarting the model from scratch each time. These capabilities are further enriched by zero-shot learning, which empowers the model to identify completely new categories that the model did not observe in any training data via induction biases and descriptors such as attributes or semantic relationships. Techniques using SAE and DeViSE use these descriptors effectively for matching the seen and unseen categorical representations of the model. Knowledge graphs enrich learning from scarce data by encoding relations among objects and their properties, embedding structured information into the model's knowledge.

## 8.6. Weakly-supervised Learning

Standard ML models require volumes of data with detailed labels, something which is extremely expensive to create. Weakly-supervised learning, on the other hand, lets models learn from data that is not necessarily thoroughly labeled. In place of pixel-level annotations--painstakingly detailing every aspect of an image--a weakly-supervised model is trainable using generally weaker signals, such as image-level tags-labeled "cat"- or bounding boxes hastily drawn around objects.

One line of work, for example, multi-instance learning, which tries to pool regions of images that can refer to the same object from these weak labels. It does iterative refinement of each area to refine predictions. Another successful line of work combines multi-instance learning with a statistical expectation-maximization approach, usually called EM. Such joint usage allows the model to infer not only labels of the whole image or objects but also the localization of informative image regions supporting accurate classification.

## 8.7. Self-supervised and Unsupervised Learning

Self-supervised learning overcomes the problem brought about by scarce labeled data by leveraging abundant unlabeled datasets in pre-training model initiatives. In self-supervised learning, so-called "pretext tasks" are at the center of the action, which in turn should guide the model to learn meaningful visual features without explicit labels. Examples include the rotation of images and the prediction of how they originally looked, jigsaw puzzles created from image fragments, or counting specific colors. In such a pre-text task, the model will learn robust representations of visual information that can be helpful across various downstream tasks. Recently, this has been done with success by contrastive self-supervised learning methods represented, but not limited to SimCLR, SwAV, and MoCo. This model performs recognition tasks of images at a competitive or superior level to those obtained by their counterparts trained with labeled data. This capability not only enables more effective fine-tuning using less labeled data but also allows transfer to purely new visual tasks.

### 8.8. Lifelong and Continual Learning

Continuous or lifelong learning is a ML paradigm that deals with developing the capability of AI models to learn incessantly, unrestricted by static datasets, much like the human brain, during their lifetimes. These systems adapt to the non-stationarity of data-that is, data changing over time avoiding the catastrophic forgetting of knowledge acquired earlier. Several techniques are, therefore, being explored in this regard. EWC and Incremental Moment Matching Regularization avoids forgetting previously learned knowledge by changing model parameters only selectively to make room for learning new information. Task-aware architectures make changes in internal structure to accommodate the different learning tasks that may incorporate new information without interference with previously learned knowledge. Dual-memory systems follow the human brain in terms of the line between short-term and long-term memories. While it places higher priority on the most recent information in its short-term memory, the model still retains a good deal of key knowledge from its past in its long-term memory. Buffer replay systems store previous data in a buffer from which to replay a subset of that data into the model to relearn and reinforce past knowledge whenever necessary to avoid forgetting. These techniques enable lifelong visual learning in AI systems, acquiring the competency to improve their understanding of the visual world incrementally with time as novel information will be encountered and adapted to.

### 8.9. Vision-Language Models

VLMs are a new class of AI models that bridge the gap between visually perceived and linguistically expressed content. The model integrates NLP with computer vision into an architecture that is capable of analyzing images and text concurrently to make sense of the visual world from a linguistic perspective. Attention mechanisms in VLMs allow them to selectively focus on those parts of an image relevant to the surrounding text, and vice-versa. As a result, one can perform tasks such as accurately captioning an image or identifying objects in an image using textual descriptions. Very recently, large pre-trained VLMs such as CLIP, ALIGN, and Oscar have been proposed. These models are very good at performing many tasks, without special training regarding those tasks-zero-shot learning: classifying images, answering questions about images, and holding conversations about the content of images. Its applications could vary from reforming learning at schools with virtual, interactive classrooms to the ability to enable people who have poor eyesight to better interact with the physical world.

### 8.10. Medical Image Analysis

DL radically changes medical imaging in collaboration with clinical experts, ranging over different applications such as accurate organ segmentation in 3D scans, the detection of abnormalities in several X-rays, MRI modalities, and continuous monitoring of patients by taking sequential scans. CNNs are of great importance when it comes to handling large image datasets, both 2D and 3D versions. These sophisticated models represent this insight by taking as input anatomical priors from doctors and anatomists and making accurate predictions, preserving sharp edges and mirroring structures in reality seamlessly.

Above all, self-supervised learning allows the first training of these models on large anonymized medical datasets. That will create a very strong foundation for later fine-tuning targeted at specific medical tasks. However, another key aspect is the interpretability of model predictions to obtain confidence from the medical professional community. This would be great for fostering confidence and acceptance of the insights derived through AI. Besides diagnostic applications, CNNs provide new views of medical scans that prove very useful in surgical planning for complex cases. Also, efficient model development is extremely important in terms of model deployment on medical devices or resource-constrained settings, thereby allowing quicker diagnosis and access to more advanced medical imaging technologies.

### 8.11. Video Understanding

Understanding video content requires an analysis of both the spatial relationships between different parts of the video and the changes that take place over time in consecutive frames. Traditional CNN architectures for this purpose, such as C3D and I3D, make use of 3D convolutions, which learn patterns from whole video clips treated as volumes. Recent breakthroughs in video captioning and action recognition go beyond simple image analysis, including language models and mechanisms of attention. The language model builds descriptive captions that can interpret visual content, while the attention mechanisms permit the models to concentrate on certain frames of the video which are important for understanding either the performed action or generating a correct caption. Apart from these, the self-supervised ways of training on large collections of videos without labels have also been very promising. This pretrains the models by recognizing general patterns in videos without explicit labels of each action/scene. These can be fine-tuned subsequently for classification or activity detection in videos.

### 8.12. Multi-Task Learning

Multitask learning with CNNs improves performance by making use of the shared representations between related tasks in a way that benefits from limited data due to reduced overfitting. For example, YOLO combines the tasks related to object segmentation and counting along with detection. It has improved its overall accuracy. With careful architecture design, balance within task relationships is always kept between independent and full cooperation for robust performance and real-world applicability. Integration of CNNs with models such as RNNs for multi-modal tasks further optimizes performance by sharing feature learning and pre-training on unlabeled data using self-supervised methods.

### 8.13. 3D Vision

Estimating the full 3D pose-place and orientation of an object from a single ordinary image is considered a very challenging task since flat images inherently lose their depth information. Early approaches relied on computer-generated data, lacking accuracy in the real world. Recent work uses large datasets of real-world images and CNNs that directly predict object poses. The PoseCNN, trained on real-world data, has performed better than traditional approaches.

A key challenge is how to arrive at consistent pose predictions irrespective of object orientation. Various loss functions, including reprojection and angular loss, have been involved in mitigating this challenge. Most methods improve the pose estimates sequentially: first by detecting objects in images and then by refining pose estimates by matching image features to projected 3D model points. Methods such as DeepIM predict object shape details and iteratively refine the poses using known algorithms.

Such dense representations enable these methods to process object parts independently, while integration from multiple cameras or depth sensors increases accuracy. For instance, MVD trains separate networks for each camera view and then combines the results for robust predictions. Depth-PoseNet fuses the RGB and depth for converting 2D predictions into accurate 3D poses. All these are being trained on multiple tasks simultaneously-including bounding boxes, key points, and pose-reaching accuracies so far possible only with marker-based motion capture systems.

### 8.14. Neural Architecture Search

Neural architecture search (NAS) eliminates the need for the manual design of CNNs. NAS utilizes advanced algorithms to autonomously create optimal architectures tailored to specific tasks, allowing human experts to concentrate on other facets of the ML process.

## 9. CNN-Powered for Image Classification, Segmentation, and Detection

### 9.1. CNN Model for Image Classification

#### 9.1.1. Ensemble CNN

The challenge of classifying the pulmonary peri-fissural nodules was addressed by taking a new approach to ensemble learning. A random forest model and OverFeat CNN were combined into one model to enhance performance classification. To this effect, features critical for nodule classification were achieved while reducing data complexity by extracting 2D slices from 3D medical images. The developed approach showed an AUC of 86.8%, indicating very good performance in nodule classification versus normal lung tissue. This application further underlines that a combination of ensemble learning and 2D data analysis may offer great potential for precise pulmonary nodule classification, which, in the future, will be very important for early detection and diagnosis.

In addition, the researchers adapted AlexNet, a pre-trained model of a DL network, and used it to classify the masses of the breast in mammograms with an accuracy of 96.7%. The mentioned adaptation showed how a pre-trained model can be put to work on medical image analysis to help with diagnosis of the breast cancer. Another approach consisted of the DL locality-sensitive method for histopathology images, which depended on small patches of tissues for the detection and classification of cell nuclei. This resulted in a very effective AUC of 91.7% and an F-score of 78.4%. By concentrating on particular areas of the images, this strategy raises the efficiency and precision of the processing and, therefore, is promising in the automatic analysis of histopathology images.

#### 9.1.2. Small-Kernel CNN

Presently, the researchers innovated a medical image-based technique that introduced a new classification scheme categorizing the lung tissue into seven classes with an accuracy as high as 85%, hence beating the benchmarks set by well-known DL architectures like AlexNet and VGGNet. A great improvement can be considered in this technique, where small kernel sizes of 2x2 are used in convolutional layers. Conventional

CNNs use larger kernels for capturing high-level features. However, this process combines the low-level textural information into it. Textural information is related to the repetitive patterns in image regions, which plays an important role in distinguishing different types of lung tissue. This may be done effectively with a smaller size kernel. Also, nonlinear activation functions have been incorporated to add more nonlinearity to the network since complex relations between features of an image and classes of tissue can be learned easily. With this proposed combination of smaller kernel sizes and specialized activation functions, greater accuracy and speed in image processing are achieved-classification takes just 20 seconds. These relative efficiencies enable this model to position itself for practical medical deployment. Generally, the study by this paper underlines the need to balance high-level and low-level features in medical image classification and calls for smaller kernels and tailored activation functions that improve performance and efficiency.

#### 9.1.3. Whole Image CNN

The researchers have advanced the segmentation of lung tissue in medical scans through the use of smaller image patches to maintain fine details and the inclusion of multiple ranges of lung attenuation in an AlexNet model modified for the task. It achieved a perfect F-score of 100% and managed to keep an average accuracy of 87.9%, thus proving efficient in carrying out medical image analysis with accuracy. Results like these point toward improved diagnostic and treatment-planning tools for a variety of conditions.

#### 9.1.4. Multicrop Pooling CNN

With the picture classification problem, the challenges of limited training data are attempted to be coped with using an innovative approach by researchers. The work proposes a three-layer CNN model with an integrated newly devised multicrop pooling. Multicrop pooling refers to the analysis of random sections of variable size within the image instead of taking up the whole image at once for analysis. This also allows the model to capture more of the scale of information and be resilient to small variances with much fewer data. Besides, the model incorporates a special kind of activation "randomized leaky ReLU" throughout its layers. In other words, by utilizing this function, the model prevents matching particular patterns in the small dataset and helps the model generate more various features. Using combined techniques, the model yielded impressive results an accuracy of 87.4% and an AUC of 93%. This study shows great promise for multi-crop pooling and specialized activation functions in the improvement of CNN performance, especially where there is a limitation or cost in acquiring large data sets.

#### 9.1.5. Bayesian CNN

The authors investigated the application of the state-of-the-art methodology in image classification based on a CNN coupled with Monte Carlo Dropout for Bayesian approximation. This would, in essence, be able to let the model be informed of the uncertainty about its predictions consideration fairly important within medical diagnosis. The current research focuses on the development of ResNet50v2 right before the final classification layer, softmax, and including MCD into the model. This adaptation allowed the model not only to predict most of the probable classes an image belongs to but also to quantify its confidence in the prediction. This estimation of uncertainty is useful for many reasons: it will enable healthcare providers to gain more insight from the model's line of reasoning and will help them reach a more diagnostic decision. Hence, results were found better than those realized by traditional CNN. The approach has proved that the inclusion of MCD in CNN enhances the diagnostic performances and reliabilities in general and it has a promising application in medical issues.

#### 9.1.6. CVR-Net

The newly proposed CNN architecture attempts to overcome the difficulties introduced when dealing with small datasets of images during classification. As explained, a multi-scale multi-encoder ensemble attempts to handle the challenge by two key methods: First, the model uses multiscale feature extraction from images. That means, instead of looking into an image and focusing on one particular scale example, trying to take in the panoramic view versus noticing a flower in the foreground model extracts both minute details and broader context out of an image. Second, the architecture involves multiple CNN encoders, probably trained over images at different scales each. Aggregating the predictions of these encoders leverages their various strengths and leads to a stronger, more accurate classification result from the model. This idea proved to be very efficient and gave about 98% on different metrics of performance. Its success is therefore underlined as a potential to overcome the limitation brought about by small datasets and enhance the effectiveness of CNNs in the image classification task. Such steps are beneficial, especially in scenes where acquiring extended datasets happens to be difficult or expensive.

#### 9.1.7. Twice TL CNN

Recently, there was a breakthrough in detecting COVID-19 using chest X-ray by the researchers who proposed a well-planned approach using the DenseNet model. First of all, the proposed model was pre-trained on the large dataset of ImageNet, resulting in immense knowledge regarding how to identify images belonging to a wide range of classes. This basic knowledge helped it catch the general visual appearance quite well. Then, training on the Chest X-ray 14 dataset increased the model's familiarity with chest anatomy and normal visual patterns of lung pathologies. Further training in specialization for COVID-19 was done by fine-tuning the model on the dataset comprised exclusively of chest X-rays from COVID-19 patients. This crucial fine-tuning empowered the model to show the minute signs suggestive of COVID-19 infection. Through this multi-stage process of training, the accuracy of the DenseNet model reached an incredible 98.9% in COVID-19 case detection. The present study underlines how TL works pre-trained models develop proprietary expertise by adapting to the targeted dataset. Thus, such approaches hold tremendous promise for improving the accuracy of DL models associated with various medical image analyses.

#### 9.1.8. CUDA ConvNet CNN

The researchers achieved a high gain in image classification performance: 80.3%, significantly outpacing benchmarks. They followed a multistep method wherein optimization of training data ranked first. They equalized the histograms so that the contrast between images would increase, therefore normalizing the distribution of pixel intensity and thereby making the images more distinguishable. Then they applied data augmentation to artificially increase the dataset, so the model would see a single image many times but with several variations. Therefore, the model will learn invariant features and be insensitive to small variations in real data.

Researchers extended this to apply the concept of TL by fine-tuning the pre-trained model on the preprocessed and augmented dataset. TL is a concept used when taking a model pre-trained on a very large dataset and tuning it to work for a particular task, whether medical or satellite image classification. Thus, an integrated approach-data preparation followed by TL has proved to significantly improve the performance of models targeting image classification tasks.

**9.1.9. Six-Layer CNN**

The HEp-2 cell image classification by these researchers was significantly advanced and is a necessity in the diagnosis of autoimmune diseases. The approach followed the classified quality improvement of the training data. Thus, after applying noise reduction and background removal as preprocessing techniques, complemented by rotations, flips, and scaling for data augmentation, they were able to achieve 96.7% average accuracy in the classification on the ICPR-2012 benchmark dataset.

Preprocessing normalized the quality of the images, removing noise and background. The further data augmentation techniques increased the variety in the dataset, which allowed the model to learn more features and increase robustness against variations that could occur in real-world images of HEp-2 cells. This work confirms that careful data preprocessing is an important step in image classification for accurate performance, especially in medical applications.

**9.1.10. Semi-Supervised CNN**

After that, the scientists proceeded to investigate a graph-based semi-supervised learning strategy in a scenario of very few labeled data. This approach leverages the relationships among the data points to enhance the accuracy of the classification model when only a subset of data is labeled. Results: The proposed approach yielded 88% AUC and 82% accuracy. This outperformance has been compared favorably with traditional DL models, where the latter faced similar constraints of a limited amount of labeled data. This methodology is useful in scenarios that need data labeling to be expensive or time-consuming. The present graph-based method of semi-supervised learning relies on the intrinsic structure in a dataset to make effective use of both labeled and unlabeled data points and, hence, boost performance beyond what is achievable by a sparsely labeled dataset.

**9.1.11. One-Dimensional CNN**

The new approach of the researchers in ECG classification using a one-dimensional deep CNN was proposed. This model unifies feature extraction and classification into one framework, where these two features are separated into traditional approaches. The 1D CNN employed learning directly from ECG data with a high accuracy of 95%, thus classifying the ECG signals. Performance of the 1D CNN thus constitutes evidence of its ability to uncover meaningful patterns from ECG signals and thus hopefully allow for an efficient, perhaps more accurate analysis of heart conditions in clinical practice.

**9.1.12. Fused CNN**

The researchers proposed a fused CNN architecture for echocardiography video classification that includes both spatial information from the video frames and temporal data describing the tracking of the heart movement. It yielded 92% accuracy in classifying videos, which is higher than that obtained by state-of-the-art single-path CNN techniques. Integrating both spatial and temporal information could offer an overall review of cardiac functioning for far better diagnosis. This current research work acts as a new direction toward the development of integrated CNN architectures for medical video analysis, majorly echocardiography, toward the treatment of patients.

Another one looked for classifying AD with a combined framework of 2D and 3D CNNs. It yielded accuracy in identifying AD at 86.7%, the detection of lesions at 78.9%, and classifying a healthy brain scan at 95.6%. The approach makes full use of the advantages of both types of CNN: 2D CNN captures the intraslice spatial relations in MRI scans, while 3D CNN analyzes the whole brain volume for complicated features. The overall representation from both the CNNs puts together a comprehensive understanding that improves the accuracy of classification across AD, lesions, and healthy brain scans. This work justifies the power of combined 2D and 3D CNN architecture to advance medical image analyses with specific diagnostics in Alzheimer's disease.

### 9.1.13. Gaussian Initialized CNN

It proposes a Gaussian initialization along with weighted class weights for the classification of DR from the given retinal fundus images. The proposed approach was able to identify healthy eyes with 95% specificity, whereas, for the detection of cases, it resulted in as low as 30% sensitivity.

It is called Gaussian initialization and normally refers to initializing weights in DL models to enhance training efficiency. Weighted class weights relate to associating different importance levels on categories during the model training, normally giving a higher priority to the less frequent classes like DR.

The low sensitivity would tend to indicate that the model prefers to avoid false positives by mislabeling a healthy eye as having DR. The weighting strategy or further optimization of the model will be important in striking a better balance between correctly classifying negative and positive DR cases. This fine-tuning will, therefore, be necessary to allow for early detection and treatment of diabetic retinopathy.

### 9.1.14. Hyperparameter Tuning Inception-v4

Researchers have proposed DL-based locality-sensitive methods for the analysis of histopathological images. The authors demonstrated how to effectively train the model based on small patches sampled from whole tissue images since the area around each patch is enough for nuclei detection and classification. It achieved outstanding performance metrics: 91.7% AUC and 78.4% F-score; thus, proving to be effective in nucleus detection and classification.

Some of the benefits of the locality-sensitive approach are outlined below: It enables the efficient processing of big histopathology images in high resolution by focusing the computational effort in small patches. The model will, consequently, be in a better position to contextualize those specific features relevant to the classification of nuclei, therefore giving higher general accuracy. This study proposes that the integration of locality-sensitive DL methods is among the promises of automation for some tasks in histopathological image analysis, such as the detection and classification of nuclei.

## 9.2. CNN-based Segmentation

### 9.2.1. Small Kernel CNN

Researchers studied the segmentation of gliomas, representing brain tumors, using a medical image by adopting a new strategy. This work tries to balance the advantages of deep neural network architectures with the requirement for consistent receptive fields. The receptive field refers to an image region that every neuron in the CNN would look at for any given prediction. They were able to do this using small filters during patch-wise training, size 3x3. This means that patch-wise training is done by dividing the image into small patches and then training the model independently on those patches. Thereby, this strategy allows the model to learn small and detailed information about gliomas while maintaining similar receptive fields throughout the network.

For high-grade and low-grade gliomas, the model architecture was adapted differently. The segmentation model of high-grade glioma consisted of eight convolutional layers and three dense layers, while four convolutional layers with three dense layers formed the model for low-grade gliomas. Max pooling was also used for reducing dimensionality in this approach, dropout techniques to avoid overfitting, and data augmentation methods so it could artificially increase the size of the training dataset.

This approach indeed turned in promising results, allowing it to achieve the rank of fourth place in the BRATS-2015 challenge high-ranking benchmark for algorithms to segment brain tumors. These findings emphasize the fact that one of the efficient ways to approach glioma segmentation in medical images is by utilizing small filters together with patch-wise training.

### 9.2.2. Fully Volumetric CNN

Researchers studied the new segmentation of subcortical structures in the brain using fully volumetric MR images that capture highly detailed 3D views of an entire brain. To refine the accuracy of the segmentation, they integrated a Markov random field into their model. MRF is a statistical tool that imposes consistency among neighboring regions within an image. In the work, MRF ensured that volumetric homogeneity was present in the segmentation output from CNNs, which was responsible for yielding realistic 3D shapes and sizes of the segmented structures. The current approach outperforms a few existing state-of-the-art techniques in accurately segmenting sub-cortical brain structures from volumetric MRI datasets.

### 9.2.3. Cascaded CNN

Researchers proposed an imbalanced label distribution in the tasks of brain segmentation using a two-phase training approach with a two-pathway CNN architecture that enables the simultaneous learning of global contextual features and finer-grained local details. This scenario is particularly suitable for tackling issues where some types of brain tissues are underrepresented in the training data. Generally speaking, the imbalanced label often leads to model preferences toward more frequent classes while compromising the overall segmentation accuracy. The two-stage strategy most probably started from the initial self-balancing stage to balance learning across all tissue types followed by the refinement stage to fine-tune for increasing accuracy of segmentation. This approach outperformed the usual two-pathway CNN to a large degree, winning major successes, including second place in the MICCAI BRATS-2013 challenge with significant increases in almost all metrics of segmentation. These results confirm that the proposed two-phase training approach is effective in alleviating the problem of label imbalance in improving the accuracy of segmentation tasks.

#### 9.2.4. Multipath CNN

In this regard, researchers have proposed a new approach for automatic MS lesion segmentation in medical images. This was introducing an automated dual-pathway CNN. For feature extraction, a convolutional pathway was used. The model uses a deconvolutional pathway for refining the lesion boundaries. The proposed CNN model first pre-trained the weights in the convolutional pathway using CRBM. It speeded up the training process, and improved GPU optimization during both training and image analysis. It presented a very high true positive rate with low false positives. However, when it came to DSC, the overall performances were relatively low, which implied that boundary delineation needed an upgrade. Weight for DSC improvement may be given by further research efforts toward better segmentation accuracy.

#### 9.2.5. Multiscale CNN

The proposed multiscale CNN architecture captures the local details as well as wider contextual information in brain tumor segmentation. It comprises three different CNNs that operate at three different scales of images. This kind of multiscale approach lets the model learn small regions of complex features and greater patterns across an entire brain scan with an accuracy of roughly 90%. This hereby presents the capability of a multiscale CNN in enhancing the precision and consistency of segmentations about brain tumors.

#### 9.2.6. Multipath and Multiscale CNN

Compared to brain lesion segmentation, two methods were adopted for two-path and multiscale CNNs. The two-path design uses dual parallel convolution pathways to process information at multiple scales. The model generates reasonably accurate boundary maps around the brain lesion segmentation. It performed robustly by getting high accuracy in segmenting lesions within seriously injured brain trauma patients and also ranked highly in stroke lesions with the challenge ISLES-SISS-2015. The proposed architectures proved to be highly accurate both for two-path and multiscale architectures and thus held great promise for medical imaging.

#### 9.2.7. FCNN

Researchers presented a fast-sweep deep CNN for the segmentation of breast cancer tissue in high-resolution histopathological images. The developed FCNN demonstrated efficient performance with an F-score accuracy of 85%, accomplishing the segmentation of big images (1000×1000 pixels) within 2.3 seconds. More importantly, the FCNN proved to be robust against intra-class variation, turning out to be effective on a wide range of tissue samples within the class of cancer. This is important in real-world applications where the manifestations of cancer can vary significantly. With speed and accuracy, the FCNN has been proven to be a promising tool for pathologists who analyze biopsies for breast cancer. Automation of the segmentation process could reduce the time required for analysis and perhaps improve diagnostic consistency.

#### 9.2.8. Probability Map CNN

A new approach was followed for the automation of cell nuclei segmentation in images of human breast cancer. It trains a CNN on a probability map so that every pixel has some probability of belonging to a cell nucleus. In this mapping, CNN significantly outperformed traditional ML approaches like SVM, RF, and DBN by a great margin in terms of precision, recall, and F-score. These metrics provide an evaluation of the model's precision in identifying cell nuclei within the images. This work depicts the probability map-based CNNs that can lead to an increase in the accuracy and speed of nucleus segmentation-related tasks, especially when considering the analysis of breast cancer.

### 9.2.9. Patch CNN

A new approach for handling histopathological images for the classification of breast and colorectal cancers was performed. The results found 100% accuracy with a higher AUC than the previous CNN-based techniques. This technique covered two approaches: patch-based CNNs and superpixel techniques. Patch-based CNNs focus on the small patches of the image of the tissue with an emphasis on recognizing the details. Superpixel techniques group similar pixels to obtain comprehensive information about the structure of the tissue. This merged the benefits for superior performance in cancer classification and proved the potential of patch-based CNNs combined with superpixel techniques in cancer diagnosis.

### 9.2.10. Greedy CNN

The researchers have proposed an improved architecture of CNN in detecting glaucoma from images of the retinal fundus by including sequential learning and greedy boosting. Greedy boosting is an optimization technique that selects the training examples in steps to minimize weighted classification errors. This algorithm improves the accuracy of segmentation for the optic cup and the optic disc in the retinal images. Both these features are informative for the diagnosis of glaucoma. Hence, the study indicated that there is the possibility of modification to CNN architecture by techniques like greedy boosting for improving tasks related to medical image analysis.

### 9.2.11. Multi-label Inference CNN

Segmentation of the retinal blood vessel was treated by researchers as a multi-label inference problem. The authors report their model to have attained high precision, sensitivity, specificity, accuracy, and AUC. Most importantly, the model focused on the green channel of RGB fundus images. This implies that the green channel is really important for segmenting blood vessels. This is an important realization that may direct the development of other models targeted at blood vessel segmentation.

### 9.2.12. U-Net

They assessed the performance of a U-net architecture for segmenting the lungs and excluding bone shadows in the analysis of lung cancer. Results using this approach showed fast processing speeds and great accuracy of lung segmentations in chest X-ray images. The U-net architecture captures contextual information more effectively, thus proving to be better for medical image segmentation tasks. This could indicate that models based on U-net might become an important tool in the analysis of lung cancer and thus possibly enable rapid and more accurate diagnoses.

## 9.3. CNNs based Object Detection

### 9.3.1. GoogLeNet CNN

These different award-winning systems attained performances close to that of human observers by relying on patch-based classification combined with extensive training based on misclassified patches. One of them reached an AUC of 92.5% in the Camelyon Grand Challenge.

### 9.3.2. Dynamic CNN

Researchers proposed a dynamic CNN for hemorrhage detection in retinal fundus images, presenting very promising results: sensitivity-93%, specificity-91.5%, and ROC AUC-98% for the detection of hemorrhages. The dynamic training sample selection is done by CNN through strategic selection, where either the most informative or most challenging examples would be selected to focus on its learning. Further, Gaussian filters are introduced within the network during training for either the preprocessing of the fundus images or the enhancement of important features inside the CNN structure. This work investigates dynamic CNNs using techniques such as dynamic sample selection and Gaussian filters that show their potential to deliver high accuracy in hemorrhage detection tasks concerning retinal fundus images.

## 10. CNN Advancements

| Adv. | Technique | Functionality | Impact | Formula |
|---|---|---|---|---|
| Transposed Convolutions and | | ▪ Upsampling Feature Maps<br>▪ Control of Upsampling<br>▪ Upscaling Low-Resolution Images<br>▪ Control over Output Resolution<br>▪ Artifact Reduction | • Enhanced Image Resolution<br>• Reduction of Artifacts<br>• Versatility in Image Processing<br>• Improved Visual Appearance | |
| Depthwise separable Convolution | | • Two-Step Convolution Process<br>• Depthwise Convolutions<br>• Pointwise Convolutions<br>• Efficient Spatial Pattern Capture<br>• Resource-Constrained Environments | • Parameter and Computation Reduction<br>• Suitability for Mobile and Embedded Applications<br>• Enabling Advanced CNN Architectures<br>• Wide Adoption<br>• Model Size Reduction | |
| Spatial Pyramid Pooling | | • Handling Variable Input Sizes<br>• Region-Based Pooling<br>• Feature Map Division<br>• Fixed-Length Representation<br>• Compatibility with Fully Connected Layers<br>• Consistent Feature Extraction | • Consistent Feature Representation<br>• Versatility in Applications<br>• Improved Adaptability<br>• Enhanced Performance | |
| Attention Mechanisms in Convolutions | | • Focused Feature Extraction<br>• Spatial Weight Assignment<br>• Selective Focus on Relevant Parts<br>• Adaptation from Transformers<br>• Capture of Long-Range Dependencies | • Enhanced Pattern Recognition<br>• Adaptability with Convolutional Networks<br>• Enhanced Relevance<br>• Contextual Understanding<br>• Increased Model Capability | |
| Shift-Invariant and Steerable Convolutions | | • Consistency in Feature Learning<br>• Orientation-Sensitive Feature Learning<br>• Shift-Invariant Convolutions<br>• Steerable Convolutions<br>• Orientation Sensitivity | • Robust Object Detection<br>• Adaptability in Text Recognition<br>• Enhanced Text Recognition<br>• Versatility | |
| Recent Advancements And Innovations | Capsule Network | • Capsules as Fundamental Units<br>• Dynamic Routing Mechanism<br>• Addressing Limitations<br>• Enhanced Representation<br>• Robustness to Transformations | • Robust Object Recognition<br>• Advanced Representation of Entities<br>• Revolutionary Advancement<br>• Improved Hierarchical Modeling<br>• Enhanced Robustness<br>• Broad Applicability | |
| | Neural Architecture Search For Convolutions | • Automated Architecture Design<br>• Exploration of Convolutional Designs<br>• Evaluation and Optimization<br>• Tailored Architectures | • Development of Advanced CNNs<br>• Optimization for Specific Applications<br>• Improvement in Autonomous Navigation | |
| | Generative Adversarial Networks | • Generative Model Components<br>• Adversarial Training | • Advancements in Image Generation<br>• Extension to Various Domains<br>• Data Augmentation<br>• Realism in Generated Images | |
| ViTs and self-attention Mechanisms | | • Segmentation into Patches<br>• Use of Self-Attention | • Evolution of Computer Vision<br>• Advancement in CNN Capabilities<br>• Promising Future Developments<br>• Conceptual Divergence | |
| Convolution Layer | Tiled Convolution | • Weight Sharing<br>• Learning invariances<br>• Separate kernel<br>• Square Root Pooling | • Enhanced Feature Learning<br>• Optimal performance<br>• Improved results on the small dataset | |
| | Transposed Convolution | • Backward Pass of Convolution<br>• Upsampling<br>• Multiple Output Activations | • Visualization<br>• Recognition and Localization<br>• Semantic Segmentation<br>• Visual Question Answering (VQA)<br>• Super-Resolution | |
| | Dilated Convolution | • Hyper-parameter Introduction<br>• Receptive Field Expansion<br>• Dilated Convolution Operation | • Large Receptive Fields for Prediction<br>• Exponential Growth of Receptive Field Size<br>• Performance in Various Tasks<br>▪ Scene Segmentation<br>▪ Machine Translation<br>▪ Speech Synthesis and Recognition | |
| | Network in Network | • Micro Network Replacement<br>• Stacking Micro Networks<br>• Enhanced Feature Maps<br>• 1×1 Convolutions<br>• Global Average Pooling | • Abstract Representation Learning<br>• Reduction in Overfitting<br>• Improved Performance | $a_{i,j,k} = \max(\mathrm{w}\frac{T}{k} X_{i,j} + b_k, 0)$<br><br>$a^n_{i,j,k_n} = \max(w^T_{k_n} a^{n-1}_{i,j,:} + b_{k_n,}, 0)$ |
| | Inception Module | • Variable Filter Sizes<br>• Dimension Reduction<br>• Pooling and Convolution Operations<br>• Depth and Width Increase | • Parameter Efficiency<br>• Performance Optimization<br>• Training Acceleration<br>• High Accuracy | |
| Pooling Layer | Lp Pooling | • Generalized Pooling Operation<br>• Adaptability to Different Pooling Strategies | • Better Generalization<br>• Flexibility in Feature Aggregation<br>• Versatility | $y_{i,j,k} = [\sum_{(m,n)\epsilon R_{ij}} (a_{m,n,k})^p ]^{1/}$ |
| | Mixed Pooling | • Combination of Max and Average Pooling<br>• Random Selection<br>• Training Process | • Reduction of Overfitting<br>• Improved Performance<br>• Backpropagation Consistency | $y_{i,j,k} = \lambda_{(m,n)\epsilon R_{ij}}^{max} a_{m,n,k} + (1 - \lambda) \frac{1}{\mathrm{I} R_{ij} \mathrm{I}} \sum_{(m,n)\epsilon R_{ij}} a_{m,n}$ |
| | Stochastic Pooling | • Random Activation Selection<br>• Probability Computation<br>• Sampling from Multinomial Distribution | • Reduction of Overfitting<br>• Utilization of Non-Maximal Activations<br>• Regularization Effect | $p_i = \frac{a_i}{\sum_{k \in R_j}.}(a_k)$ |
| | Spectral Pooling | • Dimensionality Reduction<br>• Process Steps<br>▪ DFT Computation<br>▪ Frequency Cropping<br>▪ Inverse DFT Application | • Information Preservation<br>• Output Dimensionality<br>• Computational Efficiency | |
| | Pyramid Spatial Pooling | • Fixed-Length Representation<br>• Pooling Local Spatial Bins<br>• Fixed Number of Bins<br>• Contrast with Sliding Window Pooling | • Size Invariance<br>• SPP-net (process images of different sizes without resizing)<br>• Versatility in Handling Image Sizes | |
| | Multiscale orderless Pooling | • Deep Activation Feature Extraction<br>• Global Spatial Layout<br>• Local Patch Aggregation | • Enhanced Invariance<br>• Fine-Grained Detail Capture<br>• New Image Representation | |
| Activation Function | Relu | • Non-saturated Activation<br>• Piecewise Linear Function<br>• Sparsity Induction<br>• Efficient Computation | • Training Efficiency<br>• Empirical Performance<br>• Computational Speed<br>• Induced Sparsity | $a_{i,j,k} = \max(z_{i,j,k}, 0)$ |
| | LRelu | • Gradient Preservation<br>• Parameter ($\lambda$): (A predefined constant in the range (0, 1), representing a small positive slope for the negative part of the function.) | • Solves Dying ReLU Problem<br>• Ensure Continuous Gradient Flow<br>• Accelerates training.<br>• Prevents Zero Gradients<br>• Promotes Faster Convergence<br>• Enhanced Robustness | $a_{i,j,k} = \max(z_{i,j,k}, 0) + \lambda \min(z_{i,j,k}, 0)$ |
| | PRelu | • Adaptive Parameter Learning<br>• Channel-Specific Parameter | • Improved Accuracy<br>• Low Risk of Overfitting<br>• Negligible Computational Cost<br>• Backpropagation Training | $a_{i,j,k} = \max(z_{i,j,k}, 0) + \lambda_k \min(z_{i,j,k}, 0)$ |
| | RRelu | • Randomized Parameters<br>• Randomized Nature | • Reduction of Overfitting<br>• Consistent Performance Improvement | $a_{i,j,k}^{(n)} = \max(z_{i,j,k}^{(n)}, 0) + \lambda_k^{(n)} \min(z_{i,j,k}^{(n)}, 0)$ |
| | Elu | • Positive Identity<br>• Negative Part Incorporation<br>• Saturation Function<br>• Parameter ($\lambda$): (A predefined parameter that controls the saturation level for negative inputs.) | • Faster Learning<br>• Higher Classification Accuracy<br>• Robustness to Noise<br>• Avoids Vanishing Gradient Problem | $a_{i,j,k} = \max(z_{i,j,k}, 0) + \min(\lambda(e^{z_{i,j,k}} - 1), 0)$ |
| | Maxout | • Feature Map Channels<br>• Suitability for Dropout<br>• Maximum Response | • Flexibility<br>• Compatibility with Dropout<br>• Regularization Support | $a_{i,j,k} = \max_{k\in[1,K]} z_{i,j,k}$ |
| | Probout | • Probabilistic Sampling<br>• Multinomial Distribution<br>• Probabilistic Variant of Maxout | • Balanced Properties: (maintains the beneficial properties of max out while enhancing invariance properties.<br>• Computational Consideration.<br>• Incorporation with Dropout | $p_i = \frac{e^{\lambda z_i}}{\sum_{j=1}^{k}. e^{\lambda z_j}}$<br>$\hat{p}_0 = 0.5, \hat{p}_i = \frac{e^{\lambda z_i}}{2\sum_{j=1}^{k}. e^{\lambda z_j}}$<br>$a_i = \begin{cases} 0 & if\ i = 0 \\ z_i & else \end{cases}$ |
| Loss Function | Hinge loss | • Large Margin Classification<br>• Multi-Class SVM<br>• Correct Class Labels | • L1-Loss and L2-Loss<br>• Differentiability: (The L2-Loss is differentiable.)<br>• Penalty for Misclassification<br>• Comparison with Softmax: (Studies comparing L2-SVMs with softmax activation in deep networks have shown that L2-SVMs can lead to superior performance, as demonstrated on datasets like MNIST.) | $L_{\text{hinge}} = \frac{1}{N}\sum_{i=1}^{N}.\sum_{j=1}^{K}[\max(0,1 - \delta(y^{(i)}, j)(w^T x_i))^p]$ |
| | Softmax loss | • Combination of Losses<br>• Probability Distribution<br>• Loss Calculation | • Large-Margin Softmax (L-Softmax)<br>• Margin Adjustment<br>• Empirical Validation | $L_{\text{softmax}} = -\frac{1}{N}[\sum_{i=1}^{N}.\sum_{j=1}^{K}.1\{y^{(i)} = j\}\log p_j^{(i)}]$<br>$p_j^{(i)} = \frac{e^{\lVert w_j \rVert \lVert a^{(}}}{e^{\lVert w_j \rVert \lVert a^{(i)} \rVert \psi(\theta_j)} + \sum}$<br>$\psi(\theta_j) = (-1)^k \cos(m\theta_j) - 2k, \theta_j \in [\frac{k\pi}{m}, \frac{(k+1)\pi}{m}]$ |
| | Contrastive loss | • Siamese Networks<br>• Pairwise Data<br>• Loss Calculation<br>• Distance Measure | • Handling of Pairs<br>• Single Margin Loss<br>• Double Margin Loss<br>• Margin Parameters<br>• Stability Improvement | $L_{\text{contrastive}} = \frac{1}{2N}\sum_{i=1}^{N}[(y)d^{(i,L)} + (1-y)\max(m - d^{(i,L)}, 0)]$ |
| | | | | $L_{d-\text{contrastive}} = \frac{1}{2N}\sum_{i=1}^{N}.\sum_{l=1}^{L}[(y)\max ( - m_1, 0) + (1-y) \cdot \max (m_2 - d^{(i,l)}, 0)]$ |
| | Triplet loss | • Triplet Units (Three Instances per Loss)<br>• Distance Calculation<br>• Distance Measures | • Objective Optimization<br>• Addressing False Judgments<br>• Coupled Clusters (CC) Loss<br>• Addressing False Judgments<br>• Application: Used in tasks like re-identification, verification, and image retrieval, enhancing the model's ability to differentiate between similar and dissimilar data points effectively. | $L_{\text{triplet}} = \frac{1}{N}\sum_{i=1}^{N}.\max \{d_{(a,p)}^{(i)} - d_{(a,n)}^{(i)} + m, 0\}$<br>$L_{\text{cc}} = \frac{1}{N_p}\sum_{i=1}^{N_p}.\frac{1}{2}\max \{ \Vert z_p^{(i)} - c_p \Vert_2^2 - \Vert z_n^{(*)} - c_p \Vert_2^2 + m, 0\}$ |
| | KL Divergence | • Measure of Difference<br>• Entropy and Cross-Entropy<br>• Variational Autoencoder (VAE)<br>• Application: KLD is commonly used as a measure of information loss in the objective function of various Autoencoders (AEs), including sparse AE, Denoising AE, and Variational AE (VAE).<br>• **Functionality of GANs**<br>• Generative Tasks<br>• Discriminator and Generator | • Autoencoders (AEs)<br>• Variational Autoencoders (VAEs)<br>• Jensen-Shannon Divergence (JSD)<br>• Enhanced Learning<br>• Impact of GANs<br>• Optimization Objective<br>• Stability in Training | $D_{KL}(p \Vert q) = -H(p(x)) - E_p[\log q(x)] = \sum_x . p(x)\log\, p(x) - \sum_x . p(x)\log\, q(x) = \sum_x . p(x)\log \frac{p(x)}{q(x)}$<br>$L_{\text{vae}} = E_{z\sim q_\phi(z \mid x)}[\log\, p_\theta(x \mid z)] - D_{KL}(q_\phi(z \mid x) \Vert p(z))$<br>$D_{JS}(p \Vert q) = \frac{1}{2}D_{KL}(p(x) \Vert \frac{p(x)+q(x)}{2}) + \frac{1}{2}D_{KL}(q(x) \Vert \frac{p(x)+q(x)}{2})$ |
| | | | | $\min_{G} \max_{D} L_{\text{gan}}(D,G) = \mathbb{E}_{x\sim p(x)}[\log D(x)] + \mathbb{E}_{z\sim q(z)}[\log (1 - D(G(z)))]$ |
| Regularization | Lp-Norm | • Objective Function Modification<br>• Regularization Term<br>• Regularized Loss Equation | • Convexity for $p \geq 1$<br>• Weight Decay (`2-norm)<br>• Tikhonov Regularization<br>• Sparsity for ( p < 1 )<br>• Optimization Convenience | $E(\theta,x,y) = L(\theta,x,y) + \lambda R(\theta)$ |
| | Dropout | • Overfitting Reduction<br>• Neuron Independence<br>• Improvements to Dropout<br>▪ Fast Dropout<br>▪ Adaptive Dropout | • Prevention of Co-adaptation<br>• Enhanced Generalization<br>• Improve Efficiency<br>• Preventing Overfitting<br>• SpatialDropout | $y = r * a(W^T x)$ |
| | DropConnect | • Extension of Dropout<br>• Weight Matrix Randomization<br>• Bias Masking | • Reduction of Overfitting<br>• Increased Regularization<br>• Increased Model Robustness<br>• Variation in Architecture | $y = a((R * W)x)$ |
| Optimization | Data Augmentation | • Enhancing Training Data<br>• Transformation Techniques<br>• Greedy Selection Strategy<br>• Diffeomorphisms | • Alleviating Data Scarcity<br>• Improving Model Robustness<br>• Computational Considerations<br>• Internet-Sourced Augmentation<br>• Diverse Training Data<br>• | |
| | Weight Initialization | • Critical for Training<br>• Bias and Weight Parameters<br>• Xavier Variant for ReLU<br>• Orthonormal Matrix Initialization<br>• Symmetry Breaking<br>• Scale Consideration | • Scaling Factor<br>• Standard Methods:<br>▪ Gaussian Initialization<br>▪ Xavier Initialization<br>▪ Xavier Variants<br>▪ Orthogonal Initialization<br>▪ Layer-Sequential Unit-Variance Process<br>• Layer-Sequential Unit-Variance<br>• Convergence Improvement<br>• Stability Enhancement<br>• Performance Boost | |
| | SGD | • Parameter Updates<br>• Learning Rate Control<br>• Momentum update<br>• Nesterov Momentum | • Convergence Stability<br>• Learning Rate Challenges<br>• Training Acceleration | $\theta_{t+1} = \theta_t - \eta_t \nabla_\theta L(\theta_t; x^{(t)}, y^{(t)})$ |
| | | • Parallelization<br>• Asynchronous Updates<br>• Gradient Estimation<br>• Mini-Batch Processing | • Efficiency in Large-Scale Training. | $v_{t+1} = \gamma v_t - \eta_t \nabla_\theta L(\theta_t; x^{(t)}, y^{(t)})$<br><br>$\theta_{t+1} = \theta_t + v_{t+1}$<br><br>$v_{t+1} = \gamma v_t - \eta_t \nabla_\theta L(\theta_t + \gamma v_t; x^{(t)}, y^{(t)})$ |
| | Batch Normalization | • Normalization of Layer Inputs<br>• Normalization Step: (BN fixes the means and variances of layer inputs by computing estimations after each mini-batch.)<br>• Transformation | • Reduces Internal Covariate Shift<br>• Improves Gradient Flow<br>• Regularizes the Model<br>• Enables Saturating Nonlinearities<br>• Enhanced Training Stability<br>• Wider Applicability | Normalized Input<br>$\hat{x}_k = \frac{x_k - \mu_B}{\sqrt{\delta_B^2 + \epsilon}}$<br><br>Transformed Input<br>$y_k = BN_{\gamma,\beta}(x_k) = \gamma \hat{x}_k + \beta$ |
| | Shortcut Connection | • Alleviation of Vanishing Gradient Problem<br>• Optimization of Deep Networks<br>• Dynamic Data Routing<br>• Addressing Degradation Problem<br>• Highway Networks<br>• Residual Networks (ResNets)<br>• Pre-Activation ResNets<br>• Wide ResNets<br>• Stochastic Depth ResNets<br>• ResNets of ResNets (RiR) and ResNets of ResNets (RoR)<br>• DenseNet | • Efficient Training of Deep Networks<br>• Reduction in Parameters<br>• Enhanced Network Performance<br>• Versatility in Architecture Design<br>• Improved Performance | Highway Network output<br>$x_{l+1} = \phi_{l+1}(x_l, W_H) \cdot \tau_{l+1}(x_l, W_T) + x_l \cdot (1 - \tau_{l+1}(x_l, W_T))$<br><br>ResNet Output<br>$x_{l+1} = x_l + f_{l+1}(x_l, W_F)$ |

## 11. Conclusion

CNNs have significantly advanced the state of the art in image processing, computer vision, and a range of other ML tasks, positioning themselves as foundational architectures within DL. Over the past decade, substantial research efforts have focused on improving CNN performance through innovations in convolutional operations, pooling strategies, activation functions, normalization techniques, regularization methods, and optimization frameworks. This survey has systematically reviewed these developments, offering a comprehensive taxonomy of convolutional and pooling variants, a critical analysis of architectural evolution from early designs to attention-driven and hybrid CNN-transformer models, and an exploration of CNN applications across diverse domains, including computer vision, NLP, and medical imaging. Recent architectural advancements emphasize modular block designs and attention mechanisms, which have demonstrated superior learning capacity by enabling the network to focus on the most informative features. These design patterns have led to highly efficient, scalable, and interpretable models capable of addressing complex tasks under resource-constrained conditions. The continued refinement of modular and block-based architectures is expected to enhance model adaptability and facilitate customization for task-specific requirements. Despite these advancements, several challenges persist, including data scarcity, adversarial robustness, interpretability, and the high computational cost of deep architectures. Addressing these issues remains a central focus for the research community.

Future Direction: Future research in CNNs will focus on enhancing performance, adaptability, and efficiency. A key direction is the development of hybrid architectures that integrate convolutional networks with ViTs and GNNs, improving feature representation and generalization, particularly in multi-modal and sequential data analysis. Another priority is the design of energy-efficient, edge-optimized models through NAS, quantization, and pruning, enabling deployment on resource-constrained and embedded systems without compromising accuracy. Additionally, continual learning, lifelong learning, and federated learning frameworks will address data privacy, scalability, and catastrophic forgetting, allowing models to adapt to evolving tasks while preserving previously learned knowledge in decentralized environments.

**Acknowledgments**

We thank the Artificial Intelligence Lab, Department of Computer Systems Engineering, University of Engineering and Applied Sciences (UEAS), Swat, for providing the necessary resources.

| Abbreviation | | | |
|---|---|---|---|
| **Term** | **Definition** | **Term** | **Definition** |
| CNN | Convolutional Neural Network | ReLU | Rectified Linear Unit |
| DL | Deep Learning | RNNs | Recurrent Neural Networks |
| AI | Artificial Intelligence | NLP | Natural Language Processing |
| ML | Machine Learning | TL | Transfer Learning |
| NIN | Network in Network | ROI | Region of Interest |
| GCN | Graph Convolutional Network | SPP | Spatial Pyramid Pooling |
| MOP | Multi-Scale Orderless Pooling | PCA | Principal Component Analysis |
| LAP | Lead Asymmetric Pooling | IMP | Intermap Pooling |
| RAP | Rank-Based Average Pooling | CMPE-SE | Competitive Inner-Imaging Squeeze and Excitation for Residual Network |
| CB | Channel Boosted | RAN | Residual Attention Neural |
| OCR | Optical Character Recognition | CBAM | Convolutional Block Attention Module |
| LSTM | Long Short-Term Memory | GRUs | Gated Recurrent Units |
| NAS | Neural Architecture Search | JSD | Jensen-Shannon Divergence |
| SE-Net | Squeeze and Excitation Network | AEs | Autoencoders |
| RAN | Residual Attention Neural | OCR | Optical Character Recognition |
| FFTs | Fast Fourier Transforms | BNNs | Binarized Neural Networks |
| VQ | Vector quantization | VQA | Visual Question Answering |
| CC | Coupled Clusters | VAEs | Variational Autoencoders |

| **Symbol** | **Description** |
|---|---|
| $X$ | Total $x$ coordinates of an image |
| $x$ | $x^{th}$ coordinate under consideration of an image |
| $Y$ | Total $y$ coordinates of an image |
| $y$ | $y^{th}$ coordinate under consideration of an image |
| $c$ | Channel index |
| $i_C(x,y)$ | $(x, y)$ element of $c^{th}$ channel of an image |
| $L$ | Total number of layers |
| $l$ | Layer number |
| $K_l$ | Total number of kernels of $l^{th}$ layer |
| $k_l$ | Kernel number of $l^{th}$ layer |
| $U$ | Total number of rows of $k^{th}$ kernel |
| $u$ | $u^{th}$ row under consideration |
| $V$ | Total number of columns of $k^{th}$ kernel |
| $v$ | $v^{th}$ column under consideration |
| $e_l^k(v,u)$ | $(u,v)$ element of $k^{th}$ kernel of $l^{th}$ layer |
| $F_l^k$ | Input feature matrix for $l^{th}$ layer and $k^{th}$ neuron |
| $P$ | Total number of rows of feature matrix |
| $p$ | $p^{th}$ row under consideration |
| $Q$ | Total number of columns of feature matrix |
| $q$ | $q^{th}$ column under consideration |
| $f_l^k(p,q)$ | $(p,q)$ element of feature matrix |
| $g_c(.)$ | Convolution operation |
| $g_p(.)$ | Pooling operation |
| $g_a(.)$ | Activation function |
| $g_k(.)$ | Concatenation operation |
| $\mathrm{g}_{\mathrm{t_g}}$ | Transformation gate |
| $\mathrm{g}_{\mathrm{c_g}}$ | Carry gate |
| $g_{sq}(.)$ | Squeeze operation |
| $g_{ex}(.)$ | Excitation operation |
| $\mathbf{Y}_{l+1}^{K}$ | Weight vector showing feature-maps importance learned using SE operation |
| $\mathrm{g}_{\mathrm{t}}$ | Transformation function for two layer NN implemented by SE block |
| $\mathrm{g}_{\mathrm{s_g}}$ | Sigmoid gate implemented by SE block |
| $g_{sm}$ | Soft mask |
| $g_{tm}$ | Trunk mask |
| $\mathbf{I}_B$ | Channel boosted input tensor. |